\documentclass[letterpaper]{article} %
\usepackage{aaai24}  %
\usepackage{times}  %
\usepackage{helvet}  %
\usepackage{courier}  %
\usepackage[hyphens]{url}  %
\usepackage{graphicx} %
\usepackage{natbib}  %
\usepackage{caption} %
\usepackage{algorithm}

\usepackage{newfloat}
\usepackage{listings}
\DeclareCaptionStyle{ruled}{labelfont=normalfont,labelsep=colon,strut=off} %
\floatstyle{ruled}
\newfloat{listing}{tb}{lst}{}
\floatname{listing}{Listing}
\usepackage[usenames,dvipsnames]{xcolor}
\usepackage{longtable}
\usepackage{array}
\usepackage{booktabs}
\usepackage{caption}
\usepackage{colortbl}
\usepackage{multirow}
\usepackage{diagbox}
\definecolor{rc1}{RGB}{235,235,235}
\definecolor{rc2}{RGB}{255,255,255}
\definecolor{fade}{gray}{0.4}
\definecolor{green}{rgb}{0.0, 0.5, 0.0}
\definecolor{grey}{rgb}{0.5, 0.5, 0.5}
\newcommand\xkd{XKD~}

\newcommand{\decblue}[1]{\textcolor{black}{${(\downarrow\!#1)}$}}

\usepackage{soul}

\usepackage{fontawesome}

\usepackage{pifont}%
\usepackage{graphicx}
\usepackage{amsmath}
\usepackage{booktabs}
\usepackage{enumitem}
\usepackage{longtable}
\usepackage{array}
\usepackage{booktabs}
\usepackage{caption}
\usepackage{color, colortbl}
\usepackage{multirow}
\usepackage{diagbox}
\usepackage{algorithm}
\usepackage{listings}
\definecolor{rc1}{RGB}{235,235,235}
\definecolor{rc2}{RGB}{255,255,255}
\definecolor{codeblue}{rgb}{0.25,0.5,0.5}
\definecolor{codekw}{rgb}{0.85, 0.18, 0.50}
\usepackage{booktabs}       %
\usepackage{amsfonts}       %
\usepackage{nicefrac}       %
\usepackage{microtype}      %
\usepackage{graphicx}
\usepackage{tikz}
\usepackage{comment}
\usepackage{amsmath,amssymb} %
\usepackage{color}

\newcommand{\loss}{\mathcal{L}}

\usepackage{stackengine}

\usepackage{algpseudocode}
\newcommand{\probP}{\text{I\kern-0.15em P}}
\usepackage{tablefootnote} %
\newcommand\fadetext[1]{\textcolor{black}{#1}} %

\usepackage{threeparttable}
\usepackage{multicol}

\usepackage{amsmath,amsfonts,bm}

\def\eqref#1{equation~\ref{#1}}
\def\Eqref#1{Equation~\ref{#1}}

\def\1{\bm{1}}

\DeclareMathAlphabet{\mathsfit}{\encodingdefault}{\sfdefault}{m}{sl}
\SetMathAlphabet{\mathsfit}{bold}{\encodingdefault}{\sfdefault}{bx}{n}

\def\emA{{A}}

\newcommand{\E}{\mathbb{E}}

\newcommand{\R}{\mathbb{R}}

\newcommand{\softmax}{\mathrm{softmax}}

\renewcommand\fadetext[1]{\textcolor{gray}{#1}} %

\title{XKD: Cross-modal Knowledge Distillation with Domain Alignment for Video Representation Learning}
\author {
    Pritam Sarkar\textsuperscript{\rm 1, \rm 2},
    Ali Etemad\textsuperscript{\rm 1}
}
\affiliations {
    \textsuperscript{\rm 1} Queen's University, Canada\\
    \textsuperscript{\rm 2} Vector Institute\\
    {\{pritam.sarkar, ali.etemad\}@queensu.ca}
    \\{\textcolor{magenta}{\bf\small\texttt{https://pritamqu.github.io/XKD}}}
}

\usepackage{bibentry}

\begin{document}

\maketitle

\begin{abstract}

We present XKD, a novel self-supervised framework to learn meaningful representations from unlabelled videos. XKD is trained with two pseudo objectives. First, masked data reconstruction is performed to learn modality-specific representations from audio and visual streams. Next, self-supervised cross-modal knowledge distillation is performed between the two modalities through a teacher-student setup to learn complementary information. We introduce a novel domain alignment strategy to tackle domain discrepancy between audio and visual modalities enabling effective cross-modal knowledge distillation.
Additionally, to develop a general-purpose network capable of handling both audio and visual streams, modality-agnostic variants of XKD are introduced, which use the same pretrained backbone for different audio and visual tasks. Our proposed cross-modal knowledge distillation improves video action classification by $8\%$ to $14\%$ on UCF101, HMDB51, and Kinetics400. Additionally, XKD improves multimodal action classification by $5.5\%$ on Kinetics-Sound. XKD shows state-of-the-art performance in sound classification on ESC50, achieving top-1 accuracy of $96.5\%$.
\end{abstract}

\section{Introduction} \label{sec:introduction}

Self-supervised learning aims to learn meaningful representations from unlabelled data with no human supervision \cite{simclr,simsiam,pirl,byol,mae}. Using self-supervision, recent multimodal methods \cite{avid,xdc,ravid,cmac,cmacc,crisscross} have shown great promise in learning effective representations from videos. In general, multimodal frameworks leverage the existing information in multiple data streams to learn better representations for downstream tasks toward either modality (or all). Recent audio-visual frameworks aim to perform \textit{information sharing} between networks to further enrich the learned representations \cite{afouras2020asr,chen2021distilling,ren2021learning,aytar2016soundnet,albanie2018emotion}. 
However, effective knowledge sharing between audio and video is particularly challenging due to the inherent diversity, complexity, and domain-specific nature of each modality, as well as the existence of substantial domain gaps between them \cite{ren2021learning,chen2021distilling}.

In this work, we aim to perform effective information sharing between audio and video streams to obtain more generalized representations for downstream tasks. To this end, we propose a novel self-supervised framework called XKD, which stands for Cross-modal Knowledge Distillation. Our approach consists of two sets of pseudo-tasks, (\textit{i}) masked data modelling and (\textit{ii}) cross-modal knowledge distillation. The former is performed to learn modality-specific (MS) information, while the latter distills and transfers knowledge across modalities to further enrich the learned representations. To allow for stable and effective information exchange between modalities, we introduce a domain alignment strategy. The proposed 
strategy involves $2$ steps (\textit{i}) feature refinement that identifies \textit{`what to transfer'} based on cross-modal feature relevance and (\textit{ii}) minimizing domain discrepancy to align the two representations. 
Additionally, we introduce modality-agnostic (MA) variants of our method to tackle the challenging task of learning from both modalities using a unified network (shared backbone). Modality-agnostic methods are particularly useful given their ability to accept different data streams and solve a variety of tasks using the same backbone. Moreover, they ease the challenge of designing individual models for each modality, thus attracting attention in recent studies \cite{omnivore,imagebind,vatt}.

Inspired by the recent success of Transformers in different domains \cite{bert,ast,mae,videomae2}, we use ViT \cite{vit} as the backbone of our framework for both audio and visual modalities. We use the $3$ different sizes of datasets to pretrain the framework, including AudioSet, Kinetics400, and Kinetics-Sound. 
The pretrained backbones are then evaluated on multiple datasets on a variety of downstream tasks. More specifically, UCF101, HMDB51, and Kinetics400 are used for video-related tasks; ESC50 and FSD50K are used for audio-related tasks; and Kinetics-Sound is used for multimodal evaluation.

\noindent Our contributions are summarized as follows:
\begin{itemize}[noitemsep,nolistsep,leftmargin=*]
    \item We introduce XKD, a self-supervised framework for video representation learning, which uses a novel domain alignment strategy to enable self-supervised cross-modal knowledge distillation between audio and video as two heterogeneous modalities.
    The proposed domain alignment strategy minimizes the domain gap and identifies the most transferable features between the two domains for effective cross-modal knowledge distillation.

    \item Rigorous experiments and thorough ablations are performed to analyse the proposed method. XKD achieves state-of-the-art or competitive performance on a variety of downstream tasks including video action recognition, sound classification, and multimodal fusion. %
    \item Our proposed modality-agnostic variants achieve very competitive performance compared to the modality-specific ones, enabling the use of a single pretrained encoder for a variety of audio and visual downstream tasks.

\end{itemize}

\noindent The code, pretrained models, and supplementary material are available on the project website.


\section{Related Work} \label{sec:background}

\noindent\textbf{Video self-supervised learning.}
Self-supervised learning has been widely used in a variety of different areas, including the challenging task of video representation learning \cite{schiappa2022self}. Several prior works have attempted to learn video representations through both uni-modal \cite{feichtenhofer2021large,cvrl,rotnet3d} as well as multimodal \cite{avid,mmv,brave,xiao2022maclr,coclr} pretraining. For example, prior works have explored self-supervised frameworks for learning video representations through contrastive \cite{avid,ravid,mmv}, non-contrastive \cite{crisscross,brave}, and deep clustering \cite{xdc,selavi} techniques.

\noindent\textbf{Masked data modelling.} Inspired by the success of BERT \cite{bert} in natural language processing, several prior works have attempted to learn meaningful representations through the reconstruction of masked (corrupted) inputs. Such methods employ encoder-decoder setups, where the encoder compresses the inputs into a lower dimension, while the decoder is trained for reconstruction. This simple approach shows promise in different domains including image \cite{beit,mae,bachmann2022multimae}, video \cite{bevt,videomae,videomae2}, and audio \cite{mae_ast_long,ast,chong2022masked} among others.

\noindent\textbf{Cross-modal knowledge distillation.}
The main goal of cross-modal knowledge distillation is to transfer knowledge across different modalities \cite{afouras2020asr,dai2021learning,chen2021distilling,ren2021learning,aytar2016soundnet,albanie2018emotion,elo}. For example, \cite{afouras2020asr,ren2021learning} attempted knowledge distillation between audio teachers and video students to improve visual representations. Cross-modal knowledge distillation amongst different visual modalities has been performed in \cite{dai2021learning}, where a pretrained optical flow teacher is used to improve RGB student's representation. %

\noindent\textbf{Modality-agnostic networks.}
While modality-specific models have been the preferred approach toward developing representation learning solutions due to their strong performance, they do not have the ability to learn from multiple modalities using the same backbone, which makes them harder to develop. 
Recently, \cite{omnivore,vatt} introduced modality-agnostic models capable of learning different modalities using the same backbones. In our paper, we attempt to develop modality-agnostic variants of our solution which can handle $2$ very different modalities, i.e., audio and video to solve a variety of downstream tasks.

\begin{figure*}[t]
    \centering
    \includegraphics[width=0.8\linewidth]{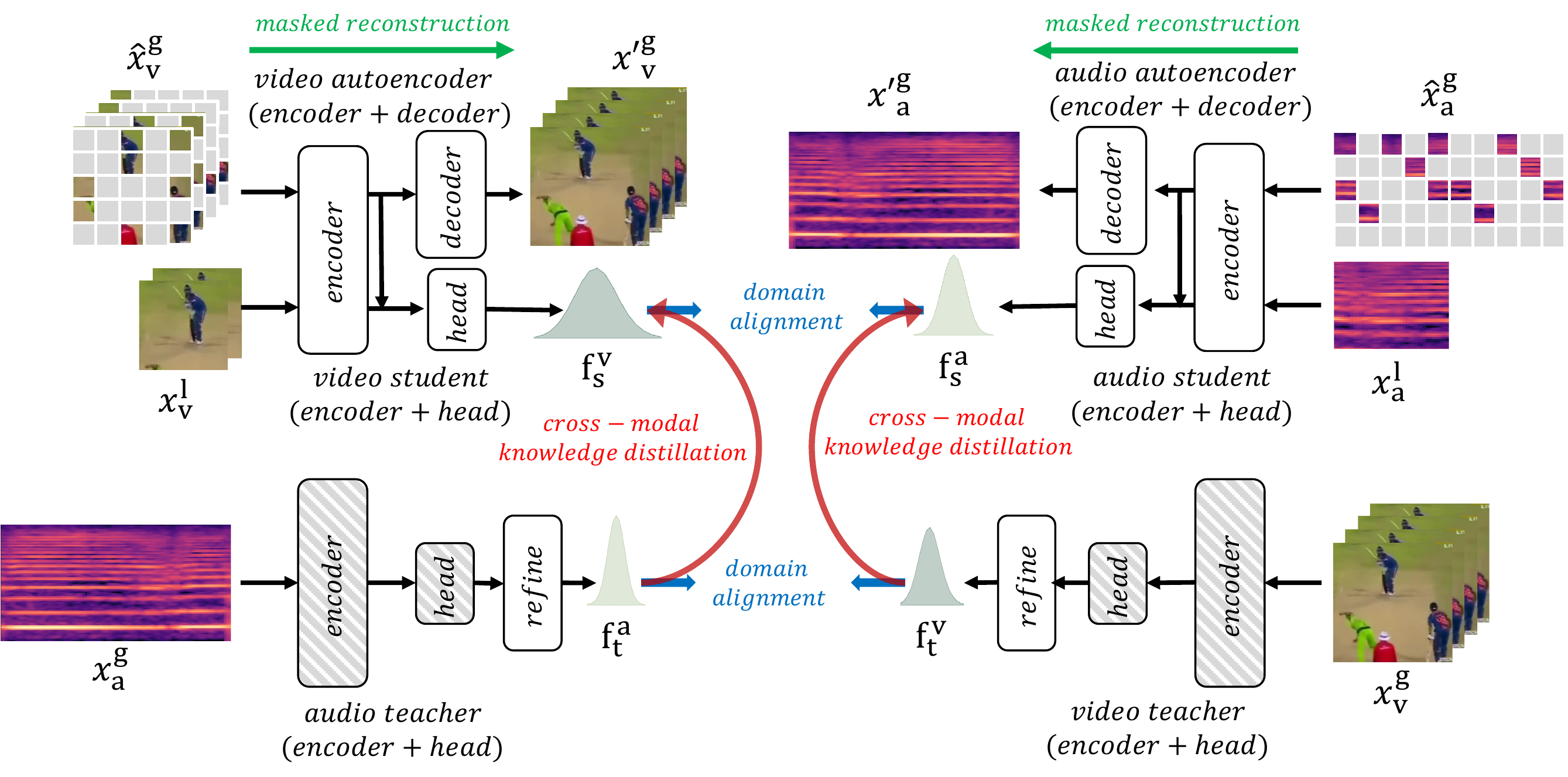}
    \caption{An \textbf{overview} of our proposed framework, which consists of $3$ steps.
    \textbf{Masked reconstruction}: autoencoders are used to learn representations from individual modalities through reconstruction of highly masked inputs. 
    \textbf{Domain alignment}:
    To enable cross-modal knowledge distillation, two domains are aligned through feature refinement and minimizing domain discrepancies. 
    \textbf{Cross-modal knowledge distillation}: The students are used to distil knowledge from their respective cross-modal teachers. %
    }
    \label{fig:xkd}
\end{figure*}

\section{Method} \label{sec:method}

\subsection{Overview}
Figure \ref{fig:xkd} presents an overview of our framework. Our method consists of two sets of autoencoders for audio ($\theta_{ae}^a$) and video ($\theta_{ae}^v$). First, we train these autoencoders to reconstruct from the masked inputs, which helps the encoders ($\theta_{e}^a$ and $\theta_{e}^v$) to learn modality-specific information. Next, to transfer knowledge between modalities to learn complementary information, we align the two domains by identifying the most transferable features and minimizing the domain gaps. Finally, audio ($\theta_{t}^a$) and video ($\theta_{t}^v$) teachers are employed to provide cross-modal supervision to the opposite modalities. The details of our proposed method are mentioned below.

\subsection{Data Embedding} \label{sec:Data Embedding}
Let be given video clip $x$, where the visual frames and audio spectrograms are denoted as $x_v$ and $x_a$, respectively. We create `global' and `local' views from each modality which are then used by the teachers and students respectively. First, we apply augmentations on $x_v$ and $x_a$ to generate the global views as $x_v^g$ and $x_a^g$. Next, we generate $n$ local views $x_v^l$ and $x_a^l$ from $x_v$ and $x_a$ respectively, where $x_v^l=\{x_{v1}^l, \dots, x_{vn}^l\}$ and $x_a^l=\{x_{a1}^l, \dots, x_{an}^l\}$. Specifically, $x_{ai}^l$ is an augmented time-frequency crop of $x_{a}$ and $x_{vi}^l$ is an augmented spatio-temporal crop of $x_{v}$. To further elaborate, $x_{v}^l$ and $x_{a}^l$ are differently augmented than $x_{v}^g$ and $x_{a}^g$. Both local and global views of audio and visual inputs are projected into an embedding space. For example, $x_a \in \R^{F \times T_a}$ are reshaped into $N_a$ smaller patches of size $f \times t_a$, where $N_a = F/f \times T_a/t_a$. Similarly, we reshape the videos $x_v \in \R^{T_v\times H\times W\times C}$ into $N_v$ smaller cuboids of size $t_v\times\!h\times\!w\times\!C$, where $N_v = T_v/t_v\!\times\!H/h\!\times\!W/w\!\times\!C$. Finally, the spectrogram patches and visual cuboids are flattened into vectors and linearly projected onto the embedding space, which are then fed to the encoders.

\subsection{Masked Data Modelling}
Inspired by the recent success {and scalability }of pretraining with masked reconstruction in different domains \cite{bert,beit,bevt,ast,data2vec,mae,mae_ast_long,videomae}, {we adopt masked data modelling in our framework} to learn modality-specific representations. 
The masked reconstruction employs an autoencoder $\theta_{ae}$, which consists of an encoder $\theta_{e}$ and a decoder $\theta_{d}$. Let $x$ be the input, which can be further expressed as a sequence of tokens $\{x_i\}_{i=1}^N$, as described in Section \ref{sec:Data Embedding}. Here, we randomly mask some of the input tokens with $m \in \{0, 1\}^N$, hence the masked tokens $x^{[m]}$ are represented as $ \{x_i | m_i=1\}_{i=1}^N$ while the corrupted inputs $\hat{x}$ are represented as $\{x_i|m_i=0\}_{i=1}^N$. Further, we drop the masked tokens $x^{[m]}$ before feeding the input to $\theta_{ae}$ for computational efficiency \cite{vatt,mae}. We train $\theta_{ae}$ to reconstruct $x$, based on input $\hat{x}$. Here, $\theta_{ae}$ is trained to minimize the reconstruction loss $\loss_\mathrm{recon}$ as:
\begin{equation} \label{eq:recon_loss}
\fontsize{9pt}{10pt}\selectfont
    \loss_\mathrm{recon}(\theta_{ae}(\hat{x}), x^{[m]}) = \frac{1}{N_m}\sum_{i=1}^{N_m}(\theta_{ae}(\hat{x}_i) - x_i^{[m]})^{2}.
\end{equation}
In particular, using \Eqref{eq:recon_loss}, we define the video and audio reconstruction losses as $\loss_\mathrm{recon}^v$ and $\loss_\mathrm{recon}^a$ for given inputs $x_v^g$ and $x_a^g$, to train $\theta_{ae}^v$ and $\theta_{ae}^a$, respectively. Here $\theta_{ae}^v$ and $\theta_{ae}^a$ denote video and audio autoencoders. To jointly train the audio-visual masked autoencoder, we define the final reconstruction loss $\loss_\mathrm{ae}$ as:
\begin{equation}
\fontsize{9pt}{10pt}\selectfont
    \mathbf{\loss_\mathrm{\bf ae}\!=\!\loss_\mathrm{\bf recon}(\theta_{ae}^v(\hat{x}_v^g), x_v^{g[m]}) + \loss_\mathrm{\bf recon}(\theta_{ae}^a(\hat{x}_a^g), x_a^{g[m]})}.
\end{equation}

\subsection{Knowledge Distillation w/ Domain Alignment}
To facilitate cross-modal knowledge sharing, we adopt a teacher-student setup \cite{mean_teacher}. The teacher ($\theta_t$) and student ($\theta_s$) are comprised of a backbone and a projector head ($\theta_{h}$), where the teacher and student network architectures are the same but differently parameterized. Moreover, we parameterize $\theta_{s}$ as $\{\theta_{e}$, $\theta_{h}\}$, where $\theta_{e}$ is the same encoder used in reconstruction (explained in the previous subsection) and $\theta_{h}$ is a newly added projector head. We define $\theta_s^v$ and $\theta_s^a$ as video and audio students, whereas, $\theta_t^v$ and $\theta_t^a$ are denoted as video and audio teachers.

As mentioned earlier, cross-modal knowledge distillation between audio and video is particularly difficult due to the inherent diversity, complexity, and domain-specific nature of these two modalities. To tackle this, we perform domain alignment by identifying the most transferable features and minimizing the domain discrepancy. The proposed domain alignment ensures meaningful target distributions are set by the teachers in order to perform successful cross-modal knowledge distillation.

\noindent \textbf{Domain Alignment.}
Both the audio and video carry a rich and diverse set of information about the source. Therefore, first, we identify the most transferable features by re-weighting the teachers' representations based on their cross-modal feature importance through a soft-selection process. Specifically, we obtain the cross-modal attention maps to calculate the cross-modal feature importance with respect to the corresponding modalities. In order to calculate the cross-modal attention maps, we first extract the modality-specific attention maps ($\emA$) from the last attention layers of the teacher networks as: 
\begin{equation} \label{eq:attn_map}
\fontsize{9pt}{10pt}\selectfont
\begin{split}
    \emA = \softmax(Q^{\mathrm{[CLS]}} \cdot \frac{K^T}{\sqrt{d}} ).
\end{split}
\end{equation}
Here, $Q$ denotes the query, $K$ is the key, and $V$ is the value. Specifically, $\emA\in \R^{H\times N}$ is calculated as the correlation between the query ($Q$) embedding of the class token (\texttt{CLS}) and the key ($K$) embeddings of all the other patches or cuboids. Note, $H$ denotes the number of attention heads and $N$ denotes the number of patches or cuboids. We obtain the visual attention $\emA_v \in \R^{H\times N_v}$ and audio attention $\emA_a \in \R^{H\times N_a}$ as per \Eqref{eq:attn_map}. Next, we obtain the respective cross-modal attention maps as $\emA_{v}^\times$ and $\emA_{a}^\times$ as:
\begin{equation}\label{eq:cmattn}
\fontsize{9pt}{10pt}\selectfont
\begin{split}
    \emA_{v}^\times = \mathrm{MeanPool}(\emA_{v} \cdot \emA_{a}^T) / \mathrm{scale_v}; %
    \\
    \emA_{a}^\times = \mathrm{MeanPool}(\emA_{a} \cdot \emA_{v}^T) / \mathrm{scale_a}.
\end{split}
\end{equation}
Here, $\emA_{v} \cdot \emA_{a}^T \in \R^{H\times N_v \times N_a}$ and $\emA_{a} \cdot \emA_{v}^T \in \R^{H\times N_a \times N_v}$, we apply $\mathrm{MeanPool}$ across the last dimension. Additionally, $\mathrm{scale_v}$ and $\mathrm{scale_a}$ are scaling factors, obtained as $\frac{1}{N_v}\sum_{i=1}^{N_v} \emA_{v}$ and $\frac{1}{N_a}\sum_{i=1}^{N_a} \emA_{a}$, used to re-scale the computed cross-modal attention maps back to their original range for numerical stability.
We identify the most transferable features obtained from the teachers as $f_{t}^v$ and $f_{t}^a$, as $\mathrm{refine}(\theta_t^v(x_v^g), \emA_{v}^\times)$  and $\mathrm{refine}(\theta_t^a(x_a^g), \emA_{a}^\times)$ respectively. Here $\emA_{v}^\times$ and $\emA_{a}^\times$ are used to re-weight the visual and audio representations respectively. We formulate $\mathrm{refine}$ as:
\begin{equation} \label{eq:refine}
\fontsize{9pt}{10pt}\selectfont
    \mathrm{refine}(\theta_t(x^g), \emA^\times) = \theta_t(x^g) \cdot \frac{1}{H}\sum_{h=1}^{H}\emA_h^\times \cdot \Omega,
\end{equation}
{where $\emA_h^\times$ represents the cross-modal attention of each head} and $\Omega$ is a non-negative scalar defined as the ratio of prior and posterior energy, expressed as: 
\begin{equation} \label{eq:refine_scaler}
\fontsize{9pt}{10pt}\selectfont
    \Omega = \frac{\parallel\! \theta_t(x^g) \!\parallel_2^2}{\parallel\!\theta_t(x^g) \cdot \emA^\times\!\parallel_2^2}.
\end{equation}
Next, to improve the knowledge transferability, we reduce the domain gaps by minimizing the Maximum Mean Discrepancy (MMD) \cite{mmd} loss, estimated as:
\begin{equation} \label{eq:mmd}
\fontsize{8.2pt}{10pt}\selectfont
    \loss_\mathrm{mmd}(\probP_a, \probP_v) = \parallel\!\E_{x_a\sim \probP_a} [k(\cdot, x_a)] - \E_{x_v\sim \probP_v}[k(\cdot, x_v)]\parallel\!_{\mathcal{H}_k}.
\end{equation}
Here $x_a$ and $x_v$ are drawn from distributions $\probP_a$ and $\probP_v$ respectively. Additionally, $\parallel \cdot \parallel_{\mathcal{H}_k}$ is the RKHS norm \cite{mmd} and and $k$ represents the Gaussian kernel with bandwidth $\sigma$, written as:
\begin{equation} \label{eq:gauss_kernel}
\fontsize{9pt}{10pt}\selectfont
    k(x_a, x_v) = \exp\left(\frac{-\parallel x_a-x_v\parallel^2}{2\sigma^2}\right).
\end{equation}
Using \Eqref{eq:mmd}, we define {domain alignment} loss $\loss_\mathrm{da}$ as:
\begin{equation} \label{eq:align_loss}
\fontsize{9pt}{10pt}\selectfont
\begin{split}
    \mathbf{\loss_\mathrm{\bf da} = \loss_\mathrm{\bf mmd}(f_s^v, f_s^a) + \space \loss_\mathrm{\bf mmd}(f_t^v, f_t^a)}.
\end{split}
\end{equation}
Here, $f_s^a$ and $f_s^v$ refer to audio and visual representations obtained from $\theta_s^a$ and $\theta_s^v$ respectively.

\noindent \textbf{Knowledge Distillation.}
We perform knowledge distillation between modalities to provide cross-modal supervision in order to learn complementary information. %
We train the student networks to match the distribution of the cross-modal teacher networks. Specifically, we train $\theta_{s}^v$ to match the distribution of $\theta_{t}^a$, while $\theta_{s}^a$ is trained to match the distribution of $\theta_{t}^v$. 
Following \cite{dino},
we normalize $f_{t}^a$ and $f_{t}^v$ with a `mean' calculated based on the current batch statistics, which helps the teachers' outputs be more uniformly distributed. Additionally, to prevent abrupt updates of batch means, we slowly update the `means' using an exponential moving average (EMA). Finally, we minimize the cross-entropy loss $H(a,b)$ formulated as $-a\log b$, where a and b represent the output probability distributions of $\theta_t$ and $\theta_s$ respectively. 
Here, the probability $P$ over $J$ distributions is obtained by using a $\softmax$ function with the temperature parameter $\tau$, where $0<\tau<1$ used to sharpen the distribution as:
\begin{equation} \label{eq:softmax_temp}
\fontsize{9pt}{10pt}\selectfont
    P(f^{(i)}) = \frac{\exp(f^{(i)}/\tau)}{\sum_{j=1}^J\exp(f^{(j)}/\tau)}.
\end{equation}
Accordingly, we define the cross-modal knowledge distillation loss $\loss_\mathrm{kd}$ as:
\begin{equation} \label{eq:cmkd_loss}
\fontsize{9pt}{10pt}\selectfont
\begin{split}
   \mathbf{\loss_\mathrm{\bf kd} = -P(f_t^v)\log(P(f_s^a))
    -P(f_t^a)\log(P(f_s^v))}.
\end{split}
\end{equation}

\subsection{Final Loss}
To train {XKD}, we define the final loss function as the combination of reconstruction, domain alignment, and cross-modal knowledge distillation losses expressed as:
\begin{equation} \label{eq:final_loss}
\fontsize{9pt}{10pt}\selectfont
\mathbf{\loss_\mathrm{\bf xkd} = \lambda_\mathrm{\bf ae}\!\times\!\loss_\mathrm{\bf ae} \!+\!  \lambda_\mathrm{\bf da}\!\times\!\loss_\mathrm{\bf da} \!+\! \lambda_\mathrm{\bf kd}\!\times\!\loss_\mathrm{\bf kd}}.
\end{equation}
Here, $\lambda_\mathrm{ae}$, $\lambda_\mathrm{da}$, and $\lambda_\mathrm{kd}$ are the loss coefficients corresponding to the three loss terms, respectively. Empirically, we set $\lambda_\mathrm{ae}$, $\lambda_\mathrm{da}$, and $\lambda_\mathrm{kd}$ as $5$, $1$, and $1$ respectively. It should be noted that $\loss_\mathrm{xkd}$ is only used to train $\theta_s$ and $\theta_{ae}$, not $\theta_t$. EMA \cite{byol,mean_teacher,mocov3} is used to slowly update $\theta_t^a$ and $\theta_t^v$ as:
\begin{equation} \label{eq:ema}
\fontsize{9pt}{10pt}\selectfont
\begin{split}
    \theta_t^a \leftarrow \lambda_a \theta_t^a + (1-\lambda_a) \theta_s^a\;\text{and}\; %
    \theta_t^v \leftarrow \lambda_v \theta_t^v + (1-\lambda_v) \theta_s^v,
\end{split}   
\end{equation}
where $\lambda_a$ and $\lambda_v$ are the EMA coefficients corresponding to $\theta_t^a$ and $\theta_t^v$.
We present the proposed algorithm in Alg. \ref{algo:pseudocode}.

\begin{algorithm}[t]
\algdef{SE}[SUBALG]{Indent}{EndIndent}{}{\algorithmicend\ }%
\algtext*{Indent}
\algtext*{EndIndent}
	\small
	\begin{algorithmic}[1]
		\State \textbf{Input}: $x_v$, $x_a$
		\State \textbf{Initalize}: $\theta_{d}^v$, $\theta_{d}^a$, $\theta_{s}^v$,                                 $\theta_{s}^a$, $\theta_{t}^v$, $\theta_{t}^a$
		\For{t \textbf{in} iterations}
            \State Obtain $x_v^g$ and $x_v^l$ from $x_v$
            \State Obtain $x_a^g$ and $x_a^l$ from $x_a$
		\Statex \textbf{\phantom{for }\fadetext{\# masked data reconstruction}} 
            \State $f_{sv}^g\!=\!\theta_{e}^v(\hat{x}_v^{g})$; $f_{sa}^g = \theta_{e}^a(\hat{x}_a^{g})$ 
            \Statex \textbf{\phantom{for }\fadetext{\# total mask reconstruction loss}} 
            \State $\loss_{ae}\!=\!\loss_{recon}(\theta_{d}^v(f_{sv}^g), x_v^{g[m]})\!+\!\loss_{recon}(\theta_{d}^a(f_{sa}^g), x_a^{g[m]})$
            
            \Statex \textbf{\phantom{for }\fadetext{\# cross-modal knowledge distil. w/ domain alignment}} 
            \State {with no-gradient:}
            \Indent
            \State $f_{tv}^g =\theta_{t}^v(x_v^g)$; $f_{ta}^g =\theta_{t}^a(x_a^g)$
            \State Obtain $\emA_{v}^\times$, $\emA_{a}^\times$ \Comment{Eqn. \ref{eq:cmattn}}%
            \Statex \textbf{\phantom{for }\phantom{for }\fadetext{\# feature refinement}} 
            \State $f_t^v\!=\!\mathrm{refine}(f_{tv}^g, \emA_{v}^\times)$ \Comment{Eqn. \ref{eq:refine}}%
            \State $f_t^a\!=\!\mathrm{refine}(f_{ta}^g, \emA_{a}^\times)$ \Comment{Eqn. \ref{eq:refine}}%
            \EndIndent
            
            \State $f_{sv}^l\!=\!\theta_{s}^v(x_v^l)$; $f_{sa}^l\!=\!\theta_{s}^a(x_a^l)$
            \State $f_s^v\!=$ Concat $(\theta_h^v(f_{sv}^g),f_{sv}^l)$
            \State $f_s^a\!=$ Concat $(\theta_h^a(f_{sa}^g), f_{sa}^l)$
            \Statex \textbf{\phantom{for }\fadetext{\# domain alignment loss}} 
            \State $\loss_{da}\!=\!\loss_{mmd}(f_s^v, f_s^a)\!+\!\space \loss_{mmd}(f_t^v, f_t^a)$
            \Statex \textbf{\phantom{for }\fadetext{\# cross-modal knowledge distillation loss}} 
            \State $\loss_{kd}\!=\!-P(f_t^v)\log(P(f_s^a))\!-\!P(f_t^a)\log(P(f_s^v))$
            \Statex \textbf{\phantom{for }\fadetext{\# final loss}} 
            \State $\loss_{xkd} = \lambda_{ae}\!\times\!\loss_{ae} \!+\!  \lambda_{da}\!\times\!\loss_{da} \!+\! \lambda_{kd}\!\times\!\loss_{kd}$

            \State Update $\theta_{d}^v$, $\theta_{d}^a$, $\theta_{s}^v$, $\theta_{s}^a$ based on $\loss_{xkd}$

            \State Update $\theta_{t}^v$ and $\theta_{t}^a$
            \Comment{Eqn. \ref{eq:ema}}%

		\EndFor
		\State \textbf{Output}: $\theta_{s}^v$, $\theta_{s}^a$, $\theta_{t}^v$, $\theta_{t}^a$
	\end{algorithmic}
	\caption{XKD Training Algorithm.}
	\label{algo:pseudocode}
\end{algorithm}

\subsection{Modality-agnostic Variant}
We take our proposed approach a step forward and attempt to train it in a modality-agnostic fashion with the goal of developing a `general' network capable of handling both modalities. This is a very challenging task in the context of our work given the very diverse nature of audio and video streams. We introduce two modality-agnostic variants XKD-MATS and XKD-MAS.
As the name suggests, in {XKD-MATS} the audio and visual teachers share their backbones, and so do the audio and visual students. In the case of {XKD-MAS}, the audio and visual students share their backbones, while the audio and visual teachers use modality-specific backbones. Please note that in all the setups, we use the \Eqref{eq:ema} to update the teachers using their respective students. Moreover, given the need to reconstruct different modalities, all of the variants use modality-specific decoders and input projection layers. The rest of the setups remain the same as the original XKD. Both variants are trained with $\loss_\mathrm{xkd}$ (see \Eqref{eq:final_loss}).

\section{Experiments and Results} \label{sec:experiment}

\subsection{Implementation Details}
\noindent\textbf{Datasets.}
We pretrain XKD on 3 datasets of different sizes including the small-scale \textbf{Kinetics-Sound} \cite{l3-kineticssound}, large-scale \textbf{Kinetics400} \cite{kinetics400}, and very large-scale \textbf{AudioSet} \cite{audioset}. We evaluate our proposed method on a variety of downstream tasks including {video action recognition}, {sound classification}, and {multimodal action classification}. We use a total of $6$ datasets for downstream tasks, namely \textbf{Kinetics400} (K$400$) \cite{kinetics400}, \textbf{Kinetics-Sound} (KS) \cite{l3-kineticssound}, \textbf{UCF101} (U$101$) \cite{ucf101}, \textbf{HMDB51} (H$51$) \cite{hmdb}, \textbf{ESC50} (E$50$) \cite{esc}, and \textbf{FSD50K} (F$50$K) \cite{fsd50k}.
The dataset details are provided in the supplementary material (Suppl. Mat.) Sec. \ref{supsec:dataset}. Unless mentioned otherwise, Kinetics400 is used for pretraining.

\noindent\textbf{Input setup.}
During pretraining, to save computation
we downsample the video input at $8$ FPS and resize the frame resolution at $112^2$. Additionally, we re-sample the audio waveforms at $16$ kHz. and generate mel-spectrograms using $80$ mel filters.
Next, we create global and local views for both audio and video. We use $4$ seconds of audio-visual input for the global views. Followed by the local views are generated by taking random temporal segments of $1$ second unless stated otherwise. The final input sizes to the teachers are $3\!\times\!32\!\times\!112^2$ and $80\!\times\!448$. Similarly, the input sizes to the students are $3\!\times\!8\!\times\!96^2$ and $80\!\times\!112$, for video and audio respectively. Moreover, the inputs to the encoders for masked reconstructions are the same as the input to the teacher networks, except they are heavily masked. We use a patch size of $4\!\times\!16$ for audio spectrograms and a cuboid size of $4\!\times16^2$ for video input. Please see additional details in the Suppl. Mat. Sec.  \ref{supsec:augmentation} and \ref{supsec:pretraining}.

\noindent\textbf{Architecture.}
We choose ViT \cite{vit} as the backbone for both audio and video, due to its stability in performance across different data streams \cite{vatt}. In particular, we experiment with two ViT variants \textbf{ViT-B} and \textbf{ViT-L}. By default, ViT-B is used as the video backbone for pretraining with Kinetics400 and Kinetics-Sound, whereas, ViT-Large is used when pretrained with AudioSet. In all the setups, ViT-B is used as the audio backbone. The additional details are in the Suppl. Mat. Sec.  \ref{supsec:pretraining} and \ref{supsec:modality_agnostic}. 

\noindent\textbf{Evaluation.} Following \cite{crisscross,avid,xdc,brave}, we evaluate XKD in both linear and finetuning setup. We redirect readers to see the details on evaluation protocols in the Suppl. Mat. Sec. 
\ref{supsec:eval_proto_linear} and \ref{supsec:eval_proto_finetune}.

\subsection{Effect of Cross-modal Knowledge Distillation} %

As discussed, the proposed \xkd is pretrained to solve masked data modelling and cross-modal knowledge distillation. Therefore, to obtain an accurate understanding of the impact of cross-modal knowledge distillation, we compare \xkd ($\mathbf{\loss_\mathrm{\bf xkd}}$) with respect to the following baselines: 
\begin{itemize}[noitemsep,nolistsep,leftmargin=30pt]
    \item masked video reconstruction ($\mathbf{\loss_\mathrm{\bf recon}^{v}}$)
    \item masked audio reconstruction ($\mathbf{\loss_\mathrm{\bf recon}^{a}}$)
    \item audio-video masked autoencoder ($\mathbf{\loss_\mathrm{\bf ae}}$).
\end{itemize}
We pretrain the above variants for the full training schedule of $800$ epochs and report linear evaluation results on the split-1 of downstream benchmarks.
 
\noindent\textbf{Visual representations.} 
The results presented in Table \ref{tab:ablation_vid} show that video representations significantly benefit from the cross-modal supervision obtained through knowledge distillation.
In comparison to $\loss_\mathrm{recon}^{v}$, $\loss_\mathrm{xkd}$ improves video action recognition by $8.6\%$, $8.2\%$, and $13.9\%$ on UCF101, HMDB51, and Kinetics-Sound, respectively. Furthermore, we challenge XKD on Kinetics400 in a linear evaluation setup which improves the top-1 accuracy by $15.7\%$. We interestingly notice that the joint audio-visual masked reconstruction ($\loss_\mathrm{ae}$) does not make considerable improvements, e.g., it improves top-1 accuracies by only $0.1\%-0.2\%$ on UCf101, HMDB51, and Kinetics-Sound.

\begin{table}[!h]
\fontsize{9pt}{10pt}\selectfont
\centering

\resizebox{0.99\columnwidth}{!}{%
    \begin{tabular}{lllll}
    \toprule
    \textbf{Loss} & \textbf{UCF101} & \textbf{HHDB51} & \textbf{Kin-Sound} & \textbf{Kinetics400} \\
    \midrule\midrule

    $\mathbf{\loss_\mathrm{\bf recon}^{v}}$ & 76.1 \decblue{8.6} & 51.1 \decblue{8.2} & 56.8 \decblue{13.9} & 30.7 \decblue{15.7}  \\ 
    $\mathbf{\loss_\mathrm{\bf ae}}$ & 76.3 \decblue{8.4} & 51.2 \decblue{8.1} & 56.9 \decblue{13.8} & 32.2 \decblue{14.2} \\
    $\mathbf{\loss_\mathrm{\bf xkd}}$
    & \textbf{84.7} & \textbf{59.3} & \textbf{70.7} & \textbf{46.4} \\    
    
    \bottomrule
    \end{tabular}%
}
\caption{Effect of \textbf{cross-modal knowledge distillation} in video action recognition. We report linear eval. top-1 acc. 
}
\label{tab:ablation_vid}
\vspace{10pt}

\centering
    \resizebox{0.55\linewidth}{!}{
        \begin{tabular}[c]{lllll}
        \toprule
        \textbf{Loss} & \textbf{FSD50K} & \textbf{ESC50} \\
        \midrule\midrule

        $\mathbf{\loss_\mathrm{\bf recon}^{a}}$ & 44.6 \decblue{1.2} & 90.0 \decblue{1.0}   \\ 
        $\mathbf{\loss_\mathrm{\bf ae}}$ & 44.3 \decblue{1.5} & 90.0 \decblue{1.0}  \\  
        $\mathbf{\loss_\mathrm{\bf xkd}}$ & \textbf{45.8} & \textbf{91.0} \\
        \bottomrule
        \end{tabular}%
    }
    \captionof{table}{Effect of \textbf{cross-modal knowledge distillation} in sound classification. We report mAP for FSD50K and top-1 acc. for ESC50.}
    \label{tab:ablation_aud}

\vspace{10pt}
\centering
\includegraphics[width=0.8\linewidth]{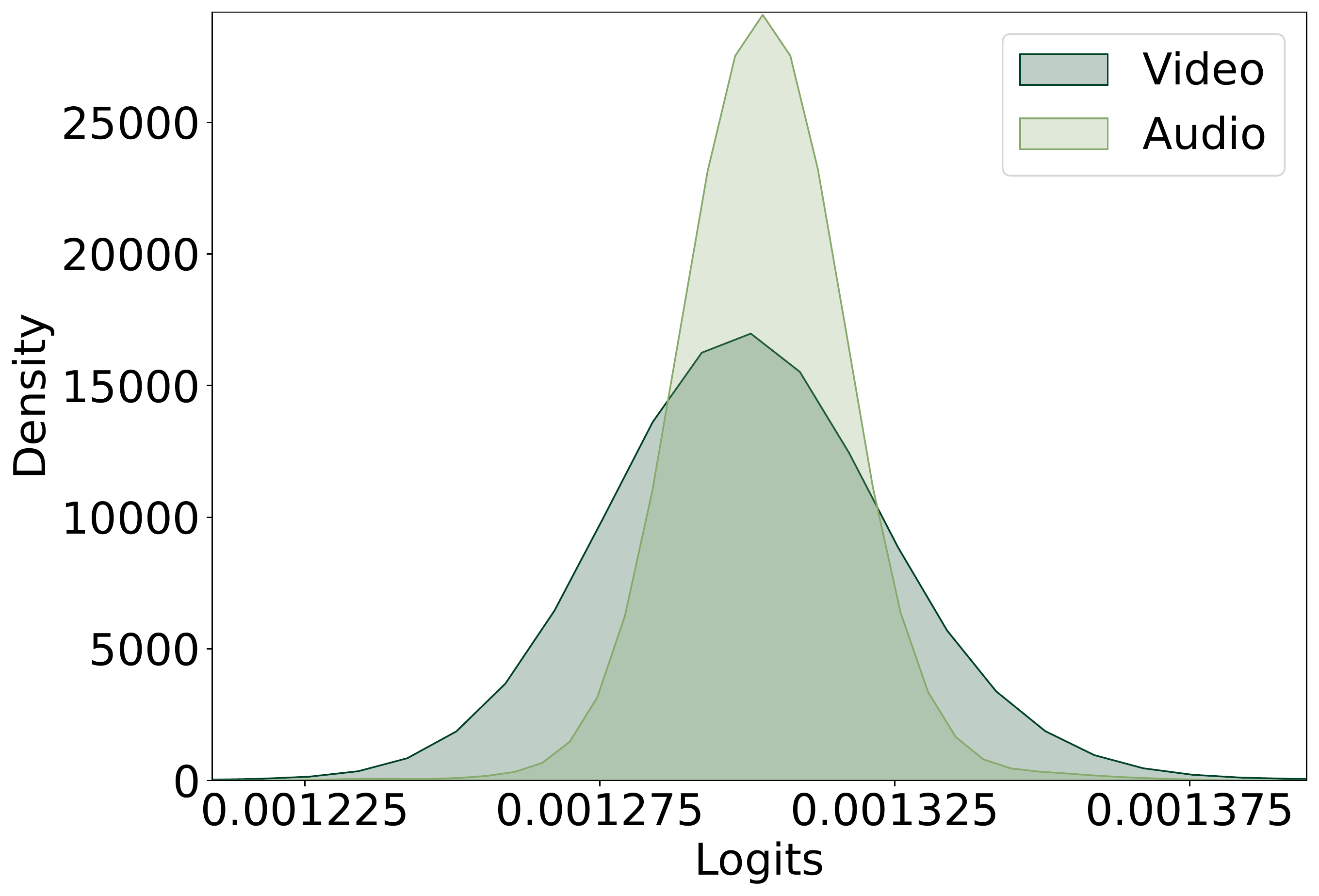}
\captionof{figure}{
    Visualizing the distribution of audio and visual representations, obtained from Kinetics400. It shows that audio has a sharper distribution compared to video. We find that a sharper distribution inherently provides better cross-modal supervision.
    }
    \label{fig:aud_vid_dist}
\end{table}

\noindent\textbf{Audio representations.} 
In Table \ref{tab:ablation_aud}, we notice that cross-modal knowledge distillation improves the performance in sound classification by $1.5\%$ and $1\%$ on FSD50K and ESC50 respectively. We note that the improvement is relatively less prominent compared to visual representations. Our thorough literature review in this regard reveals that a similar phenomenon has also been noticed amongst earlier works \cite{chen2021distilling,ren2021learning} that have attempted cross-modal knowledge distillation between audio and video in a semi-supervised setting. We conclude that while audio-to-video knowledge distillation is highly effective, video-to-audio provides a less substantial improvement. This is likely since a sharper distribution is preferred to provide supervision \cite{dino,mocov3}. As shown in Figure \ref{fig:aud_vid_dist}, the distribution of audio is sharper while the distribution of video is quite wider in nature.
We find that such constraints can be overcome to some extent by applying aggressive sharpening of visual representations. 
Please find additional discussions on temperature scheduling in the Suppl. Mat. Sec.  \ref{supsec:temperature}. 
Additionally, readers are redirected to the Suppl. Mat. Sec. \ref{supsec:ema} to \ref{supsec:mask_ratio} for studies on EMA schedule, local views, and mask ratio.

\begin{table}[t]
\fontsize{9pt}{10pt}\selectfont
\centering

\resizebox{0.99\columnwidth}{!}{%
    \begin{tabular}{lcccp{0.25\columnwidth}}
    \toprule
    \textbf{Loss} & \textbf{UCF} & \textbf{HMDB} & \textbf{Kinetics} & \multicolumn{1}{l}{\textbf{Remarks}} \\
    \midrule\midrule

    \multirow{2}{*}{$\mathbf{\loss_\mathrm{\bf ae}}$} & 
    \multirow{2}{*}{76.3} & 
    \multirow{2}{*}{51.2} & 
    \multirow{2}{*}{32.2} & {multimodal baseline.} \\ \midrule
    \multirow{3}{*}{${~+\!\loss_\mathrm{kd}}$} & 
    \multirow{3}{*}{76.3} & 
    \multirow{3}{*}{51.2} & 
    \multirow{3}{*}{32.2} & {without $\loss_\mathrm{da}$, knowledge distillation fails.}
    \\ \midrule
    \multirow{3}{*}{$~+\!\loss_\mathrm{da}$} & 
    \multirow{3}{*}{77.8} & 
    \multirow{3}{*}{51.4} & 
    \multirow{3}{*}{33.7} & {without $\loss_\mathrm{kd}$, $\loss_\mathrm{da}$ has marginal impact.} \\ \midrule

    \multirow{3}{*}{$~+\!\loss_\mathrm{kd}\!+\!\loss_\mathrm{da}$}
    & 
    \multirow{3}{*}{\textbf{84.7}} & 
    \multirow{3}{*}{\textbf{59.3}} & 
    \multirow{3}{*}{\textbf{46.4}} & {$\loss_\mathrm{da}+\loss_\mathrm{kd}$ improves the accuracy by $8\!-\!14\%$.} \\

    \bottomrule
    \end{tabular}%
}
\caption{Effect of \textbf{domain alignment}. $\loss_\mathrm{da}$ and $\loss_\mathrm{kd}$ are complementary to each other. While $\loss_\mathrm{da}$ and $\loss_\mathrm{kd}$ are not effective when applied separately, their combined optimization significantly improves the performance.
}
\label{tab:ablatida}
\end{table}

\subsection{Effect of Domain Alignment}

We conduct a thorough ablation study investigating the impact of domain alignment in our proposed framework, as presented in Table \ref{tab:ablatida}. First, without domain alignment, the model fails to perform cross-modal knowledge distillation due to domain discrepancy, and the model behaves as an audio-visual masked autoencoder. Second, quite expectedly, naively aligning the two domains without knowledge distillation has a minor impact. Last, our results exhibit that the proposed $\loss_{da}$ and $\loss_{kd}$ are complementary to each other, and their combined effect significantly improves the performance, e.g., by $8.1-14.2\%$ on UCF101, HMDB51, and Kinetics400. 
In addition to the absence of domain alignment, we identify two more factors that could cause training instability, discussed in the Suppl. Mat. Sec.  \ref{supsec:collapse} and \ref{supsec:projector}. Please see alternative design choices for domain alignment in the Suppl. Mat. Sec.  \ref{supsec:domain_alignment}.

\subsection{Effect of Feature Refinement ($\mathrm{\bf refine}$)}
To study the impact of $\mathrm{refine}$ in cross-modal knowledge distillation, we modify ${\loss_\mathrm{da}}$ in the final loss function (\Eqref{eq:final_loss}) as 
{$ \loss_\mathrm{mmd}(\theta_t^a(x_a^g), \theta_t^v(x_v^g)) + 
                \loss_\mathrm{mmd}(f_s^v, f_s^a)$}. 
The results presented in Table \ref{tab:alignment} demonstrate that $\mathrm{refine}$ improves downstream performance by $5.8\%$, $3.7\%$, and $3.0\%$ on HMDB51, UCF101, and ESC50 respectively. In Figure \ref{fig:refine}, we visualize the attention from the last layer with and without $\mathrm{refine}$, which further confirms the ability of $\mathrm{refine}$ in identifying the most transferable and key features for enhanced downstream performance. 
Please see alternative design choices of feature refinement Suppl. Mat. Sec. \ref{supsec:feature_refinement}. 

\begin{table}[!h]
\centering
\resizebox{0.52\linewidth}{!}{
    \begin{tabular}{lcc}
    \toprule
     & \textbf{without} & \textbf{with} \\ \midrule\midrule
     \textbf{HMDB51}  & {53.5} \decblue{5.8} & \textbf{59.3} \\
     \textbf{UCF101}  & {81.0} \decblue{3.7} & \textbf{84.7} \\
     \textbf{ESC50} & {88.0} \decblue{3.0} & \textbf{91.0} \\
    \bottomrule
    \end{tabular}
    }
    \caption{\textbf{Effect of} $\mathrm{\bf refine}$ is presented.
    $\mathrm{\bf refine}$ significantly improves downstream task performance on both audio and video tasks. 
    }
    \label{tab:alignment}
    
\vspace{10pt}
\centering
\setlength\tabcolsep{0pt}
\renewcommand{\arraystretch}{-5}
\resizebox{0.9\linewidth}{!}{
\begin{tabular}{lcc}
      &
     \bf \footnotesize without refine & \bf \footnotesize {with refine} 
     \\
     \rotatebox{90}{\tiny ~~~~~~~~~~Crying} 
     &
     \includegraphics[width=0.48\linewidth]{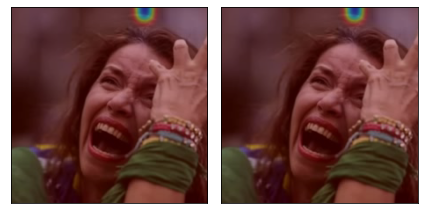}    &
     \includegraphics[width=0.48\linewidth]{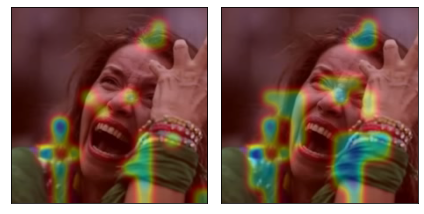}
     \\
     \rotatebox{90}{\tiny ~~~~~~~~~~Stretching} &
      \includegraphics[width=0.48\linewidth]{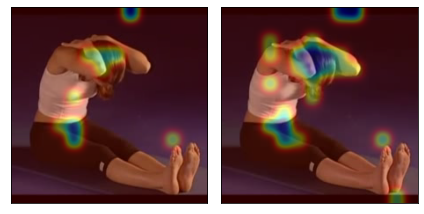}  &
     \includegraphics[width=0.48\linewidth]{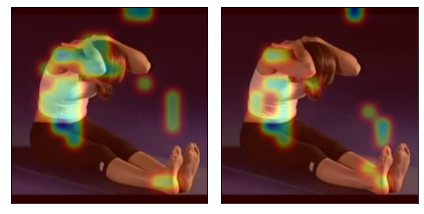}
     \\
    \end{tabular}
    }
    \captionof{figure}{Visualizing the \textbf{effect of} $\mathrm{\bf refine}$ in identifying the most transferable and key visual features. Please see more qualitative examples in the Suppl. Mat. Sec. \ref{supsec:qualitative_refinement}.
    }
    \label{fig:refine}

\end{table}

\begin{table}[]

\fontsize{9pt}{10pt}\selectfont
\centering
    \resizebox{0.7\linewidth}{!}{%
    \begin{tabular}{lccc}
    \toprule
         & \textbf{MATS} & \textbf{MAS} & \textbf{MS} \\ \midrule\midrule
         \textbf{HMDB51} & 53.5\decblue{5.8} & 55.2\decblue{4.1} & \textbf{59.3} \\ %
         \textbf{UCF101} & 80.3\decblue{4.4} & 81.5\decblue{3.2} & \textbf{84.7} \\ %
         \textbf{ESC50} & 90.3\decblue{0.7} & 89.5\decblue{1.5} & \textbf{91.0} \\ %
        \bottomrule
    \end{tabular}
    }
    \caption{Comparison between \textbf{modality-agnostic} (MATS, MAS) and \textbf{modality-specific} (MS) variants.}
    \label{tab:ma}
    
\vspace{10pt}
\fontsize{9pt}{10pt}\selectfont
\centering
    \resizebox{0.6\linewidth}{!}{%
    \begin{tabular}{lcc}
    \toprule
         & 
         \textbf{Student} & \textbf{Teacher} \\ \midrule\midrule
         \textbf{HMDB51} &
         57.8\decblue{1.5} & \textbf{59.3} \\ %
         \textbf{UCF101} & 
         84.4\decblue{0.3} & \textbf{84.7}  \\ %
         \textbf{ESC50} & 
         90.3\decblue{0.7} & \textbf{91.0} \\ %
        \bottomrule
    \end{tabular}
    }
\caption{Comparison between  \textbf{students} vs. \textbf{teachers}. %
}
\label{tab:teacher_student}

\end{table}

\subsection{Comparing Modality-Agnostic vs. -Specific}
In Table \ref{tab:ma}, we compare the performance of modality-agnostic (MA) variants with the modality-specific (MS) one. Amongst the MA variants, XKD-MATS works slightly better on ESC50, whereas, XKD-MAS shows better performance on both UCF101 and HMDB51. The results are promising, as these variants show minor performance drops (e.g., $0.7\%-4.1\%$) in comparison to the default XKD. Please note that such performance drops are expected as we keep the encoder size fixed for both MA and MS variants \cite{vatt}. To further elaborate, while we dedicate $2$ separate backbones of $87$M parameters to learn audio and visual representations for the MS variant, we use just one backbone of $87$M parameters for \textit{both} audio and visual modalities in the MA variants. Therefore, the network parameters or model weights become saturated relatively quickly, limiting their performance. A simple solution could be to use a larger backbone for MA variants,
which could be explored in future.

\subsection{Comparing Teacher vs. Student}
In Table \ref{tab:teacher_student}, we present the comparison between teachers and students. Our results exhibit that due to the slow weight update schedule using EMA (see \Eqref{eq:ema}), the teachers become slightly stronger learners compared to the students. We study the impact of different EMA schedules (presented in the Suppl. Mat. Sec.  \ref{supsec:ema}) 
and find that EMA coefficients as $0.997$ work best in our setup. As presented in Table \ref{tab:teacher_student}, the video teacher outperforms the video student by $0.3\%$ and $1.5\%$ on UCF101 and HMDB51. Next, the audio teacher outperforms the audio student by $0.7\%$ on ESC50. Please note that by default, we use the teachers to perform downstream tasks. 
We present a detailed comparison between the modality-specific and modality-agnostic variants using both teacher and student encoders in the Suppl. Mat. Sec.  \ref{supsec:xkd_variants}.

\subsection{Scalability}

\begin{figure}[]
    \centering
    \includegraphics[width=0.8\columnwidth]{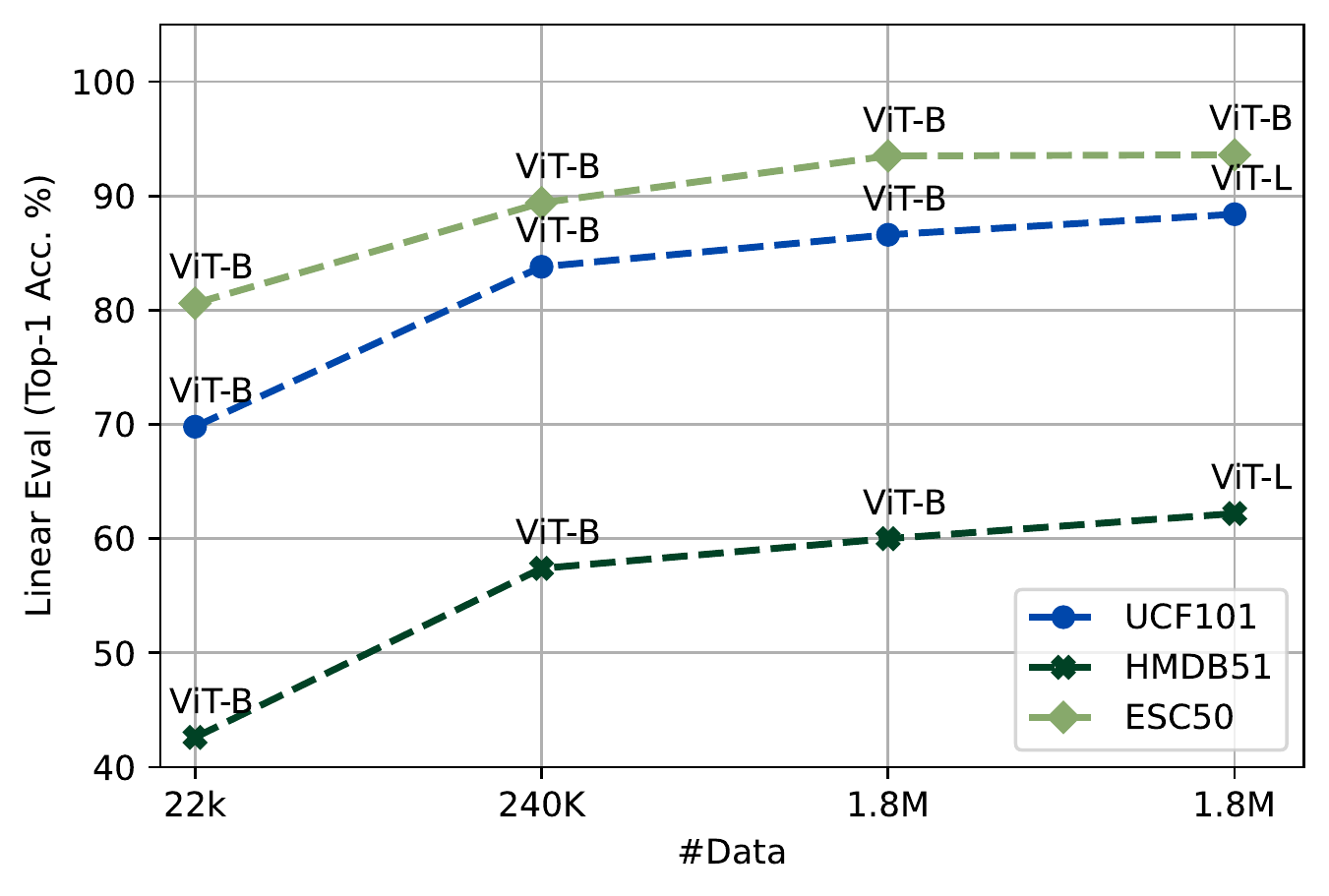}
    \caption{We study \textbf{scalability} both in terms of the dataset (x-axis) and network (ViT-B/ViT-L) size. Our results exhibit that XKD is scalable to both larger datasets and networks.}
    \label{fig:lt_scalability}
\end{figure}

We study the scalability of XKD on $3$ pretraining datasets of different sizes,  similar to \cite{crisscross,cmacc}, i.e., Kinetics-Sound ($22$K), Kinetics400 ($240$K), and AudioSet ($1.8$M). Additionally, we experiment with $2$ different sizes of ViT variants, i.e., ViT-B ($87$M) and ViT-L ($ 305$M). We try both ViT-B and ViT-L as the video backbone when pretrained on AudioSet.
We report the linear evaluation top-1 accuracy averaged over all the splits on UCF101, HMDB51, and ESC50. Figure \ref{fig:lt_scalability} shows that XKD continues to perform better as we increase the number of training samples and/or the size of the network.
Such scalability is a much-desired property, which shows XKD can likely be scaled on even larger datasets like HowTo100M \cite{howto100m} or larger networks like ViT-22B \cite{dehghani2023scaling}. 
We also study the effect of longer pretraining of XKD, presented in the Suppl. Mat. Sec.  \ref{supsec:epoch_acc}.

\begin{table}[!h]
\fontsize{9pt}{10pt}\selectfont
\centering
\setlength\tabcolsep{2pt}
\resizebox{\columnwidth}{!}{
\begin{tabular}{lccllcllclcc}
\toprule
\multirow{2}{*}{\bf Method} & \multirow{2}{*}{\bf Pretrain} & \multirow{2}{*}{\bf Mod.}  
&& \multicolumn{2}{c}{\textbf{UCF101}}
&& \multicolumn{2}{c}{\textbf{HMDB51}}
&& \multicolumn{2}{c}{\textbf{Kinetics400}}  \\ \cmidrule{5-6} \cmidrule{8-9} \cmidrule{11-12}
&  & 
&& \textbf{Lin.} & \textbf{FT.} 
&& \textbf{Lin.} & \textbf{FT.} 
&& \textbf{Lin.} & \textbf{FT.}  \\
\midrule\midrule

{AVSlowFast} \shortcite{xiao2020audiovisual} & {K400} & {VA} && 77.4 & 87.0  && 44.1 & 54.6  && - & -  \\ 
SeLaVi \shortcite{selavi} & K400 & VA && - & 83.1 && - & 47.1 && - & -  \\
XDC \shortcite{xdc} & K400 & VA  && - & 86.8 && - & 52.6 && - & -  \\
CMACC \shortcite{cmacc} & K400 & VA && - & 90.2 && - & 61.8 && - & -  \\
{AVID} \shortcite{avid} & {K400} & {VA} && {72.3}$^*$ & 87.5  && 41.4$^*$ & 60.8  && 44.5  & - \\ 
CMAC \shortcite{cmac} & K400 & VA  && - & 90.3 && - & 61.1 && - & -  \\
{GDT} \shortcite{gdt} & {K400} & {VA} && 70.1$^*$ & 90.9  && 38.5$^*$ & 62.3  && -  & - \\ 
STiCA \shortcite{stica} & K400 & VA  && - & 93.1 && - & 67.0 && - & -  \\
{CrissCross} \shortcite{crisscross} & {K400} & {VA} && \textbf{83.9} & 91.5  && {50.0} & 64.7  && 44.5  & - \\

\midrule
\textbf{XKD} & K400 & VA &&  {83.8} & \textbf{94.1}  && \textbf{57.4} & \textbf{69.0}  && \textbf{51.4} & \textbf{77.6}  \\ 
{\textbf{XKD-MAS}} & K400 & VA && 81.7 & 93.4  && 55.1 & 65.9  && 50.1 & 75.9  \\ 
{\textbf{XKD-MATS}} & K400 & VA && 80.1 & 93.1  && 53.1 & 65.7  && 48.8 & 75.7  \\ 
\midrule

{XDC} \shortcite{xdc} & {AS} & {VA} & & {85.3} & {93.0} & & {56.0} & {63.7} & & {-} & {-} \\
{MMV} \shortcite{mmv} & {AS} & {VA}   & & {83.9} & {91.5} & & {60.0} & {70.1} & & {-} & {-} \\ 
{CM-ACC} \shortcite{cmacc} & {AS} & {VA}  & & {-} & {93.5} & & {-} & {67.2} & & {-} & {-} \\
{BraVe$^{**}$} \shortcite{brave} & {AS} & {VA}  & & \textbf{90.0} & {93.6} && \textbf{63.6} & {70.8} && {-} & {-} \\
{AVID} \shortcite{avid} & {AS} & {VA} & & {-} & {91.5} & & {-} & {64.7} & & {48.9} & {-} \\
{CrissCross} \shortcite{crisscross} & {AS} & {VA} & & {87.7} & {92.4} & & {56.2} & {67.4} & & {50.1} & {-} \\
\midrule
\textbf{XKD} & AS & VA &&  {88.4} & \textbf{95.8}  && {62.2} & \textbf{75.7}  && \textbf{56.5} & \textbf{80.1}  \\ 

\midrule
\fadetext{VideoMAE} \shortcite{videomae} & \fadetext{U101/H51} & \fadetext{V}  & & \fadetext{-} & \fadetext{91.3} & & \fadetext{-} & \fadetext{62.6} & & \fadetext{-} & \fadetext{-} \\
\fadetext{BEVT} \shortcite{bevt} & \fadetext{K400} & \fadetext{V}  & & \fadetext{-} & \fadetext{-} & & \fadetext{-} & \fadetext{-} & & \fadetext{-} & \fadetext{76.2} \\
\fadetext{VideoMAE} \shortcite{videomae} & \fadetext{K400} & \fadetext{V}  & & \fadetext{-} & \fadetext{-} & & \fadetext{-} & \fadetext{-} & & \fadetext{-} & \fadetext{79.0} \\
\fadetext{VideoMAE}$^{**}$ \shortcite{videomae} & \fadetext{K400} & \fadetext{V}  & & \fadetext{80.0}$^*$ & \fadetext{96.1} & & \fadetext{54.3}$^*$ & \fadetext{73.3} & & \fadetext{-} & \fadetext{80.0} \\ %
\fadetext{CPD} \shortcite{cpd} & \fadetext{K400} & \fadetext{VT}  && \fadetext{-} & \fadetext{90.5} && \fadetext{-} & \fadetext{63.6} && \fadetext{-} & \fadetext{-} \\
\fadetext{CoCLR} \shortcite{coclr} & \fadetext{K400} & \fadetext{VF} && \fadetext{74.5} & - && \fadetext{46.1} & -   && - & -   \\
\fadetext{BraVe}$^{**}$ \shortcite{brave} & \fadetext{AS} & \fadetext{VFA}  & & \fadetext{93.2} & \fadetext{96.9} & & \fadetext{69.9} & \fadetext{79.4} & & \fadetext{-} & \fadetext{-} \\
\fadetext{MIL-NCE} \shortcite{milnce} & \fadetext{HT} & \fadetext{VT}  & & \fadetext{-} & \fadetext{91.3} & & \fadetext{-} & \fadetext{61.0} & & \fadetext{-} & \fadetext{-} \\ 
\fadetext{VATT} \shortcite{vatt} & \fadetext{AS+HT} & \fadetext{VAT} && \fadetext{89.6} & \fadetext{-} && \fadetext{65.2} & \fadetext{-} && \fadetext{-} & \fadetext{81.1} \\
\fadetext{VATT-MA} \shortcite{vatt} & \fadetext{AS+HT} & \fadetext{VAT} && \fadetext{84.4} & \fadetext{-} && \fadetext{63.1} & \fadetext{-} && \fadetext{-} & \fadetext{79.9} \\
\fadetext{ELo} \shortcite{elo} & \fadetext{YT} & \fadetext{VFA}  & & \fadetext{-} & \fadetext{93.8} & & \fadetext{-} & \fadetext{67.4} & & \fadetext{-} & \fadetext{-} \\
\toprule
\multicolumn{12}{p{1.25\linewidth}}{{%
Mod.: Modality, 
V: Video, A: Audio, T: Text, F: Flow. 
$^*$computed by us using official checkpoints. 
$^{**}$industry-level computation, e.g., VideoMAE uses $64$ vs. ours $8$ GPUs, BraVe pretrains with very high temporal resolutions (128 frames) compared to others (8-32). 
More comparisons with VideoMAE are in the Suppl. Mat. Sec.  \ref{supsec:comp_videomae}.
}}
\end{tabular}%
}
\caption{Comparison on \textbf{video action recognition}. XKD outperforms or achieves competitive performance compared to state-of-the-art methods.
}
\label{tab:sota_action}
\end{table}

\subsection{Video Action Recognition}
Following the standard practice in \cite{xdc,crisscross,cmacc}, we compare XKD with other leading \textit{audio-visual self-supervised} frameworks in Table \ref{tab:sota_action} in both linear evaluation (\textbf{Lin.}) and finetuning (\textbf{FT.}) on UCF101, HMDB51, and Kinetics400. 
We note a large variability in the experiment setups amongst the prior works, however, they are included for a more inclusive and thorough comparison.
We report top-1 accuracy averaged over all the splits for both UCF101 and HMDB51.

The results presented in Table \ref{tab:sota_action} show that XKD outperforms or achieves competitive performance amongst the leading audio-visual self-supervised methods. For example, XKD pretrained on Kinetics400 outperforms AVID and CrissCross in linear evaluation on Kinetics400, by a very large margin. Moreover, XKD outperforms powerful state-of-the-art models like STiCA, BraVe, and CrissCross among several others when finetuned on UCF101. Additionally, XKD outperforms prior works that are pretrained with massive datasets and tri-modalities like Elo pretrained with video, audio, and optical flow from YouTube8M (YT) \cite{yt8m}, compared to XKD pretrained with audio and video from AudioSet 2M.
Our modality-agnostic variants also achieve encouraging results, e.g., a minor performance drop is noticed between XKD-MAS and XKD, e.g., $0.7\%$ on UCF101 and $1.7\%$ on Kinetics400, when finetuned.

\begin{table}[!t]
\fontsize{9pt}{10pt}\selectfont
    \centering
    
    \resizebox{0.99\linewidth}{!}{%
    \begin{tabular}{lcccclcc}
    \toprule
    \multirow{2}{*}{\textbf{Method}}
    & \multirow{2}{*}{\textbf{Pretrain}} 
    & \multirow{2}{*}{\textbf{Mod.}} 
    & \multicolumn{2}{c}{\textbf{ESC50}} && \multicolumn{2}{c}{\textbf{FSD50K}}  \\ \cmidrule{4-5} \cmidrule{7-8}
    &&& \textbf{Lin.} & \textbf{FT.} && \textbf{Lin.} & \textbf{FT.} \\ \midrule\midrule
    XDC \shortcite{xdc} & K400 & VA & 78.0 & - && - & - \\ 
    AVID \shortcite{avid} & K400 & VA & 79.1 & - && - & - \\ 
    STiCA \shortcite{stica} & K400 & VA & 81.1 & - && - & - \\ 
    CrissCross \shortcite{crisscross} & K400 & VA & 86.8 & - && - & - \\ \midrule
    \textbf{XKD} & K400 & VA & \textbf{89.4} & \textbf{93.6} && \textbf{45.8} & \textbf{54.1} \\ 
    \textbf{XKD-MAS} & K400 & VA & 87.3 & 92.7 && 43.0 & 52.3 \\
    \textbf{XKD-MATS} & K400 & VA & 88.7 & 92.9 && 43.4 & 53.8 \\
    \midrule
    {AVTS} \shortcite{avts} & {AS} & {VA} & {$80.6$} & - && {-} & {-} \\
    {AVID} \shortcite{avid} & {AS} & {VA} & {$89.2$} & - && {-} & {-} \\
    {GDT} \shortcite{gdt} & {AS} & {VA} & {$88.5$} & - && {-} & {-} \\ 
    CrissCross \shortcite{crisscross} & {AS} & {VA} & {$90.5$} & - && {-} & {-} \\ 
    \midrule
    \textbf{XKD} & AS & VA & \textbf{93.6} & \textbf{96.5} && \textbf{51.5} & \textbf{58.5} \\ 
    \midrule
    \fadetext{AudioMAE} \shortcite{audiomae} & \fadetext{AS} & \fadetext{A} & - & \fadetext{93.6} && \fadetext{-} & \fadetext{-}  \\  
    \fadetext{MaskSpec} \shortcite{chong2022masked} & \fadetext{AS} & \fadetext{A} & - & \fadetext{89.6} && \fadetext{-}& \fadetext{-}  \\
    \fadetext{MAE-AST} \shortcite{baade2022mae} & \fadetext{AS} & \fadetext{A} & - & \fadetext{90.0} && \fadetext{-}& \fadetext{-}   \\ 
    \fadetext{BYOL-A} \shortcite{byol_audio} & \fadetext{AS} & \fadetext{A} & \fadetext{-} & - && \fadetext{$44.8$}  \\ 
    \fadetext{Aud. T-former} \shortcite{verma2021audio} & \fadetext{AS} & \fadetext{A} & \fadetext{-} & \fadetext{-} && \fadetext{-} & \fadetext{53.7}  \\ 
    \fadetext{SS-AST} \shortcite{ast} & \fadetext{AS} & \fadetext{A} & - & \fadetext{88.8} && \fadetext{-} & \fadetext{-}  \\   
    \fadetext{PSLA} \shortcite{psla} & \fadetext{AS} & \fadetext{A} & \fadetext{-} & \fadetext{-} && \fadetext{-} & \fadetext{55.8}  \\ 
    \fadetext{VATT} \shortcite{vatt} & \fadetext{AS+HT} & \fadetext{VAT} & \fadetext{$84.7$} & \fadetext{-} && \fadetext{-} & \fadetext{-} \\
    \fadetext{VATT-MA} \shortcite{vatt} & \fadetext{AS+HT} & \fadetext{VAT} & \fadetext{$81.2$} & \fadetext{-} && \fadetext{-} & \fadetext{-} \\
    \fadetext{PaSST(SL)} \shortcite{passt} & \fadetext{AS+IN} & \fadetext{AI} & \fadetext{-} & \fadetext{95.5} && \fadetext{-} & \fadetext{58.4} \\
    \toprule
    \multicolumn{8}{p{1.05\linewidth}}{{\footnotesize 
    Here, I: Image, IN: ImageNet \cite{imagenet}.}}
    \end{tabular}%
    }
    \caption{{Comparison on \textbf{sound classification}.} XKD outperforms the prior state-of-the-art methods. 
    }
    \label{tab:sota_audio}
\end{table}

\subsection{Sound Classification}

In Table \ref{tab:sota_audio}, we compare the performance of our proposed method using linear evaluation (\textbf{Lin.}) and finetuning (\textbf{FT.}) on sound classification using $2$ popular audio benchmarks ESC50 and FSD50K. Following \cite{fsd50k,esc}, we report top-1 accuracy averaged over all the splits on ESC50 and mean average precision on FSD50K. XKD outperforms prior works like XDC, AVID, and CrissCross on ESC50 in both linear evaluation and finetuning. Moreover, XKD and its MA variants outperform VATT and VATT-MA on ESC50 by $4.7\%$ and $7.5\%$, even though VATT is pretrained with $136$M videos, compared to XKD which is only trained on $240$K samples. Additionally, when evaluated on FSD50K, XKD outperforms BYOL-A, AudioTransformer, and PSLA among others. Lastly, XKD shows state-of-the-art performance on ESC50, achieving top-1 finetuned accuracy of $96.5\%$ when pretrained with AudioSet.

\begin{table}[]
    \centering
    \fontsize{9pt}{10pt}\selectfont
    \begin{tabular}{lc}
    \toprule
     \textbf{Method} & \textbf{Audio + Video}  \\ \midrule\midrule
       CrissCross &  {66.7} \\
       $\loss_{ae}$ (AV-MAE) & {75.7} \\ \midrule
       \textbf{XKD} & \textbf{81.2} \\ 
       \textbf{XKD-MAS} & 78.8  \\ 
       \textbf{XKD-MATS} & 78.3 \\ 
    \bottomrule
    \end{tabular}%
    \caption{
    \textbf{Multimodal} action classification by late fusion on Kinetics-Sound.
    }
    \label{tab:fusion}
\end{table}

\subsection{Multimodal Fusion}%
Following \cite{crisscross}, we evaluate XKD in multimodal action classification using Kinetics-Sound. We extract fixed audio and visual embeddings from the pretrained encoders and concatenate them together (i.e., late fusion), followed by a linear SVM classifier is trained and top-1 accuracy is reported.
The results presented in Table \ref{tab:fusion} show that XKD outperforms CrissCross by $14.5\%$ and baseline audio-visual masked autoencoder ($\mathbf{\loss_{\bf ae}}$) by $5.5\%$. 

\subsection{In-painting.}
\begin{figure}[!t]

\centering
\setlength\tabcolsep{0pt}
\renewcommand{\arraystretch}{-40} %
\resizebox{0.9\linewidth}{!}{%
\begin{tabular}{lccc}
& \tiny \bf Original & \tiny \bf Masked & \tiny \bf Reconstructed \\
\rotatebox{90}{\tiny ~~~~~~~~~Video} &
\includegraphics[width=0.3\linewidth]{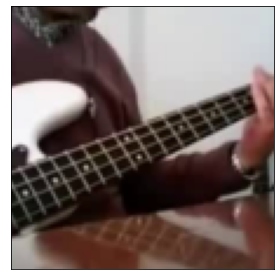}  &
\includegraphics[width=0.3\linewidth]{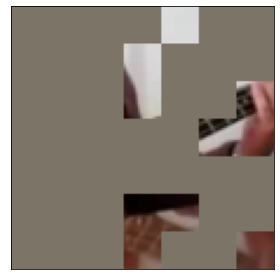}  &
\includegraphics[width=0.3\linewidth]{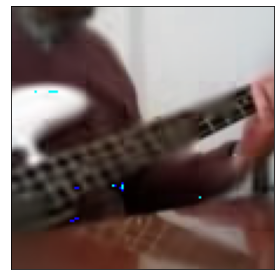} \\
\rotatebox{90}{\tiny ~~~~~~Audio} &
\includegraphics[width=0.3\linewidth]{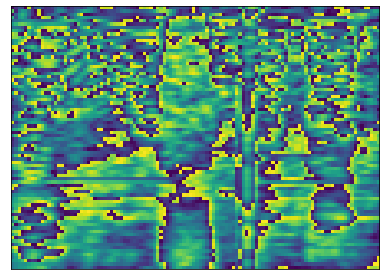}  &
\includegraphics[width=0.3\linewidth]{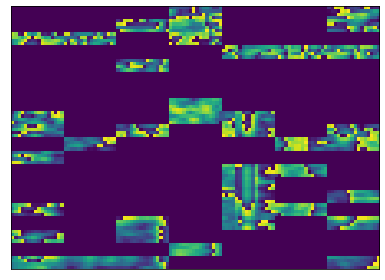}  &
\includegraphics[width=0.3\linewidth]{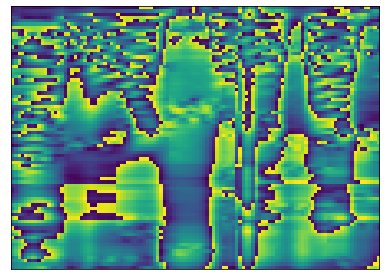} 
\end{tabular}

}
\captionof{figure}{\textbf{Reconstruction} examples from highly masked inputs, video and audio mask ratios are $80\%$ and $70\%$.%
}
\label{fig:inpaint}

\end{figure}

We present reconstruction examples in Figure \ref{fig:inpaint}, which shows that XKD retains its reconstruction ability even when a very high masking ratio is applied, for both audio and video modalities. 
This makes XKD also suitable for in-painting tasks. 
More examples are in the Suppl. Mat. Sec.  \ref{supsec:qualitative_inpainting}.

\section{Summary} \label{sec:conclusion}

In this work, we propose XKD, a novel self-supervised framework to improve video representation learning using cross-modal knowledge distillation. To effectively transfer knowledge between audio and video modalities, XKD aligns the two domains by identifying the most transferable features and minimizing the domain gaps. Our study shows that cross-modal knowledge distillation significantly improves video representations on a variety of benchmarks. Additionally, to develop a general network with the ability to process different modalities, we introduce modality-agnostic variants of XKD which show promising results in handling both audio and video using the same backbone. We believe that our approach can be further expanded to perform cross-modal knowledge distillation between other modalities as well (e.g., vision and language), which could be investigated in the future.

\section*{Acknowledgments}
We are grateful to the Bank of Montreal and Mitacs for funding this research. We are thankful to SciNet HPC Consortium for helping with the computation resources.

\small{
\bibliography{refs}
}
\clearpage
\appendix
\begin{center}
    \Large
    \textbf{Supplementary Material}
\end{center}

\setcounter{table}{0}
\setcounter{figure}{0}
\setcounter{equation}{0}
\renewcommand{\theequation}{S\arabic{equation}}
\renewcommand{\thetable}{S\arabic{table}}
\renewcommand\thefigure{S\arabic{figure}}

\noindent The organization of the supplementary material is as follows:
\begin{itemize}[]
    \item \ref{supsec:implementation}: Implementation Details
    \begin{itemize}
        \item \ref{supsec:dataset} Datasets
        \item \ref{supsec:augmentation} Augmentation
        \item \ref{supsec:pretraining} Pretraining
        \item \ref{supsec:modality_agnostic} Modality-agnostic setups
        \item \ref{supsec:eval_proto_linear} Linear evaluation
        \item \ref{supsec:eval_proto_finetune} Finetuning
    \end{itemize}
    \item \ref{supsec:addl_exp}: Additional Experiments and Results
    \begin{itemize}
        \item \ref{supsec:temperature} Effect of temperature schedules
        \item \ref{supsec:ema} Effect of different EMA schedules
        \item \ref{supsec:local_view} Effect of local views
        \item \ref{supsec:mask_ratio} Effect of different mask ratio
        \item \ref{supsec:collapse} Avoiding collapse and training instability
        \item \ref{supsec:projector} Design choices for projector head
        \item \ref{supsec:domain_alignment} Design choices for domain alignment
        \item \ref{supsec:feature_refinement} Design choices for feature refinement
        \item \ref{supsec:xkd_variants} Comparing different variants of XKD 
        \item \ref{supsec:epoch_acc} Training schedule
        \item \ref{supsec:comp_videomae} Detailed comparison with VideoMAE
    \end{itemize}
    \item \ref{supsec:qualitative}: Qualitative Analysis
    \begin{itemize}
        \item \ref{supsec:qualitative_refinement} Effect of feature refinement
        \item \ref{supsec:qualitative_inpainting} XKD reconstruction results
    \end{itemize}
\end{itemize}

\newpage

\section{Implementation Details}\label{supsec:implementation}

\subsection{Datasets} \label{supsec:dataset}
The details of the datasets are presented.

\noindent\textbf{Kinetics400.}
Kinetics400\cite{kinetics400} is a large-scale action recognition audio-visual dataset consisting of $400$ action classes. It has a total of $240$K video clips with an average duration of $10$ seconds. Following \cite{crisscross,xdc,avid,videomae,videomae2}, we use the Kinetics400 for both pretraining and downstream evaluation. 

\noindent\textbf{AudioSet.} AudioSet \cite{audioset} is a very large-scale audio-visual dataset comprised of $1.8$M videos and collected from YouTube. The samples from AudioSet are of average duration of $10$ seconds and spread over $632$ audio events. We use AudioSet to conduct experiments on large-scale pretraining. Please note that none of the labels are used in pretraining.

\noindent\textbf{UCF101.} UCF101\cite{ucf101} is a popular action recognition dataset used for downstream evaluation. It consists of a total of $13$K clips of an average duration of $7$ seconds and spread over $101$ action classes. We use UCF101 for downstream evaluation on video action recognition.

\noindent\textbf{HMDB51.} HMDB51 \cite{hmdb} is one of earlier video action recognition datasets. It contains $7$K video clips distributed over $51$ action classes. HMDB51 is used for downstream evaluation on video action recognition.

\noindent\textbf{Kinetics-Sound.} Kinetics-Sound\cite{l3-kineticssound} is originally a subset of Kinetics400, which has a total of $22$K video clips consisting of $32$ action classes. Following \cite{crisscross}, we use Kinetics-Sound for multimodal action classification by late fusion. We particularly choose Kinetics-Sound for multimodal evaluation as it consists of hand-picked samples from Kinetics400 which are prominently manifested both audibly and visually. In addition to downstream evaluation, Kinetics-Sound is also used to conduct experiments on small-scale pretraining.

\noindent\textbf{ESC50.} ESC50 is a popular sound classification dataset that consists of a total of $2$K samples of $50$ environmental sound classes, with an average duration of $5$ seconds. Following \cite{vatt,crisscross,avid,brave}, we use ESC50 for downstream evaluation of audio representations.

\noindent\textbf{FSD50K.} FSD50K is a popular audio dataset of human-labeled sound events, which consists of $50$K audio samples distributed over $200$ classes. Moreover, FSD50K is a multi-labeled dataset, where samples are between $0.3$ to $30$ seconds. Following \cite{byol_audio} We use FSD50K for downstream evaluation of audio representations.

\subsection{Augmentation} \label{supsec:augmentation}
We adopt standard augmentation strategies \cite{crisscross,simclr,byol_audio} to create the global and local views from the raw audio and visual streams. Particularly, we apply {Multi-scale crop}, {Color Jitter}, and {Random Horizontal Flip} to create global visual views. To create local views, we apply Gray Scale and Gaussian Blur, in addition to the augmentations applied on the global views with minor changes in the augmentation parameters. For example, different from global visual views, we apply aggressive cropping to create local visual views. 
The details of the augmentation parameters are presented in Table \ref{stab:params_vid_aug}. 
Next, {Volume Jitter} is applied to the audio stream to create the global audio views. We apply Random Crop in addition to the Volume Jitter to create the local audio views. The details of the audio augmentation parameters are presented in Table \ref{stab:params_aud_aug}. 

\begin{table}[!h]
    \fontsize{9pt}{10pt}\selectfont
    \setlength\tabcolsep{2pt}
        \centering

        \resizebox{\linewidth}{!}{%
        \begin{tabular}{llccc}
        \toprule
        \textbf{Augmentation} & \textbf{Param} & \textbf{Global View} & \textbf{Local View} & \textbf{Linear Eval.}\\ \midrule\midrule
        Multi-Scale Crop & Crop Scale & [0.2, 1] & [0.08, 0.4] & [0.08, 1] \\ \midrule
        Horizontal Flip  & Probability & 0.5 & 0.5 & 0.5 \\ \midrule
        \multirow{4}{*}{Color Jitter} & Brightness & 0.4 & 0.4 & 0.4\\ 
        & Contrast & 0.4 & 0.4 & 0.4\\
        & Saturation & 0.4 & 0.4 & 0.4\\
        & Hue & 0.2 & 0.2 & 0.2 \\ \midrule
        Gray Scale & Probability & n/a & 0.2 & 0.2\\ \midrule
        Gaussian Blur & Probability & n/a & 0.5 & n/a \\ 
        
        \bottomrule
        \end{tabular}%
        }
        \caption{\textbf{Visual augmentation} parameters for pretraining and linear evaluation are presented. n/a: not applied.
        }
        \label{stab:params_vid_aug}
\end{table}

\begin{table}[!h]
    \fontsize{9pt}{10pt}\selectfont
    \setlength\tabcolsep{2pt}
        \centering

        \resizebox{\linewidth}{!}{%
        \begin{tabular}{llccc}
        \toprule
        \textbf{Augmentation} & \textbf{Param} & \textbf{Global View} & \textbf{Local View} & \textbf{Linear Eval.} \\ \midrule\midrule
        Volume Jitter & Scale & $\pm$ 0.1 & $\pm$ 0.2 & $\pm$ 0.1 \\ 
        \multirow{2}{*}{Random Crop} & Range & n/a & [0.6, 1.5] & n/a\\ 
        & Crop Scale & n/a & [1, 1.5] & n/a \\ 
        
        \bottomrule
        \end{tabular}%
        }
        \caption{\textbf{Audio augmentation} parameters for pretraining and linear evaluation are presented. n/a: not applied.}
        \label{stab:params_aud_aug}
\end{table}

\subsection{Pretraining} \label{supsec:pretraining}
We train \xkd using an AdamW \cite{adamw1,adamw2} optimizer and warm-up cosine learning rate scheduler. %
During ablation studies, the models are trained for $400$ epochs using Kinetics400. We use ViT-B \cite{vit} as the backbone for both audio and visual modalities. Following \cite{mae}, we use a shallow decoder for reconstruction and adopt a similar projector head to the one proposed in \cite{dino}. Please see the additional architecture details in Table \ref{stab:params_hyper_net}. We perform distributed training with a batch size of $256$ using $8$ NVIDIA V100 $32$ GB GPUs in parallel. As mentioned in Table \ref{stab:params_hyper_net}, we use visual frames with resolutions of $96^2$ and $112^2$, such lower resolutions allow us to train the proposed \xkd with a limited computation setup. Additionally, we perform mixed-precision \cite{amp} training to save the computation overhead. 
In our setup, it takes approximately $10$ days to train the \xkd for $800$ epochs with Kinetics400. 
We present the additional details for the training parameters in Table \ref{stab:params_hyper}. Please note that the parameters are tuned based on Kinetics400 and the same parameters are used for both AudioSet and Kinetics-Sound. Due to resource constraints, we do not further tune the parameters for individual datasets. As the AudioSet is very large, we pretrain XKD for $400$ epochs, whereas, the models are pretrained for $800$ epochs for Kinetics400 and Kinetics-Sound.

\begin{table}[!h]
    \fontsize{9pt}{10pt}\selectfont
        \centering
        \begin{tabular}{lr}
        \toprule
        \textbf{Name} & \textbf{Value} \\ \midrule\midrule
        Video (global) & $32\times 112^2$ \\
        Video (local) & $8\times 96^2$ \\
        Video Patch & $4\times 16^2$  \\ \midrule
        Audio (global) & $80\times 448$ \\
        Audio (local) & $80\times 112$ \\
        Audio Patch & $4\times 16$  \\ \midrule
        \multirow{4}{*}{Backbone} & ViT-Base (87M) \\
         & embed dim:768 \\
         & depth:12 \\
         & num heads:12 \\ \midrule
        \multirow{4}{*}{Decoder} & Transformer \\
         & embed dim:384 \\
         & depth:4 \\ 
         & num heads:12 \\   \midrule
        \multirow{5}{*}{Projector} & MLP \\
        & out dim: 8192 \\
        & bottleneck dim: 256 \\
        & hidden dim: 2048 \\
        & num layers: 3 \\
         \bottomrule
        \end{tabular}%
        \caption{\textbf{Architecture} details of XKD.}
        \label{stab:params_hyper_net}
\end{table}

\begin{table}[!h]
    \fontsize{9pt}{10pt}\selectfont
        \centering
        \begin{tabular}{lr}
        \toprule
        \textbf{Name} & \textbf{Value} \\ \midrule\midrule
        Epochs & $800$ \\
        Batch Size & $256$ \\
        Optimizer & \texttt{AdamW} \\
        Optimizer Betas & $[0.9, 0.95]$ \\
        LR Scheduler & \texttt{Cosine} \\
        LR Warm-up Epochs & $30$ \\
        LR Warm-up & $0$ \\
        LR Base & $0.0001$ \\
        LR Final & $0$ \\
        Weight Decay & $0.3$ \\
        EMA Scheduler & \texttt{Cosine} \\
        EMA Base & $0.997$ \\
        EMA Final & $1$ \\
        Video Mask Ratio & $0.85\%$ \\
        Audio Mask Ratio & $0.80\%$ \\
        Video Student Temp. & $0.1$ \\
        Audio Student Temp. & $0.1$ \\
        Video Teacher Temp. & $0.1$ \\
        Audio Teacher Temp. & $0.07$ \\
        Center Momentum & $0.9$ \\

        \bottomrule
        \end{tabular}%
        \caption{XKD \textbf{pretraining} setups for Kinetics400.}
        \label{stab:params_hyper}
\end{table}

\subsection{Modality-agnostic setups} \label{supsec:modality_agnostic}
In Table \ref{tab:ma_details}, we summarize the detailed network setups of the modality-agnostic variants. The student encoders are shared in XKD-MAS, while both the teachers' and students' backbones are shared in XKD-MATS. Please note that in none of the setups, the input projection layer and the decoders are shared.

\begin{table}[!h]
\fontsize{9pt}{10pt}\selectfont
\setlength\tabcolsep{2pt}
    \centering

    \resizebox{\linewidth}{!}{%
    \begin{tabular}{lcccc}
    \toprule
        \textbf{Variant} & \textbf{Input Projection} & \textbf{Teacher} & \textbf{Student} & \textbf{Decoder}  \\ \midrule \midrule
        XKD-MAS & \textit{NS} & \textit{NS} & \textit{S} & \textit{NS} \\
        XKD-MATS & \textit{NS} & \textit{S} & \textit{S} & \textit{NS} \\ \midrule
        XKD & \textit{NS} & \textit{NS} & \textit{NS} & \textit{NS} \\
        \bottomrule
    \end{tabular}
    }
    \caption{Summary of parameter sharing in \textbf{modality-agnostic} setups. Here,
    \textit{S} and \textit{NS} refer to shared and not-shared parameters between audio and visual modalities respectively.}
    \label{tab:ma_details}
\end{table}

\subsection{Linear evaluation}\label{supsec:eval_proto_linear}

\noindent\textbf{UCF101 and HMDB51.}
To perform linear evaluation, we feed $32$ frames with a spatial resolution of
$112^2$ to the pretrained visual encoder. 
We extract the fixed features as the mean of all the patches from the last layer of the encoder. 
We randomly select $25$ samples per clip during training, while during testing, we uniformly fetch $3$ clips per sample to extract frozen features. Next, the features are used for action classification using a linear SVM kernel. We sweep a range of cost values $\{0.00001, 0.0001, 0.001, 0.01, 0.1, 1\}$ and report the best top-1 accuracy. We present the augmentation parameters applied on the training set in Table \ref{stab:params_vid_aug}. Please note that none of the augmentations are applied to the test set. We then tune the models using split 1 of the datasets and report the top-1 accuracies averaged over all the splits.

\noindent\textbf{ESC50.}
We feed $4$ seconds of audio input to the pretrained audio encoder and extract approximately $10$-epochs worth of fixed features. Similar to UCF101 and HMDB51 we extract the features as the mean of all the patches from the last layer of the encoder. During training, we apply Volume Jitter as augmentation and no augmentation is applied at test time. Similar to the setup of UCF101 and HMDB51, we sweep the same range of cost values to find the best model. We report the final top-1 accuracy averaged over all the splits. 

\noindent\textbf{FSD50K.}
To perform the downstream evaluation on FSD50K we follow the same setup as ESC50, with the exception of using a linear fully-connected layer instead of SVM. During training, we randomly extract $10$ clips per sample and feed them to the pretrained audio encoder. Moreover, during test, we uniformly select $8$ clips per sample and report the mean average precision (mAP) 
\cite{fsd50k}. The extracted features are used to train a fully-connected layer using an AdamW \cite{adamw1,adamw2} optimizer for $50$ epochs at a fixed learning rate of $0.001$. Moreover, we apply weight decay of $1e-5$ and dropout of $0.3$ to prevent overfitting. We use a batch size of $256$ and train on a single RTX6000 $24$ GB NIVIDIA GPU.

\noindent\textbf{Kinetics-Sound and Kinetics400.}
We mainly use Kinetics-Sound for 
multimodal evaluation using late fusion. 
To perform late fusion, we simply concatenate the corresponding audio and visual features, and the joint representations are then directly used to perform action classification using a linear kernel SVM. Finally, we report the top-1 accuracy.

We follow similar setups to those used for UCF101 and HMDB51 to perform linear evaluation on Kinetics400, with the exception of randomly sampling $3$ clips per video during training and testing to extract the features (we do not sample more clips during training due to memory constraints). Lastly, we report top-1 accuracy in action classification. In addition to the SVM (used in ablation study and analysis), we also perform FC-tuning (used to compare with the prior works in Table \ref{tab:sota_action}) %
on Kinetics400, considering Kinetics400 is sufficiently large compared to UCF101 and HMDB51. Similar to the setup mentioned earlier, we extract fixed features, followed by a linear fully-connected layer is trained to perform action recognition. We use an AdamW \cite{adamw1,adamw2} optimizer with a Cosine scheduler having a base learning rate of $0.01$ and batch size of $512$ to train for $20$ epochs.

\subsection{Finetuning}\label{supsec:eval_proto_finetune}

\noindent\textbf{Kinetics400, UCF101, and HMDB51.}
We use the pretrained visual encoder and add a fully-connected layer to finetune on Kinetics400, UCF101, and HMDB51. Different from our pretraining setup, we use a spatial resolution of $224^2$ in finetuning. During training, we apply Random Augmentation \cite{randaug} in addition to Multi-scale Crop and Random Horizontal Flip on the visual data, similar to \cite{videomae,videomae2,mae}. We use an AdamW \cite{adamw1,adamw2} optimizer to train the network for $100$ epochs using a cosine learning rate scheduler. In addition to the cosine learning rate scheduler, we employ layer-wise-layer-decay \cite{beit} which further stabilizes the training. We perform distributed training with a batch size of $96$ using $16$ NVIDIA V100 $32$ GB GPUs in parallel. Additionally, we employ MixUp \cite{mixup}, CutMix \cite{cutmix}, and label smoothing \cite{label-smoothing}, which boosts the finetuning performance. 
We present the details of the finetuning parameters in Table \ref{stab:params_hyper_finetune}. During training, we randomly sample $1$ clip per sample, while during testing, $3$ clips are uniformly selected per sample. We report the sample level prediction (top-1 and top-5 accuracy) averaged over all the clips. 

\begin{table}[!h]
    \fontsize{9pt}{10pt}\selectfont
        \centering
        \resizebox{0.99\linewidth}{!}{
        \begin{tabular}{lccc}
        \toprule
        \textbf{Name} & \textbf{Kinetics400}  & \textbf{UCF101}  & \textbf{HMDB51} \\ \midrule\midrule
        Epochs & \multicolumn{3}{c}{$100$} \\
        Batch Size & \multicolumn{3}{c}{$96$} \\
        Optimizer & \multicolumn{3}{c}{\texttt{AdamW}}  \\
        Optimizer Betas & \multicolumn{3}{c}{$[0.9, 0.999]$}  \\
        LR Scheduler & \multicolumn{3}{c}{\texttt{Cosine}}  \\
        LR Warm-up Epochs & $30$ & $5$ & $20$ \\
        LR Warm-up & \multicolumn{3}{c}{$0$}  \\
        LR Base & $0.0005$ & $0.0005$ & $0.0001$ \\
        LR Final & \multicolumn{3}{c}{$1.0e^{-06}$}  \\
        Weight Decay & \multicolumn{3}{c}{$0.05$} \\
        RandAug & \multicolumn{3}{c}{(9, .05)} \\
        Label Smoothing & \multicolumn{3}{c}{$0.1$} \\
        MixUp & $0.6$ & $0.6$ & $1.0$ \\
        CutMix & $0.5$ & $0.5$ & $1.0$  \\
        Drop Path & \multicolumn{3}{c}{$0$} \\
        Dropout & \multicolumn{3}{c}{$0.5$} \\
        Layer-wise LR Decay & $0.45$ & $0.45$ & $0.55$ \\
        
        \bottomrule
        \end{tabular}%
        }
        \caption{\textbf{Video finetuning} parameters.}
        \label{stab:params_hyper_finetune}
\end{table}

\noindent\textbf{ESC50 and FSD50K.}
We use the pretrained audio encoder and add a linear layer on top of it to finetune on ESC50 and FSD50K. During training, we randomly sample $1$ audio segment of $4$ seconds for both datasets. However, in the case of validation, we uniformly sample $3$ and $8$ clips per sample for ESC50 and FSD50K respectively. The number of clips per sample during testing is chosen based on the average length of the samples to ensure the full length of the audio signal is captured. Following \cite{crisscross}, we apply strong augmentations including Volume Jitter, Time-frequency Mask, Timewarp, and Random Crop. We use an AdamW optimizer with a batch size of $64$ to train the network for $100$ epochs. We use standard categorical cross-entropy error for ESC50, and binary cross-entropy error for FSD50K, as FSD50K is a multi-label multi-class classification dataset. We report top-1 accuracy for ESC50 and mAP for FSD50K. The networks are trained on 4 NVIDIA V100 32 GB GPUs. We list the additional hyperparameters in Table \ref{stab:params_hyper_finetune_audio}.

\begin{table}[!h]
    \fontsize{9pt}{10pt}\selectfont
        \centering
        \resizebox{0.75\linewidth}{!}{
        \begin{tabular}{lcc}
        \toprule
        \textbf{Name} & \textbf{ESC50}  & \textbf{FSD50K}  \\ \midrule\midrule
        Epochs & 100 & 30 \\
        Batch Size & \multicolumn{2}{c}{$64$} \\
        Optimizer & \multicolumn{2}{c}{\texttt{AdamW}}  \\
        Optimizer Betas & \multicolumn{2}{c}{$[0.9, 0.999]$}  \\
        LR Scheduler & cosine & fixed  \\
        LR Warm-up Epochs & $10$ & - \\
        LR Base & $0.0001$ & $0.0001$ \\
        LR Final & 0 & - \\
        Weight Decay & 0.005 & $1.0e-05$ \\
        Early stop & no & yes \\
        Volume Jitter & \multicolumn{2}{c}{$0.2$} \\
        Time-mask & \multicolumn{2}{c}{$[0, 20]$} \\
        Frequency-mask & \multicolumn{2}{c}{$[0, 10]$}  \\
        Num of masks & \multicolumn{2}{c}{$2$} \\
        Timewarp window & \multicolumn{2}{c}{$20$}  \\
        Drop Path & \multicolumn{2}{c}{$0$} \\
        Dropout & $0.5$ & $0.3$ \\
        Layer-wise LR Decay & $0.65$ & $0.65$ \\
        
        \bottomrule
        \end{tabular}%
        }
        \caption{\textbf{Audio finetuning} parameters.}
        \label{stab:params_hyper_finetune_audio}
\end{table}

\section{Additional Experiments and Results} \label{supsec:addl_exp}

Following, we present an in-depth study analysing the key concepts of our proposed framework. Our thorough analysis includes the effect of temperature and EMA schedules; and the impact of local views and masking ratios; design choices for domain alignment, feature refinement, and projector head; and the performance of different XKD variants; among others. The models are pretrained for $400$ epochs on Kinetics400 and we report linear evaluation top-1 accuracy using the split-1 of UCF101, HMDB51, and ESC50 unless stated otherwise.

\begin{table}[]
\centering
\setlength\tabcolsep{1.25pt}
    \resizebox{\linewidth}{!}{%
    \begin{tabular}{l>{\color{black}}c>{\color{black}}c>{\color{black}}c>{\color{black}}c>{\color{black}}c>{\color{black}}c>{\color{black}}c>{\color{black}}c>{\color{black}}c>{\color{black}}c}
    \toprule
          \textbf{Audio} 
          & $0.04$ & $0.06$ & $0.07$ & $0.08$ 
          & $0.07$ & $0.07$ & $0.07$ & $0.07$ 
          & $[0.04,0.06]$ & $[0.04,0.06]$ \\ \midrule
          \textbf{Video} & $0.04$ & $0.06$ & $0.07$ & $0.08$ & $0.09$ & $0.10$ & $0.11$ & $0.12$ &
            $[0.04,0.06]$ & $[0.09,0.11]$\\ 
            \midrule\midrule
        HMDB51 & 55.3 & \underline{57.5} &  \underline{57.5} & 55.6 & 54.4 & 55.9 & 54.5 & 53.8 &\textbf{58.5} & 52.7 \\
         UCF101 & 80.0 & 81.0 &  \underline{82.0} & 81.6 & 81.0 & 81.9 & 79.5 & 78.5 &\textbf{82.6} & 79.0 \\
         ESC50 & 85.8 & 85.3 &  84.3 & 85.8 & 87.0 & \textbf{89.0} & \textbf{89.0} & 87.3 & \underline{88.0} & 86.5 \\
        \bottomrule

    \end{tabular}%
    }
    \caption{\textbf{Effect of temperature schedules.} We highlight the best with \textbf{bold} and the second best with \underline{underline}. We find that a temperature schedule of $[0.4, 0.6]$ (increasing $0.4$ to $0.6$ using a Cosine scheduler) achieves an overall stable performance. Interestingly, excessively increasing the video temperature to $0.10$ or $0.11$ (i.e., sharpening the video distribution) improves the performance of audio downstream tasks, however, it hurts the performance of visual tasks. On the other hand, a fixed temperature of $0.6$ or $0.7$ shows stable performance in video tasks, but it hurts audio performance. By default, we use $[0.4, 0.6]$ in all the experiments.
    }
    \label{tab:temp}
        
\end{table}

\subsection{Effect of temperature schedules} \label{supsec:temperature}
We study the effect of a wide range of temperature settings ($\tau$ in Equation \ref{eq:softmax_temp}) %
in our proposed framework. The results presented in Table \ref{tab:temp} show that a relatively lower temperature works well on the audio teacher ($0.04$ to $0.07$), whereas a high temperature ($0.1$ to $0.11$) is required for the video teacher. Please note that the students' temperatures are kept fixed at $0.1$. To find the best performance, we sweep a range of combinations and notice that for both audio and video teachers, increasing the temperatures from $0.04$ to $0.06$ works fairly well for both modalities. Additionally, we notice that setting fixed temperatures of $0.07$ and $0.1$ for audio and video teachers respectively, works slightly better on ESC50; however, a performance drop is noticed for video, specifically on HMDB51. We conjecture that carrying more information and thus the wider distribution of the video stream (see Figure \ref{fig:aud_vid_dist} in the main paper) %
requires a higher temperature in order to further sharpen the distribution. Based on the results in Table \ref{tab:temp}, we use a scheduler to increase the temperature from $0.04$ to $0.06$ for both audio and video teachers throughout our experiments, unless mentioned otherwise.

\begin{figure}[!h]
    \centering
    \includegraphics[width=0.75\linewidth]{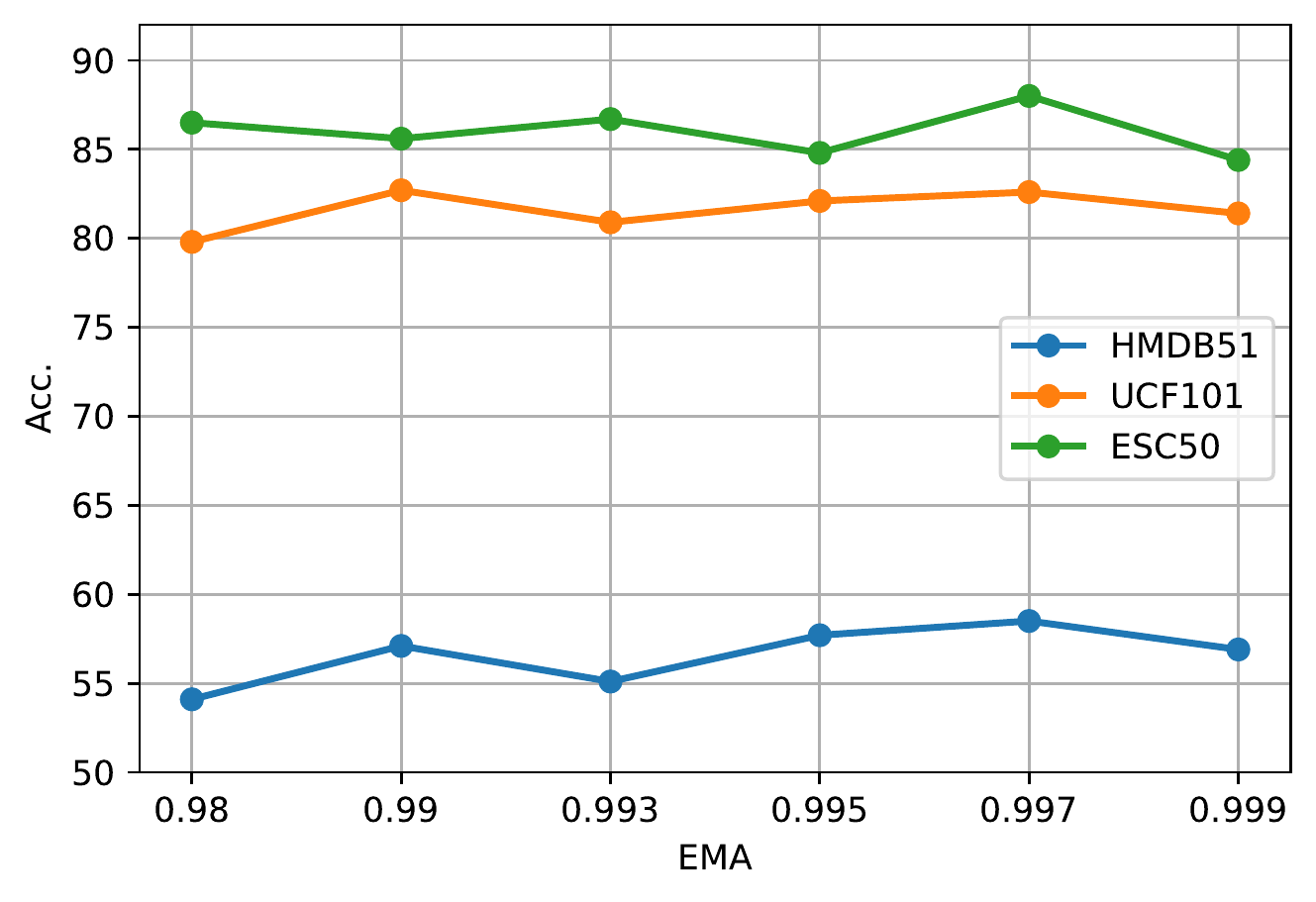}
    \caption{\textbf{Effect of EMA schedules.} EMA of $0.997$ achieves the most stable performance in both audio (ESC50) and visual (HMDB51 and UCF101) tasks.}
    \label{fig:ema}
\end{figure}

\subsection{Effect of different EMA schedules}\label{supsec:ema}
We update the weights of the teachers using EMA as described in Equation \ref{eq:ema}. %
In particular, we set the values of EMA coefficients using a cosine scheduler with a final value of $1$. This means that we update the teachers more frequently at the beginning of training, and slow down their weight update towards the end of the training. This strategy results in a very smooth update of the networks' parameters and results in stable performance \cite{mean_teacher}. To find the optimal setup, we vary a wide range of EMA coefficients between $0.98$ to $0.999$. The results presented in Figure \ref{fig:ema} show that the EMA coefficient of $0.997$ works well on both audio and video-related tasks.

\subsection{Effect of local views}\label{supsec:local_view}

\noindent\textbf{Temporal window size.}
We study the effect of the size of temporal windows used to create the local views for training the student networks. For the video stream, we create the local views by taking spatial crops of $96\!\times\!96$ pixels and vary the temporal windows in the range of $\{1, 2, 4\}$ seconds. Similarly for the audio stream, we explore temporal windows in the range of $\{1, 2\}$ seconds. 
Our results presented in Table \ref{tab:local_views_temp} reveal some interesting findings. First, local views comprised of audio and visual sequences of just $1$ second work the best on all three benchmarks. Interestingly, we notice that the downstream audio classification performance drops significantly when XKD uses long visual local views (see column $4-1$ in Table \ref{tab:local_views_temp}), while the downstream video classification performance remains more or less unchanged. On the other hand, we notice considerable performance drops on downstream video classification when using large audio segments to create the local views (see columns $2-2$ and $1-2$ in Table \ref{tab:local_views_temp}).
These observations indicate that local views comprised of relatively shorter temporal windows are most effective when learning from the cross-modal teachers, which in turn results in improved downstream classification performance.

\begin{table}[!h]
\fontsize{9pt}{10pt}\selectfont
\setlength\tabcolsep{6pt}
\centering

\begin{tabular}{lccccc}
    \toprule
         \textbf{Vid-Aud (sec.)} & $\mathbf{1\!-\!1}$ & $\mathbf{2\!-\!2}$ & $\mathbf{2\!-\!1}$ & $\mathbf{1\!-\!2}$ & $\mathbf{4\!-\!1}$ \\ \midrule\midrule
         HMDB51 & \textbf{58.5} & \underline{55.2} & 57.1 & 55.6 & 58.4 \\ %
         UCF101 & \textbf{82.6} & 81.1 & 82.3 & \underline{81.0} & \textbf{82.6} \\ %
         ESC50 & \textbf{88.0} & 86.8 & 87.3 & 86.8 & \underline{82.3} \\
        \bottomrule

\end{tabular}%
\caption{\textbf{Effect of different temporal window sizes for local views.} Here \textbf{bold} shows the best and \underline{underline} shows the worst results.
Our results indicate that local views comprised of relatively
shorter temporal windows are most effective to learn
from the cross-modal teachers. 
}
\label{tab:local_views_temp}
\end{table}

\noindent\textbf{Number of local views.}
We further investigate the effect of varying the number of local views. This experiment shows that single local views for both audio and visual modalities work fairly well. We also find that adding more visual local views (e.g., $2$ or $3$) when using a single audio local view improves the performance on HMDB51 and UCF101, as shown in columns $2-1$ and $3-1$ in Table \ref{tab:local_views_num}. Additionally, we notice that adding more audio local views worsens the model's performance in video classification (see columns $1\!-\!2$ and $2\!-\!2$ in Table \ref{tab:local_views_num}). Lastly, we notice considerable improvements in sound classification when using $2$ local views for both audio and video. 

\begin{table}[!h]
\fontsize{9pt}{10pt}\selectfont
\setlength\tabcolsep{6pt}
\centering

\begin{tabular}{lccccc}
    \toprule
         \textbf{Vid-Aud (no.)} & $\mathbf{1\!-\!1}$ & $\mathbf{1\!-\!2}$ & $\mathbf{2\!-\!1}$ & $\mathbf{2\!-\!2}$ & $\mathbf{3\!-\!1}$\\ \midrule\midrule
         HMDB51 & 58.5 & 56.3 & \textbf{58.8} & \underline{56.1} & 57.5 \\
         UCF101 & {82.6} & 79.5 & 82.2 & \underline{79.0} & \textbf{82.9} \\
         ESC50 & {88.0} & \underline{85.8} & 87.0 & \textbf{90.8} & 88.8 \\ %
        \bottomrule

\end{tabular}%
\caption{\textbf{Effect of additional local views.} We highlight the best results in \textbf{bold} and the worst with \underline{underline}. We find that XKD learns very strong representations just by using a single local view. Additionally, we encounter a few setups which show superior results on different downstream benchmarks. However, we do not see one single variant that performs the best in all setups.
}
\label{tab:local_views_num}
\end{table}

\subsection{Effect of different mask ratio}\label{supsec:mask_ratio}
\begin{figure}[!h]
    \centering
    \includegraphics[width=\linewidth]{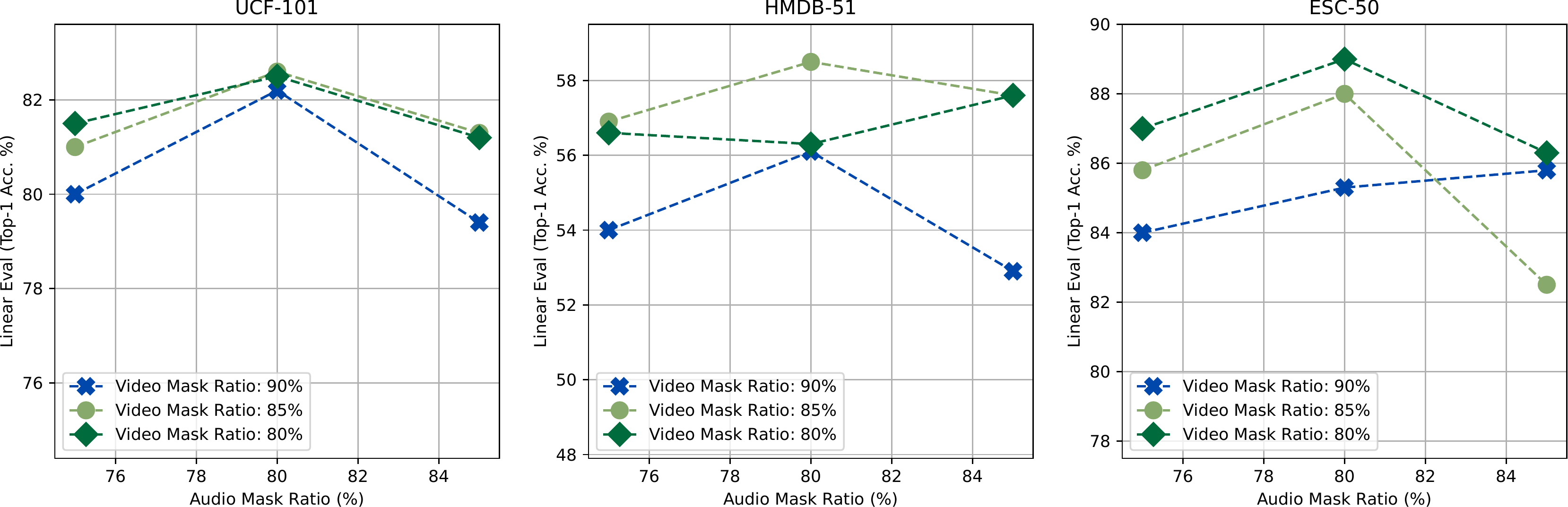}
    \caption{\textbf{Effect of different masking ratios.} We find that a video mask ratio of $85\%$ works best on both UCF101 and HMDB51. Additionally, an audio mask ratio of $80\%$ achieves superior results on ESC50.
    }
    \label{fig:ablation_masking}
\end{figure}

We study the effect of different amounts of masking ratios in our proposed framework. Specifically, we test different combinations of audio-masking ratio vs. video-masking ratio to test if there is an internal dependency that might result in optimal performance for both modalities. We sweep a wide ranges of mask ratios, particularly $\{0.80, 0.85, 0.90\}$ for video masking and $\{0.75, 0.80, 0.85\}$ for audio masking. The results presented in Figure \ref{fig:ablation_masking} show that a video mask ratio of $85\%$ and an audio mask ratio of $80\%$ work well on both UCF101 and HMDB51. However, we notice that a slightly lower video mask ratio (e.g., 80\%) improves audio performance. 
We conjecture that this setup benefits from the availability of more visual patches which provides better supervision through cross-modal knowledge distillation. 
Please note that unless stated otherwise, we use a visual mask ratio of $85\%$ and an audio mask ratio of $80\%$.

\begin{figure}[!h]
    \centering
    \includegraphics[width=0.99\linewidth]{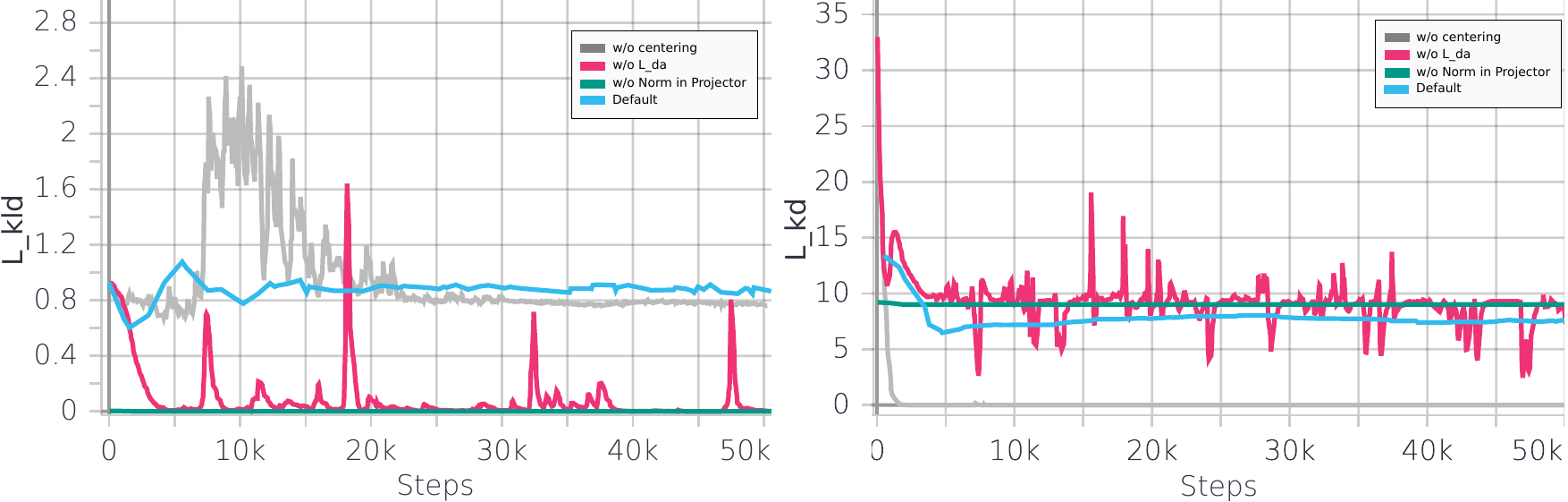}
    \caption{Visualizing \textbf{pretraining instability} and \textbf{collapse} in cross-modal knowledge distillation.  
    }
    \label{fig:collapse}
\end{figure}

\subsection{Avoiding collapse and training instability}\label{supsec:collapse}

\begin{figure}[]
    \centering
    \includegraphics[width=\linewidth]{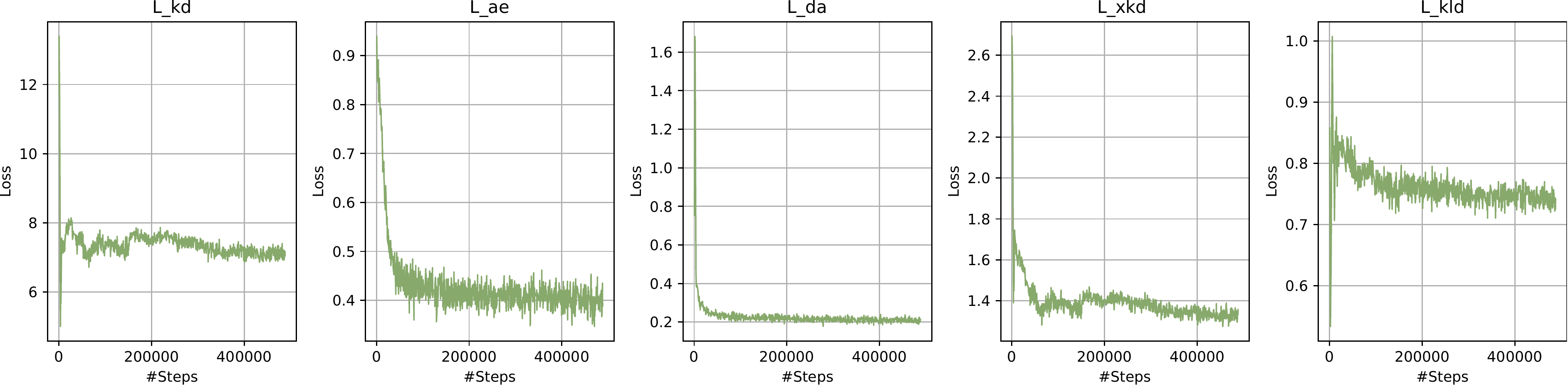}
    \caption{\textbf{Stable} training curves are presented.
    }
    \label{fig:loss}
\end{figure}

Here we summarize the setups that cause a collapse and instability in cross-modal knowledge distillation. We identify three such instances, (\textit{i}) without normalization layer in the projectors, (\textit{ii}) without $\loss_{da}$, and (\textit{iii}) without normalizing the teacher's representations based on the current batch statistics, i.e., Centering \cite{dino}. To identify collapse, we mainly track (not optimize) the Kullback–Leibler divergence between teacher and students ($\loss_{kld}$) 
and Knowledge Distillation ($\loss_{kd}$) losses, as shown in Figure \ref{fig:collapse}. A collapse can be identified if either $\loss_{kd}$ or $\loss_{kld}$ is zero. First, when no normalization layer is added in the projector head, the $\loss_{kld}$ becomes zero, which indicates that the models output a constant vector. Second, without $\loss_{da}$, the $\loss_{kld}$ also reaches zero (with minor spikes), which indicates training collapse. 
Third, we notice that without the Centering, $\loss_{kld}$ does not become zero, while the $\loss_{kd}$ reaches zero, which also indicates training collapse. 
In addition to the ablation studies presented in Tables \ref{tab:ablation_vid} and \ref{tab:ablation_aud} (in the main paper), 
we attempt to train XKD without the masked reconstruction loss (i.e., setting $\lambda_{ae}$ to 0 in Equation \ref{eq:final_loss}), %
and quite expectedly we face training collapse. This is due to the fact that
in order to perform effective cross-modal knowledge distillation, the model first needs to learn meaningful modality-specific representations.  
To provide more insights into the training process, we present the default training curves in Figure \ref{fig:loss}.

\subsection{Design choices for projector head}\label{supsec:projector}
The configuration of the projector heads plays an important role in the performance of our method. We experiment with a variety of different setups including normalization, output dimension, and the number of layers. 

\begin{figure}[!h]
    \centering
    \includegraphics[width=0.75\linewidth]{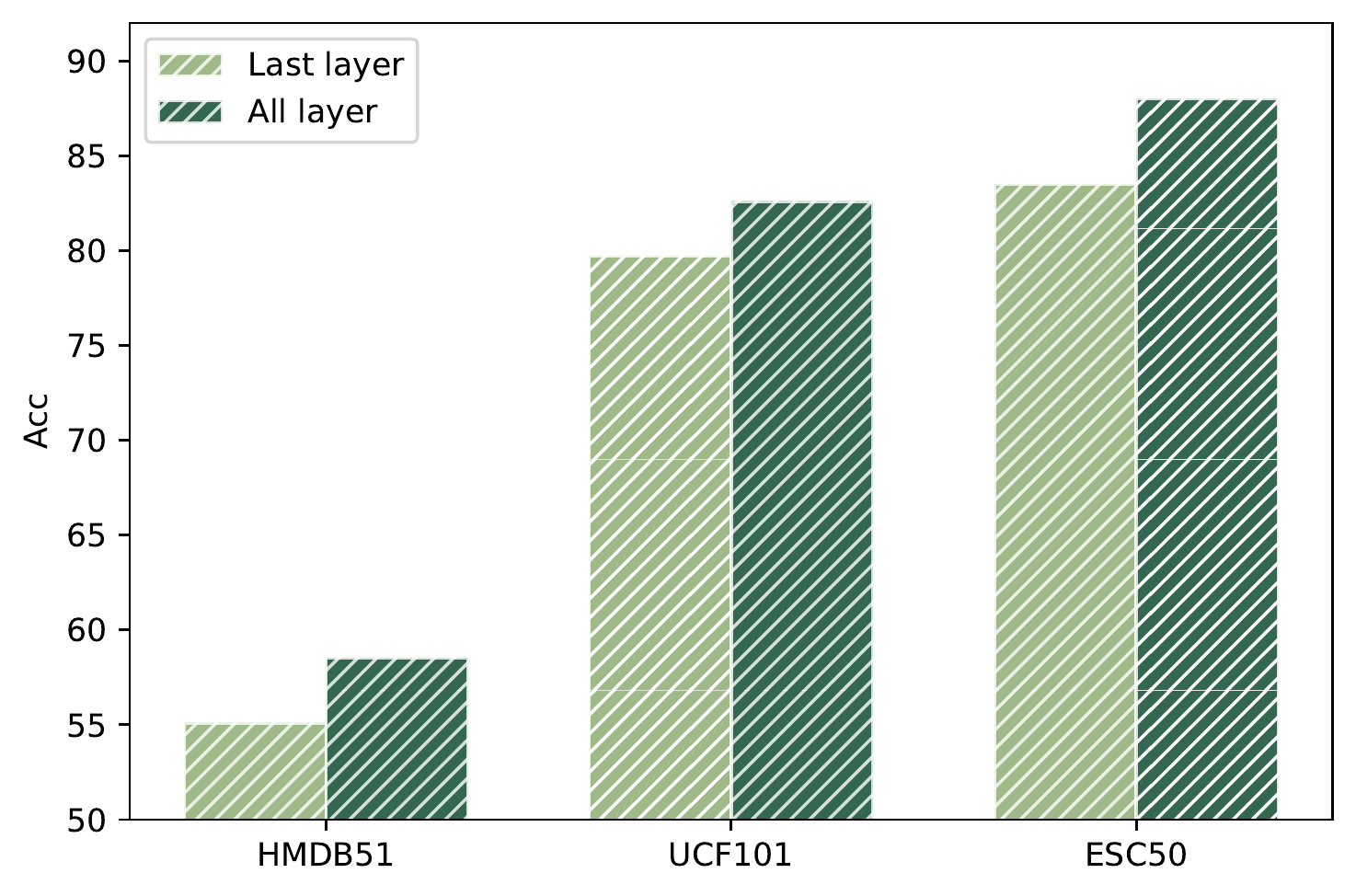}
    \caption{\textbf{Effect of normalization layers in the projector head.} We find that applying normalization to all the layers of the projector head improves the performance across all the downstream tasks.}
    \label{fig:projector}
\end{figure}

\noindent\textbf{Effect of normalization.}
{We investigate three setups,} (\textit{a}) no normalization, (\textit{b}) normalization on the last layer, and (\textit{c}) normalization on every layer. Our study shows that the cross-modal knowledge distillation fails in the absence of the normalization layer (please see additional discussions in \ref{supsec:collapse}). We then observe that adding normalization on the last layer prevents the model from collapsing while adding normalization on every layer shows significant improvements ($2.9\%$ to $4.5\%$)  as presented in Figure \ref{fig:projector}.

\begin{table}[!h]
\fontsize{9pt}{10pt}\selectfont
    \centering

    \begin{tabular}{lccc}
    \toprule
         \textbf{No. of layers} & \bf 2 & \bf  3 & \bf  4  \\ \midrule\midrule
         HMDB51 & 55.4 & \textbf{58.5} & 55.7 \\
         UCF101 & 81.0 & \textbf{82.6} & 81.0  \\
         ESC50  & 83.4 & \textbf{88.0} & 84.6  \\
    \bottomrule
    \end{tabular}
    \caption{\textbf{Effect of different number of projector layers.} We notice an MLP projector head with $3$ layers achieves the best performance.}
    \label{tab:proj_layers}
\end{table}

\begin{table}[!h]
\fontsize{9pt}{10pt}\selectfont
    \centering

        \begin{tabular}{lcccc}
        \toprule
             \textbf{Output dim.} & \bf  4096 & \bf  8192 & \bf  16384 & \bf  32768 \\ \midrule\midrule
             HMDB51 & 53.2 & \textbf{58.5} & 57.0 & 54.4 \\
             UCF101 & 78.1 & \textbf{82.6} & 81.5 & 81.7 \\
             ESC50 & 87.5 & \textbf{88.0} & 84.1 & 83.8 \\
        \bottomrule
        \end{tabular}
        \caption{\textbf{Effect of different projector output dimensions.} We notice that the output dimensions of $8192$ achieve the best performance. Please note that in all the setup the hidden layers consist of fixed dimensions of $2048$.}
        \label{tab:proj_11_dim}
\end{table}

\noindent\textbf{Effect of the number of layers and output dimension.}
We further explore different setups for the projector head. First, we present the effect of varying the number of layers and output dimensions in Tables \ref{tab:proj_layers} and \ref{tab:proj_11_dim}. The results show that a projector head consisting of $3$ layers performs most steadily on all the datasets. We also find that an output dimension of $8192$ shows stable performance on all the downstream benchmarks.

\begin{table}[h]
    \centering
    \resizebox{0.7\columnwidth}{!}{
    \begin{tabular}{lccc}
    \toprule
     & $\mathbf{\loss_\mathrm{da}^1}$ & $\mathbf{\loss_\mathrm{da}^2}$ &
      $\mathbf{\loss_\mathrm{da}}$ (\textbf{default}) \\ \midrule\midrule
     HMDB51   & 56.1 & 55.4 & \textbf{58.5} \\
     UCF101   & 81.1 & 82.2 & \textbf{82.6} \\
     ESC50  & 87.3 & 87.5 &  \textbf{88.0} \\
    \bottomrule
    \end{tabular}
    }
\caption{
\textbf{Exploring the design choices for domain alignment.} 
We find that just minimizing the MMD loss between the teachers and between the students compared to the other variants works best for effective cross-modal knowledge distillation.
}
\label{tab:alignment_supp}
\end{table}

\subsection{Design choices for domain alignment}\label{supsec:domain_alignment}
We conduct a thorough study exploring the optimal setup for domain alignment ($\loss_\mathrm{da}$). Recall, $\loss_\mathrm{da}$ is defined as:
\begin{equation*}
\begin{split}
    \loss_\mathrm{da} = \loss_\mathrm{mmd}(f_s^v, f_s^a) + \space \loss_\mathrm{mmd}(f_t^v, f_t^a) 
\end{split}
\end{equation*}
We explore the following setups.
First, 
instead of minimizing the MMD loss between teachers $(f_t^a, f_t^v)$ and between students $(f_s^a, f_s^v)$ as done in $\loss_\mathrm{da}$, we  minimize the distance between the teachers and students as:
\begin{equation}
\begin{split}
    \loss_\mathrm{da}^1\!=\!\loss_\mathrm{mmd}(f_t^a, f_s^v)\!+\!\loss_\mathrm{mmd}(f_t^v,f_s^a).
\end{split}
\end{equation}
Next, we study the effect of optimizing  $\loss_\mathrm{da}^1$ in addition to $\loss_\mathrm{da}$, expressed as: 
\begin{equation}
    \loss_\mathrm{da}^2 = \loss_\mathrm{da} + \loss_\mathrm{da}^1.
\end{equation}
The results presented in Table \ref{tab:alignment_supp} indicate drops in performance when using $\loss_\mathrm{da}^1$ and $\loss_\mathrm{da}^2$. 
In other words, we find that just minimizing the MMD loss between representations of a similar nature is more effective for domain alignment and minimizing domain discrepancy. To further clarify, teachers are slowly updated than the students and teachers provide global representations while students focus on local representations. We use $\loss_\mathrm{da}$ by default in all our experiments.

\subsection{Design choices for feature refinement} \label{supsec:feature_refinement}
Here, we perform additional studies to evaluate our design choices for the $\mathrm{refine}$ step.

\noindent\textbf{Cross-modal attention map.}
To calculate cross-modal feature relevance, we use the cross-modal attention maps as mentioned in Equation \ref{eq:cmattn}.
We find that a potential alternative to our default 
Equation \ref{eq:cmattn}
can be \Eqref{eq:cmattn_sm}. In particular, the scaling operation in 
Equation \ref{eq:cmattn}
can be simply replaced by a $\softmax$ function. 
\begin{equation}\label{eq:cmattn_sm}
\fontsize{9pt}{10pt}\selectfont
\begin{split}
    \emA_{v}^{'\times} = \softmax(\mathrm{MeanPool}(\emA_{v} \cdot \emA_{a}^T)); %
    \\
    \emA_{a}^{'\times} = \softmax(\mathrm{MeanPool}(\emA_{a} \cdot \emA_{v}^T))
\end{split}
\end{equation}
We notice that this technique works slightly better on audio representations in comparison to our default setup; however, a considerable performance drop is noticed on visual representations. Please see the results in Table \ref{tab:addl_exp}. Please note that unless mentioned otherwise, we use the default Equation \ref{eq:cmattn} in all the setups.

\begin{table}[!h]
\fontsize{9pt}{10pt}\selectfont
\centering

    \centering
    \begin{tabular}{lccc}
    \toprule
         & $\mathbf{\emA^{'\times}}$ & \textbf{w/o \texttt{CLS}} & \textbf{default} \\ \midrule\midrule
         HMDB51  & 54.6 & 54.5 & \textbf{58.5} \\
         UCF101  & 79.8 & 79.3 & \textbf{82.6} \\
         ESC50  & \textbf{88.3} & 84.5 &  88.0 \\
        \bottomrule
    \end{tabular}
\caption{\textbf{Exploring design choices for feature refinement.} We find that an alternative strategy to obtain the cross-modal attention maps ($\mathbf{\emA^{'\times}}$) may work marginally better on audio, but it significantly degrades the video performance. Additionally, we notice that the models benefit from the availability of \texttt{CLS} token in both audio and visual tasks. 
These results confirm the superiority of our proposed design choices for effective cross-modal knowledge distillation.}
\label{tab:addl_exp}
\end{table}

\noindent\textbf{CLS token.}
As mentioned in 
Equation \ref{eq:attn_map}, by default we use the \texttt{CLS} tokens to obtain $\emA_v$ and $\emA_a$, which are then used to generate $\emA_{v}^{'\times}$ and $\emA_{a}^{'\times}$. Here, we explicitly study if \texttt{CLS} tokens are necessary in our framework. To test this, we modify 
Equation \ref{eq:attn_map}
and calculate the intra-modal attention map as the correlation of the mean of the query embeddings and the key embeddings of all other patches. Please note that the rest of the setup remains the same. Table \ref{tab:addl_exp} shows that removing the \texttt{CLS} token significantly degrades the model performance on both audio and visual representations.

\subsection{Comparing different variants of XKD}\label{supsec:xkd_variants}
We conduct a thorough comparison between XKD and its different variants. In Table \ref{tab:ma_supp}, we present a detailed comparison between the modality-specific and modality-agnostic variants using both teacher and student encoders. As the teachers and students may converge at different rates due to their difference in the weight update schedule, these variants are trained for full training schedules of $800$ epochs. We report top-1 accuracy of linear evaluations using the split-1 of UCF101, HMDB51, and ESC50. As shown in Table \ref{tab:ma_supp}, modality-agnostic variants show performance drops (e.g., $3.2\%$ on UCF101 and $0.7\%$ on ESC50) in comparison to the modality-specific one. Moreover, as presented in the main paper, the performance gap between the modality-agnostic and modality-specific variants further reduces when finetuned (e.g., 0.7\% on UCF101, 1.7\% on Kinetics400, and 0.3\% on FSD50K).
Amongst the modality-agnostic backbones, XKD-MATS works slightly better on ESC50, whereas, XKD-MAS shows better performance on both UCF101 and HMDB51. Lastly, when comparing the performance between the teachers and students, teachers show slightly better performance on all the benchmarks and in all the setups. Following \cite{dino}, we adopt the terminology of `teacher-student' to simply refer to the stronger network as a teacher and the slightly weaker one as a student.

\begin{table}[]
\fontsize{9pt}{10pt}\selectfont
\setlength\tabcolsep{2pt}
\centering
    \resizebox{0.99\columnwidth}{!}{%
    \begin{tabular}{lcclcclcc}
    \toprule
        & \multicolumn{2}{c}{\textbf{XKD-MATS}} && \multicolumn{2}{c}{\textbf{XKD-MAS}} && \multicolumn{2}{c}{\textbf{XKD}} \\ \cmidrule{2-3} \cmidrule{5-6} \cmidrule{8-9} 
         & T (MA) & S (MA) && T (MS) & S (MA) && T (MS) & S (MS) \\ \midrule\midrule
         HMDB51 & 53.5 & 52.4 && 56.4 & \underline{55.2}\decblue{4.1} &&\textbf{59.3} & 57.8 \\ %
         UCF101& 80.3 & 80.0 && 81.9 & \underline{81.5}\decblue{3.2} && \textbf{84.7} & 84.4 \\ %
         ESC50 & \underline{90.3}\decblue{0.7} & 89.8 && 91.0 & 89.5  && \textbf{91.0} & 90.3 \\ %
        \bottomrule
    \end{tabular}
    }
\caption{\textbf{Comparison between teachers (T) and students (S) for both MA and MS variants.} We highlight the performance drop in the \underline{best} MA variants with respect to the \textbf{best} MS variant. Moreover, teachers always perform better with respect to the students.}
\label{tab:ma_supp}

\end{table}

\subsection{Training schedule}\label{supsec:epoch_acc}

\begin{figure}[!h]
    \centering
    \includegraphics[width=0.99\columnwidth]{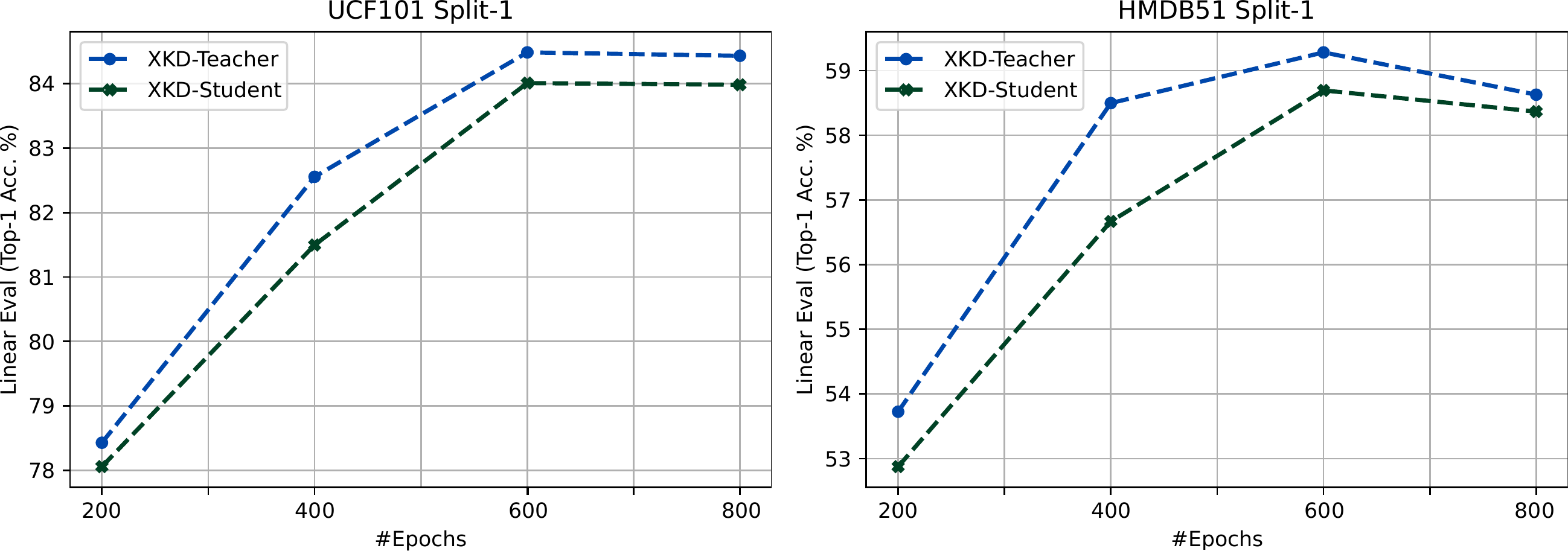}
    \caption{\textbf{Effect of longer pretraining schedules} with respect to the downstream task performance. We notice that the teacher always outperforms the students throughout the training. Additionally, little or no improvements are noticed beyond $600$ epochs. 
    }
    \label{fig:ablation_xkd_vid}
\end{figure}

In Figure \ref{fig:ablation_xkd_vid}, we present the effect of longer pretraining schedules on video action classification for both teacher and student networks. We pretrain the XKD for up to $800$ epochs and find that pretraining for around $600$ epochs shows improvement, while beyond that point, a slight performance drop is noticed on HMDB51 and no change is noticed on UCF101. Moreover, we find that the teacher always outperforms the student in downstream evaluation.

\begin{table*}[!t]
\fontsize{9pt}{10pt}\selectfont
\setlength\tabcolsep{2pt}
\centering

\begin{tabular}{lcccccccc}
\toprule
{\bf Method} & {\bf Backbone} & {\bf Input} & {\bf Patch Size} 
& {\bf \# Tokens} & \bf UCF101 & \bf HMDB51 & \bf Kinetics-Sound & \bf Kinetics400$^\dag$ \\ \midrule\midrule

\fadetext{VideoMAE$^{**}$ (224) \cite{videomae}} & \fadetext{ViT-B} &  \fadetext{$16\times224^2$} & \fadetext{$2\times16^2$} & \fadetext{1568} & \fadetext{80.0} & \fadetext{54.3} & \fadetext{-} & \fadetext{-} \\
{VideoMAE} (112) & ViT-B &  $16\!\times112^2$ & $2\!\times16^2$ & 392 & 77.1 & 51.6 & - & - \\
{VideoMAE}$^*$ & {ViT-B} & $32\!\times112^2$ & $4\!\times16^2$ & 392 & 78.9 & 51.5  & 56.8 & 30.7/38.4 \\

\midrule
\textbf{XKD} & ViT-B & $32\!\times112^2$ & $4\!\times16^2$ & \textbf{392} & \textbf{83.8} & \textbf{57.4} & \textbf{70.7} & \textbf{46.4/51.4}   \\ 
\bottomrule
\end{tabular}%
\caption{\textbf{Comparison between VideoMAE and XKD} in linear evaluation. We report top-1 accuracy, averaged over all the splits for UCF101 and HMDB51. 
As Kinetics400 is sufficiently large, we perform linear evaluations in both setups, SVM and FC-tuning. 
XKD outperforms VideoMAE$^*$ by $4.9\%$ on UCF101, $5.9\%$ on HMDB51, $13.9\%$ on Kinetics-Sound, and by $15.7\%$ on Kinetics400. VideoMAE$^{**}$ (224) uses $1568$ visual tokens, whereas, XKD uses $392$ tokens, reducing the pretraining computational cost for the video encoder by $4$ times.
Note: $^{*}$ identical setup to XKD, $^{**}$ industry level computation, $^\dag$SVM/FC. 
}
\label{tab:addl_comparison_mae}

\end{table*}

\subsection{Detailed comparison with VideoMAE}\label{supsec:comp_videomae}

In addition to the earlier results presented in the main paper, here we perform a thorough comparison of our method to video masked reconstruction, i.e., VideoMAE \cite{videomae}, on video action recognition, and present the results in Table \ref{tab:addl_comparison_mae}. In this experiment, we strictly follow the standard linear evaluation protocol \cite{avid,xdc,crisscross,xiao2020audiovisual} since it is a more appropriate method for evaluating self-supervised representation learning as pointed out in \cite{xiao2020audiovisual,cvrl}. Our comparisons are mainly 3-fold. 

First, we compare XKD to VideoMAE \cite{videomae} and notice that XKD outperforms VideoMAE\footnote{We perform linear evaluations using the official checkpoints released at: \url{https://github.com/MCG-NJU/VideoMAE/tree/main}} by $3.8\%$ and $3.1\%$ on UCF101 and HMDB51, respectively. The improvements in XKD's results clearly show the benefits of cross-modal knowledge distillation in learning better and more discriminative representations. However, it should be noted that VideoMAE \cite{videomae} is pretrained using industry-level computation ($64$ $32$GB V100 GPUs) with a higher frame resolution of $224^2$, whereas, we use a lower resolution of $112^2$ (using just $8$ V100 GPUs). We use a frame resolution of $112^2$ to reduce the computation cost, i.e., while VideoMAE \cite{videomae} is pretrained with $1568$ visual tokens, XKD uses $392$ tokens, reducing the computation cost by almost $4\times$. 

Therefore, to have a fair comparison between these two methods, we pretrain VideoMAE with a frame resolution of $112^2$ using the publicly available VideoMAE \cite{videomae} codes. We refer to this variant as `VideoMAE (112)' as presented in Table \ref{tab:addl_comparison_mae}. XKD outperforms VideoMAE (112) by $6.7\%$ and $5.8\%$ on UCF101 and HMDB51. Lastly, we find another minor difference in the input setup between XKD and VideoMAE. VideoMAE uses 16 frames as input with a temporal patch size of $2$, whereas we use $32$ frames and a temporal patch size of $4$. Therefore, for a more fair comparison, we pretrain VideoMAE in an identical setup to ours, i.e, with visual inputs of $32\times112^2$ and patch size of $4\times16$, and denote it as VideoMAE$^*$ in Table \ref{tab:addl_comparison_mae}. As presented in Table \ref{tab:addl_comparison_mae}, XKD outperforms VideoMAE$^*$ by $4.9\%$ and $5.9\%$ on UCF101 and HMDB51, respectively. 
Lastly, XKD outperforms VideoMAE$^*$ by a very large margin of $13.9$ and $15.7$ on Kinetics-Sound and Kinetics400. Such large performance improvement demonstrates that XKD learns better features compared to video-masked reconstruction frameworks.

\section{Qualitative analysis} \label{supsec:qualitative}

\subsection{Effect of feature refinement} \label{supsec:qualitative_refinement}
In Figure \ref{fig:refine_supp}, we present several examples showing the impact of our proposed $\mathrm{refine}$ strategy in identifying the key visual features.

\subsection{XKD reconstruction results} \label{supsec:qualitative_inpainting}
In Figures \ref{fig:inpaint_supp}, \ref{fig:inpaint_supp1} and \ref{fig:inpaint_supp2}, we present several examples showing the reconstruction capability of XKD with respect to varying masked inputs. Additional examples of reconstruction of video frames and audio spectrograms from highly masked inputs are presented in Figures \ref{fig:frame_recon} and \ref{fig:spec_recon}. XKD demonstrates high reconstruction performance even when a high mask ratio is applied.

\begin{table*}[]%
\centering
\fontsize{9pt}{10pt}\selectfont
\setlength\tabcolsep{2pt}
\renewcommand{\arraystretch}{-5} %
\setlength{\arrayrulewidth}{2pt}
\renewcommand{\columnseprulecolor}{\color{red}}

\resizebox{\linewidth}{!}{%
    \begin{tabular}{lcccrlccc}
     & \bf \Huge raw frames & \bf \Huge without refinement & \bf \Huge with refinement &&& \bf \Huge raw frames & \bf \Huge without refinement & \bf \Huge with refinement 
     \\
     \rotatebox{90}{\Large ~~~~~~~~~~abseiling} 
     & \includegraphics[width=0.48\linewidth]{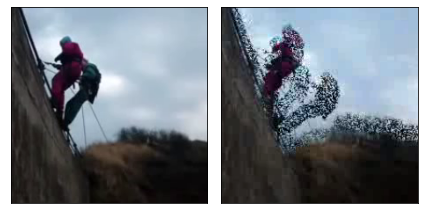}    
     & \includegraphics[width=0.48\linewidth]{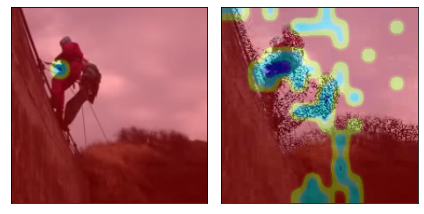}
     & \includegraphics[width=0.48\linewidth]{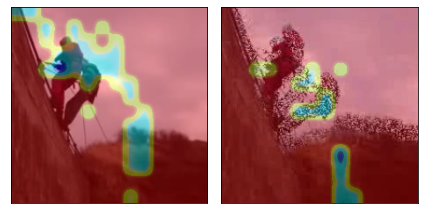} &&
     \rotatebox{90}{\Large ~~~~~~~~arm wrestling} 
     & \includegraphics[width=0.48\linewidth]{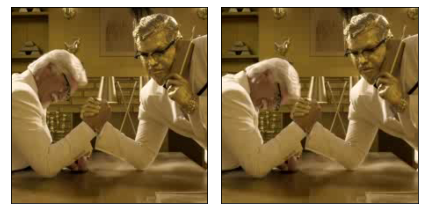}    
     & \includegraphics[width=0.48\linewidth]{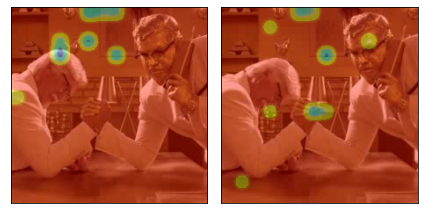}
     & \includegraphics[width=0.48\linewidth]{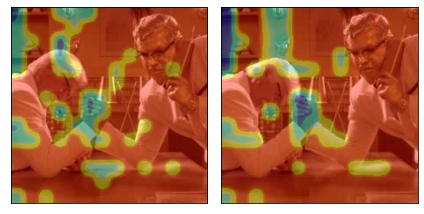} \\
     \rotatebox{90}{\Large ~~~~~~~bending metal} 
     & \includegraphics[width=0.48\linewidth]{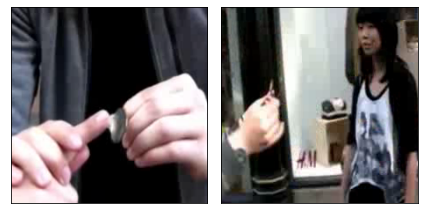}    
     & \includegraphics[width=0.48\linewidth]{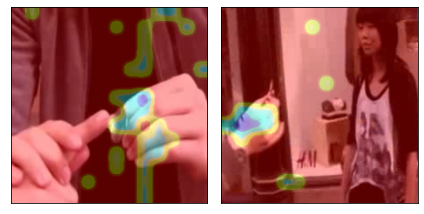}
     & \includegraphics[width=0.48\linewidth]{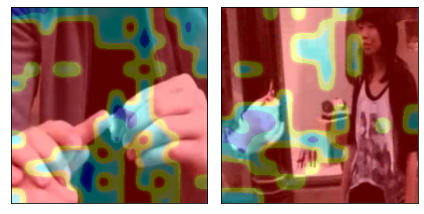} &&     
     \rotatebox{90}{\Large ~~~~~~~~~doing nails} 
     & \includegraphics[width=0.48\linewidth]{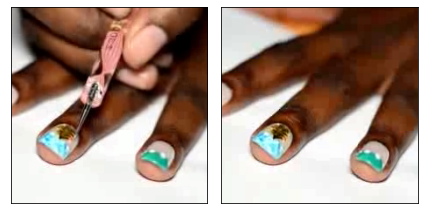}    
     & \includegraphics[width=0.48\linewidth]{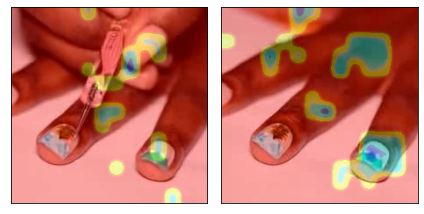}
     & \includegraphics[width=0.48\linewidth]{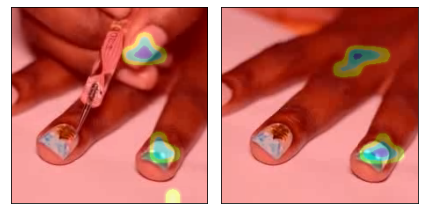} \\
     \rotatebox{90}{\Large ~~~~~~~~~~driving car} 
     & \includegraphics[width=0.48\linewidth]{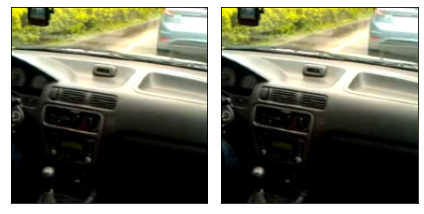}    
     & \includegraphics[width=0.48\linewidth]{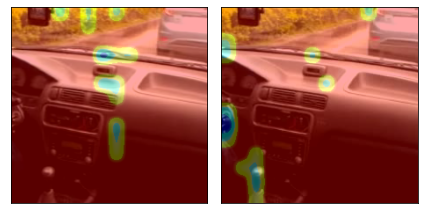}
     & \includegraphics[width=0.48\linewidth]{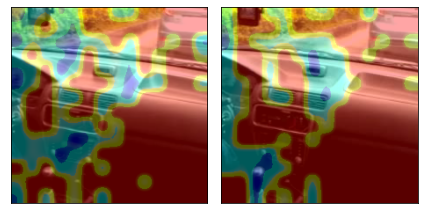} &&
     \rotatebox{90}{\Large ~~~~~~eating ice cream} 
     & \includegraphics[width=0.48\linewidth]{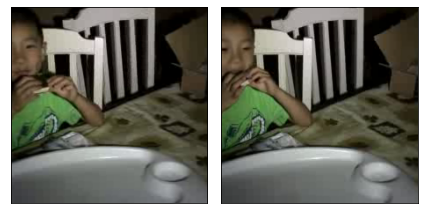}    
     & \includegraphics[width=0.48\linewidth]{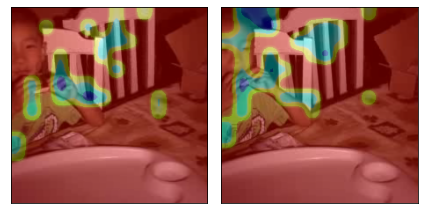}
     & \includegraphics[width=0.48\linewidth]{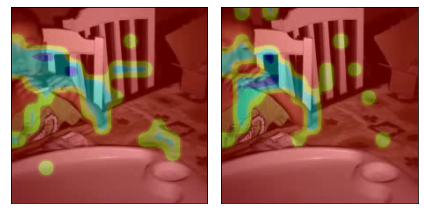} \\
     \rotatebox{90}{\Large ~~~~~~~shuffling cards} 
     & \includegraphics[width=0.48\linewidth]{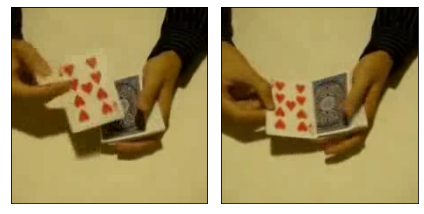}    
     & \includegraphics[width=0.48\linewidth]{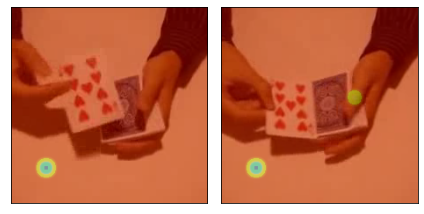}
     & \includegraphics[width=0.48\linewidth]{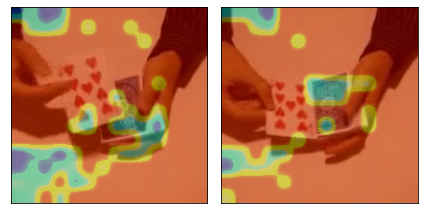} &&
     \rotatebox{90}{\Large ~~~~~~~~~side kick} 
     & \includegraphics[width=0.48\linewidth]{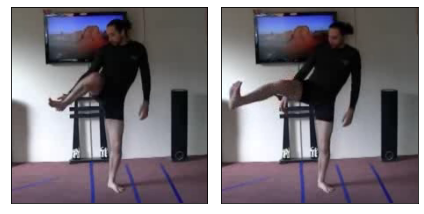}  
     & \includegraphics[width=0.48\linewidth]{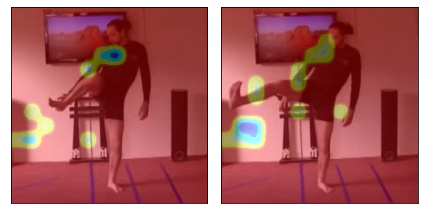}
     & \includegraphics[width=0.48\linewidth]{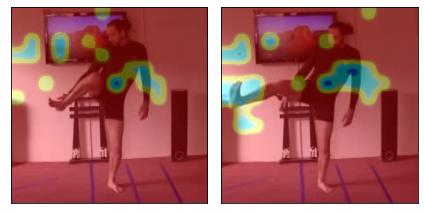} \\
     \rotatebox{90}{\Large passing american football} 
     & \includegraphics[width=0.48\linewidth]{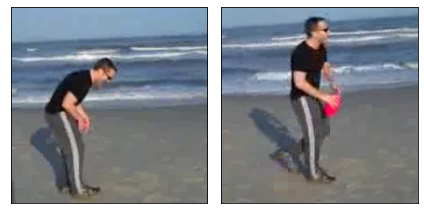}    
     & \includegraphics[width=0.48\linewidth]{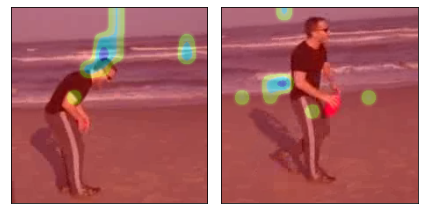}
     & \includegraphics[width=0.48\linewidth]{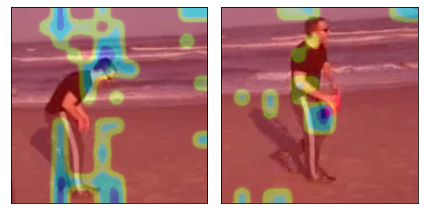} &&
     \rotatebox{90}{\Large ~~~~~~~playing harp} 
     & \includegraphics[width=0.48\linewidth]{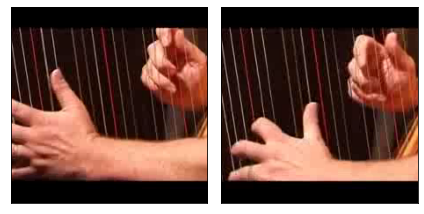}    
     & \includegraphics[width=0.48\linewidth]{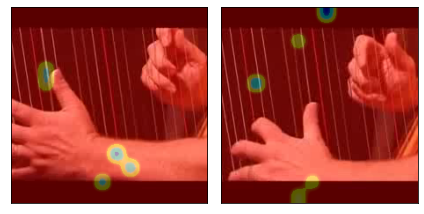}
     & \includegraphics[width=0.48\linewidth]{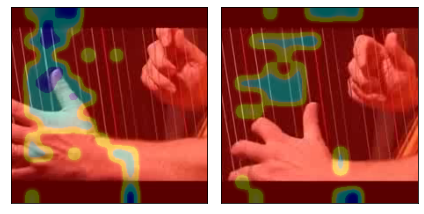} \\
     \rotatebox{90}{\Large ~~~~~~~~playing organ} 
     & \includegraphics[width=0.48\linewidth]{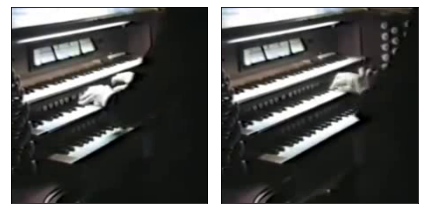}    
     & \includegraphics[width=0.48\linewidth]{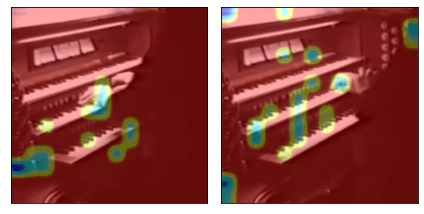}
     & \includegraphics[width=0.48\linewidth]{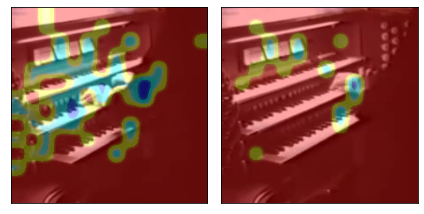} &&
     \rotatebox{90}{\Large ~~~~~~~~playing poker} 
     & \includegraphics[width=0.48\linewidth]{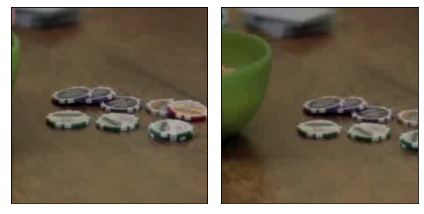}    
     & \includegraphics[width=0.48\linewidth]{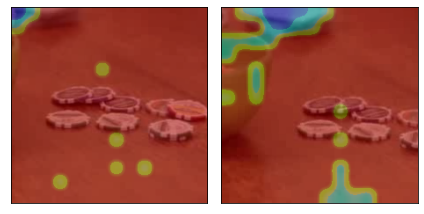}
     & \includegraphics[width=0.48\linewidth]{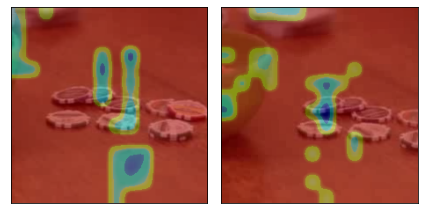} \\
     \rotatebox{90}{\Large ~~~~~~~~robot dancing} 
     & \includegraphics[width=0.48\linewidth]{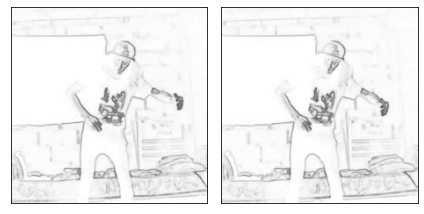}    
     & \includegraphics[width=0.48\linewidth]{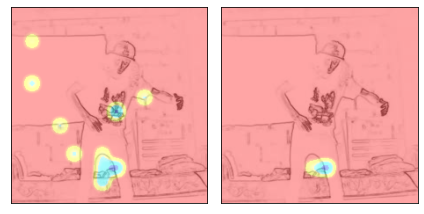}
     & \includegraphics[width=0.48\linewidth]{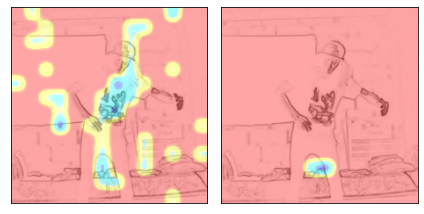} &&
     \rotatebox{90}{\Large ~~~~~~~~shining shoes} 
     & \includegraphics[width=0.48\linewidth]{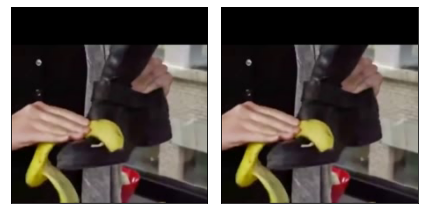}    
     & \includegraphics[width=0.48\linewidth]{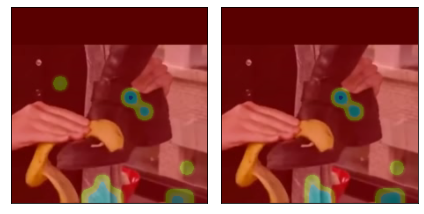}
     & \includegraphics[width=0.48\linewidth]{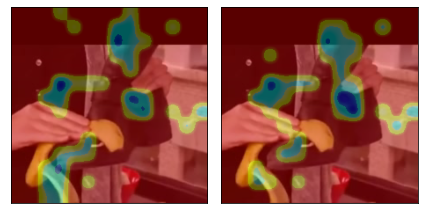} \\
     \rotatebox{90}{\Large ~~~~~~~~~~~~~skiing} 
     & \includegraphics[width=0.48\linewidth]{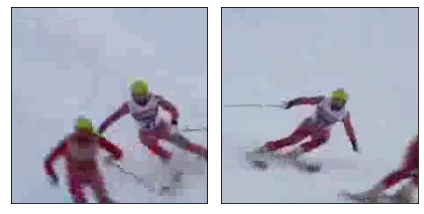}    
     & \includegraphics[width=0.48\linewidth]{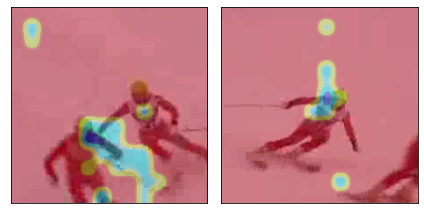}
     & \includegraphics[width=0.48\linewidth]{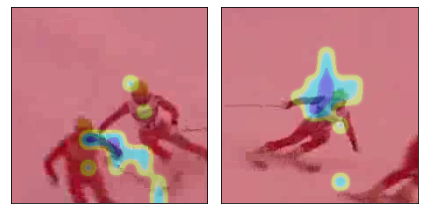} &&
     \rotatebox{90}{\Large ~~~~~~~~spray painting} 
     & \includegraphics[width=0.48\linewidth]{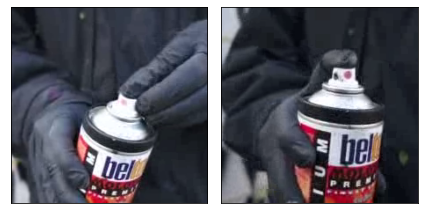}    
     & \includegraphics[width=0.48\linewidth]{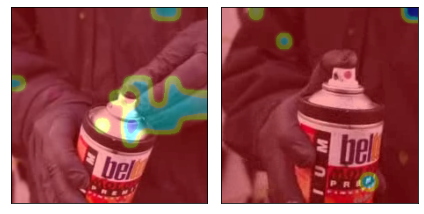}
     & \includegraphics[width=0.48\linewidth]{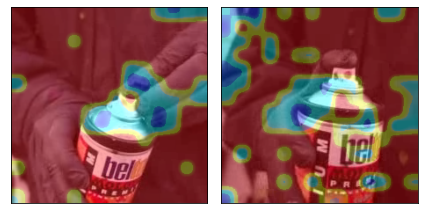} \\
    \end{tabular}
}
\captionof{figure}{Visualizing the \textbf{effect of} $\mathrm{\textbf{refine}}$ in identifying the key visual features. (Best viewed in color.)}
\label{fig:refine_supp}
\end{table*}

\begin{table*}
\centering
\setlength\tabcolsep{0pt}
\renewcommand{\arraystretch}{-10} %
\resizebox{0.9\linewidth}{!}{%
\begin{tabular}{lcc}
    \multicolumn{3}{c}{Original frames}\\
     \multicolumn{3}{c}{
     \includegraphics[height=0.1\linewidth]{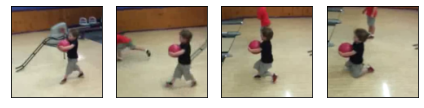}
     }\\
     & Masked frames & Reconstructed frames \\
     \rotatebox{90}{~~~~~~$60\%$}&
      \includegraphics[height=0.1\linewidth]{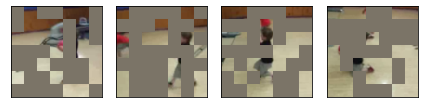} 
     &
     \includegraphics[height=0.1\linewidth]{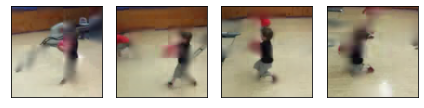}
     \\
     \rotatebox{90}{~~~~~~$70\%$}&
      \includegraphics[height=0.1\linewidth]{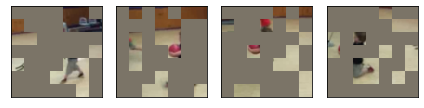} 
     &
     \includegraphics[height=0.1\linewidth]{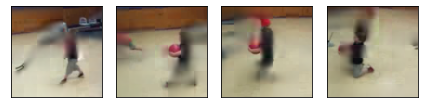}
     \\
     \rotatebox{90}{~~~~~~$80\%$}&
     \includegraphics[height=0.1\linewidth]{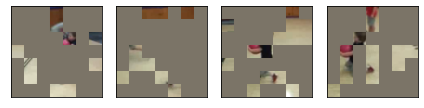} 
     &
     \includegraphics[height=0.1\linewidth]{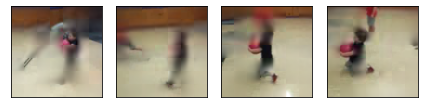}
     \\
     \rotatebox{90}{~~~~~~$90\%$}&
     \includegraphics[height=0.1\linewidth]{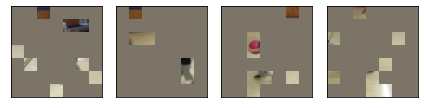} 
     &
     \includegraphics[height=0.1\linewidth]{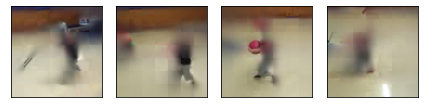}
     \\
     \multicolumn{3}{c}{Original spectrogram}\\
     \multicolumn{3}{c}{
     \includegraphics[height=0.1\linewidth,width=0.39\linewidth]{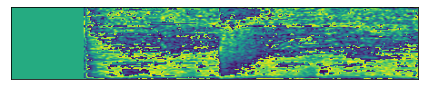}
     }\\
     & Masked spectrogram & Reconstructed spectrogram \\
     \rotatebox{90}{~~~~~~$50\%$}&
      \includegraphics[height=0.1\linewidth,width=0.39\linewidth]{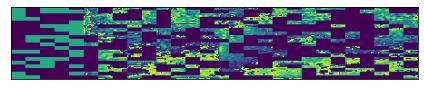} 
     &
     \includegraphics[height=0.1\linewidth,width=0.39\linewidth]{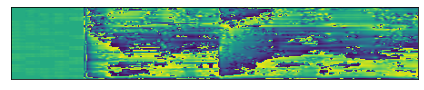}
     \\
     \rotatebox{90}{~~~~~~$60\%$}&
      \includegraphics[height=0.1\linewidth,width=0.39\linewidth]{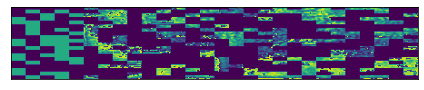} 
     &
     \includegraphics[height=0.1\linewidth,width=0.39\linewidth]{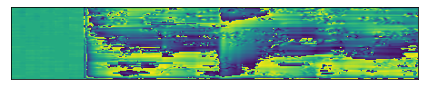}
     \\
     \rotatebox{90}{~~~~~~$70\%$}&
     \includegraphics[height=0.1\linewidth,width=0.39\linewidth]{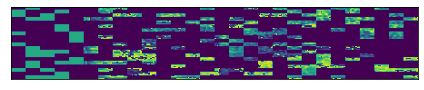} 
     &
     \includegraphics[height=0.1\linewidth,width=0.39\linewidth]{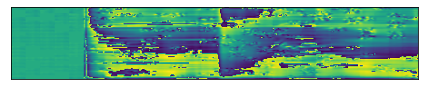}
     \\
     \rotatebox{90}{~~~~~~$80\%$}&
     \includegraphics[height=0.1\linewidth,width=0.39\linewidth]{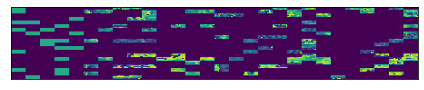} 
     &
     \includegraphics[height=0.1\linewidth,width=0.39\linewidth]{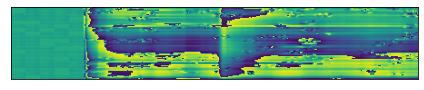}
     \\

\end{tabular}
}
\captionof{figure}{Examples of reconstruction with \textbf{varying masked inputs}. We use the pretrained XKD and during inference, the video mask ratio is varied from $60\%$ to $90\%$ and the audio mask ratio is varied from $50\%$ to $80\%$. XKD demonstrates high reconstruction performance even when a very high mask ratio is applied for both audio and visual modalities. 
}
\label{fig:inpaint_supp}
\end{table*}

\begin{table*}
\centering
\setlength\tabcolsep{0pt}
\renewcommand{\arraystretch}{-10} %
\resizebox{0.9\linewidth}{!}{%
\begin{tabular}{lcc}
    \multicolumn{3}{c}{Original frames}\\
     \multicolumn{3}{c}{
     \includegraphics[height=0.1\linewidth]{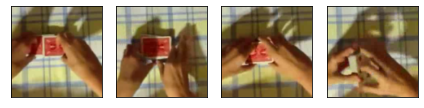}
     }\\
     & Masked frames & Reconstructed frames \\
     \rotatebox{90}{~~~~~~$60\%$}&
      \includegraphics[height=0.1\linewidth]{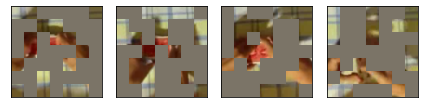} 
     &
     \includegraphics[height=0.1\linewidth]{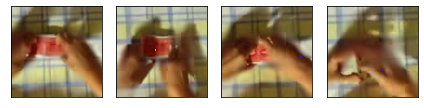}
     \\
     \rotatebox{90}{~~~~~~$70\%$}&
      \includegraphics[height=0.1\linewidth]{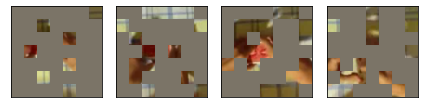} 
     &
     \includegraphics[height=0.1\linewidth]{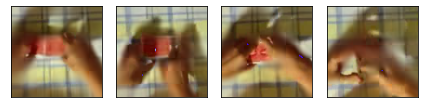}
     \\
     \rotatebox{90}{~~~~~~$80\%$}&
     \includegraphics[height=0.1\linewidth]{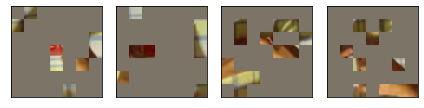} 
     &
     \includegraphics[height=0.1\linewidth]{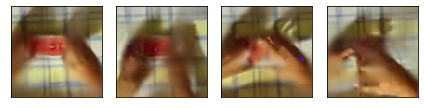}
     \\
     \rotatebox{90}{~~~~~~$90\%$}&
     \includegraphics[height=0.1\linewidth]{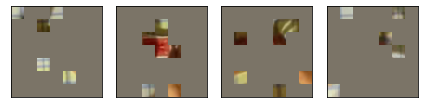} 
     &
     \includegraphics[height=0.1\linewidth]{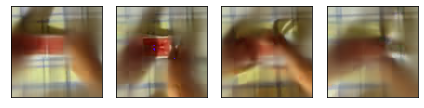}
     \\
     \multicolumn{3}{c}{Original spectrogram}\\
     \multicolumn{3}{c}{
     \includegraphics[height=0.1\linewidth,width=0.39\linewidth]{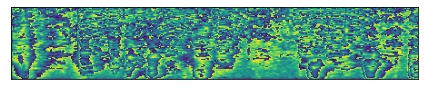}
     }\\
     & Masked spectrogram & Reconstructed spectrogram \\
     \rotatebox{90}{~~~~~~$50\%$}&
      \includegraphics[height=0.1\linewidth,width=0.39\linewidth]{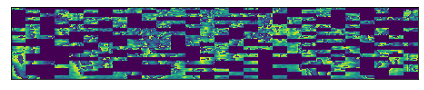} 
     &
     \includegraphics[height=0.1\linewidth,width=0.39\linewidth]{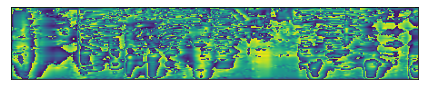}
     \\
     \rotatebox{90}{~~~~~~$60\%$}&
      \includegraphics[height=0.1\linewidth,width=0.39\linewidth]{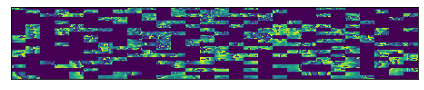} 
     &
     \includegraphics[height=0.1\linewidth,width=0.39\linewidth]{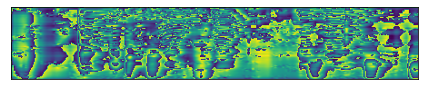}
     \\
     \rotatebox{90}{~~~~~~$70\%$}&
     \includegraphics[height=0.1\linewidth,width=0.39\linewidth]{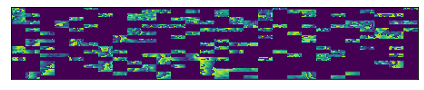} 
     &
     \includegraphics[height=0.1\linewidth,width=0.39\linewidth]{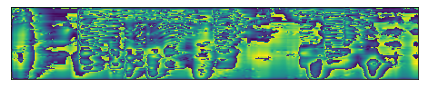}
     \\
     \rotatebox{90}{~~~~~~$80\%$}&
     \includegraphics[height=0.1\linewidth,width=0.39\linewidth]{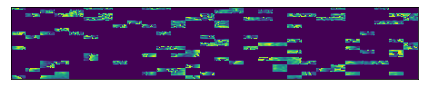} 
     &
     \includegraphics[height=0.1\linewidth,width=0.39\linewidth]{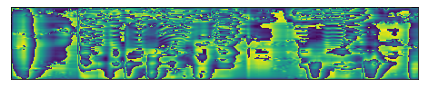}
     \\

\end{tabular}
}
\captionof{figure}{Examples of reconstruction with \textbf{varying masked inputs}. We use the pretrained XKD and during inference, the video mask ratio is varied from $60\%$ to $90\%$ and the audio mask ratio is varied from $50\%$ to $80\%$. XKD demonstrates high reconstruction performance even when a very high mask ratio is applied for both audio and visual modalities.
}
\label{fig:inpaint_supp1}
\end{table*}

\begin{table*}
\centering
\setlength\tabcolsep{0pt}
\renewcommand{\arraystretch}{-10} %
\resizebox{0.9\linewidth}{!}{%
\begin{tabular}{lcc}
    \multicolumn{3}{c}{Original frames}\\
     \multicolumn{3}{c}{
     \includegraphics[height=0.1\linewidth]{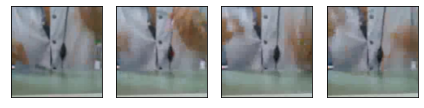}
     }\\
     & Masked frames & Reconstructed frames \\
     \rotatebox{90}{~~~~~~$60\%$}&
      \includegraphics[height=0.1\linewidth]{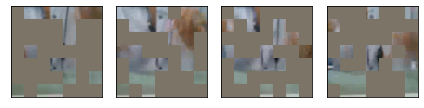} 
     &
     \includegraphics[height=0.1\linewidth]{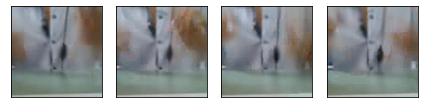}
     \\
     \rotatebox{90}{~~~~~~$70\%$}&
      \includegraphics[height=0.1\linewidth]{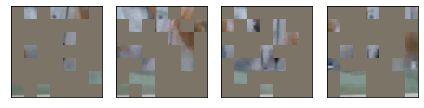} 
     &
     \includegraphics[height=0.1\linewidth]{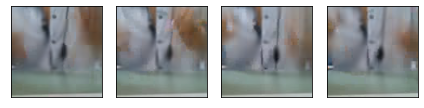}
     \\
     \rotatebox{90}{~~~~~~$80\%$}&
     \includegraphics[height=0.1\linewidth]{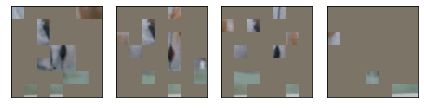} 
     &
     \includegraphics[height=0.1\linewidth]{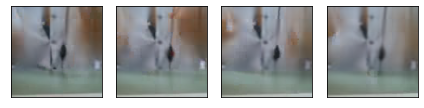}
     \\
     \rotatebox{90}{~~~~~~$90\%$}&
     \includegraphics[height=0.1\linewidth]{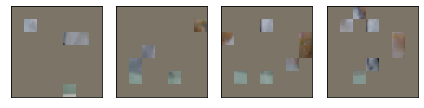} 
     &
     \includegraphics[height=0.1\linewidth]{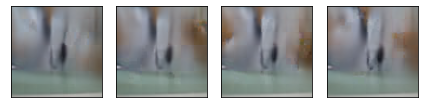}
     \\
     \multicolumn{3}{c}{Original spectrogram}\\
     \multicolumn{3}{c}{
     \includegraphics[height=0.1\linewidth,width=0.39\linewidth]{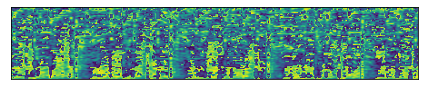}
     }\\
     & Masked spectrogram & Reconstructed spectrogram \\
     \rotatebox{90}{~~~~~~$50\%$}&
      \includegraphics[height=0.1\linewidth,width=0.39\linewidth]{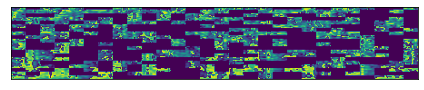} 
     &
     \includegraphics[height=0.1\linewidth,width=0.39\linewidth]{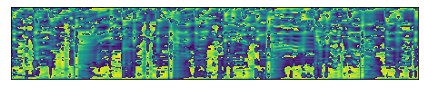}
     \\
     \rotatebox{90}{~~~~~~$60\%$}&
      \includegraphics[height=0.1\linewidth,width=0.39\linewidth]{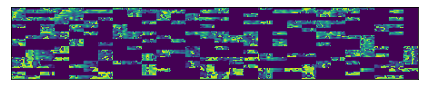} 
     &
     \includegraphics[height=0.1\linewidth,width=0.39\linewidth]{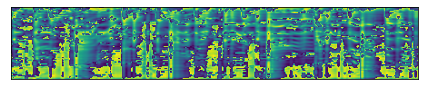}
     \\
     \rotatebox{90}{~~~~~~$70\%$}&
     \includegraphics[height=0.1\linewidth,width=0.39\linewidth]{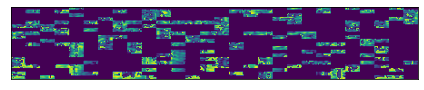} 
     &
     \includegraphics[height=0.1\linewidth,width=0.39\linewidth]{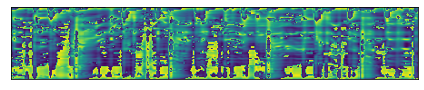}
     \\
     \rotatebox{90}{~~~~~~$80\%$}&
     \includegraphics[height=0.1\linewidth,width=0.39\linewidth]{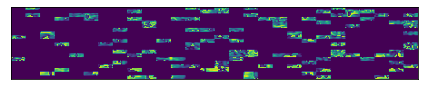} 
     &
     \includegraphics[height=0.1\linewidth,width=0.39\linewidth]{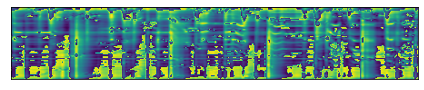}
     \\

\end{tabular}
}
\captionof{figure}{Examples of reconstruction with \textbf{varying masked inputs}. We use the pretrained XKD and during inference, the video mask ratio is varied from $60\%$ to $90\%$ and the audio mask ratio is varied from $50\%$ to $80\%$. XKD demonstrates high reconstruction performance even when a very high mask ratio is applied for both audio and visual modalities.
}
\label{fig:inpaint_supp2}
\end{table*}

\begin{table*}
\centering
\setlength\tabcolsep{0pt}
\renewcommand{\arraystretch}{-10} %
\resizebox{0.9\linewidth}{!}{%
\begin{tabular}{ccc}
\textbf{Original} & \textbf{Masked} & \textbf{Reconstructed} \\
\includegraphics[width=0.28\textwidth]{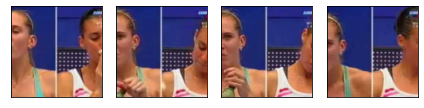}
&
\includegraphics[width=0.28\textwidth]{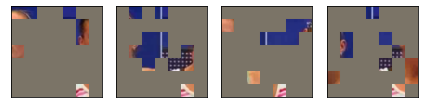}
& 
\includegraphics[width=0.28\textwidth]{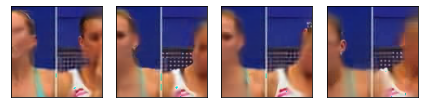}\\
\includegraphics[width=0.28\textwidth]{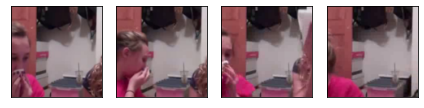}
&
\includegraphics[width=0.28\textwidth]{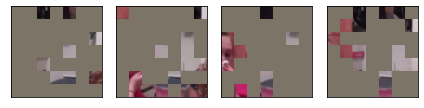}
& 
\includegraphics[width=0.28\textwidth]{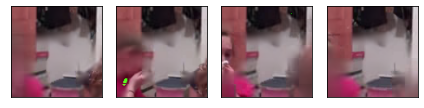}\\
\includegraphics[width=0.28\textwidth]{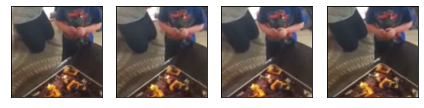}
&
\includegraphics[width=0.28\textwidth]{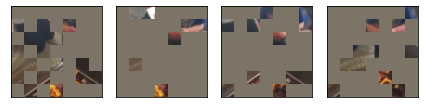}
& 
\includegraphics[width=0.28\textwidth]{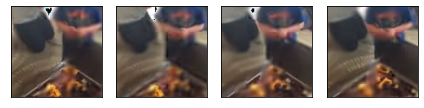}\\
\includegraphics[width=0.28\textwidth]{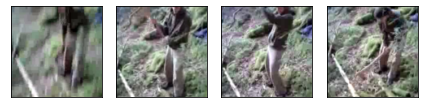}
&
\includegraphics[width=0.28\textwidth]{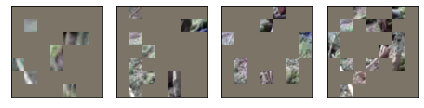}
& 
\includegraphics[width=0.28\textwidth]{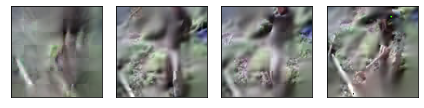}\\
\includegraphics[width=0.28\textwidth]{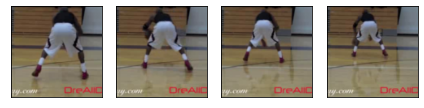}
&
\includegraphics[width=0.28\textwidth]{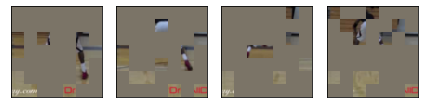}
& 
\includegraphics[width=0.28\textwidth]{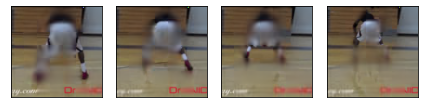}\\
\includegraphics[width=0.28\textwidth]{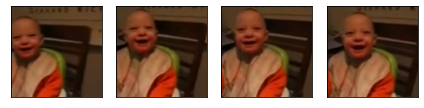}
&
\includegraphics[width=0.28\textwidth]{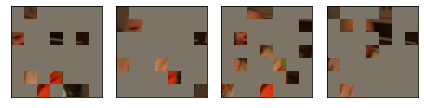}
& 
\includegraphics[width=0.28\textwidth]{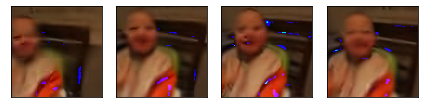}\\
\includegraphics[width=0.28\textwidth]{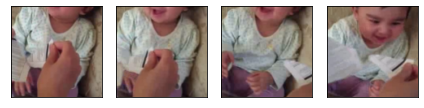}
&
\includegraphics[width=0.28\textwidth]{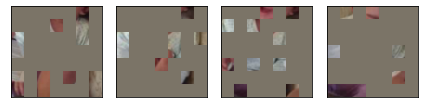}
& 
\includegraphics[width=0.28\textwidth]{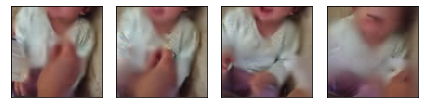}\\
\includegraphics[width=0.28\textwidth]{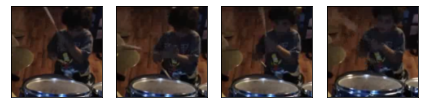}
&
\includegraphics[width=0.28\textwidth]{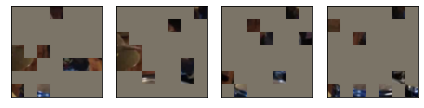}
& 
\includegraphics[width=0.28\textwidth]{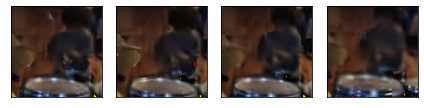}\\
\includegraphics[width=0.28\textwidth]{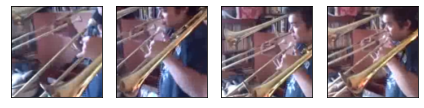}
&
\includegraphics[width=0.28\textwidth]{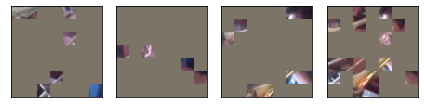}
& 
\includegraphics[width=0.28\textwidth]{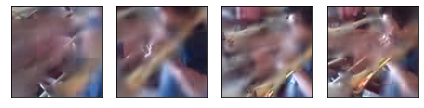}\\
\includegraphics[width=0.28\textwidth]{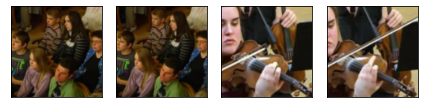}
&
\includegraphics[width=0.28\textwidth]{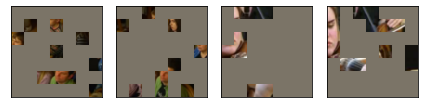}
& 
\includegraphics[width=0.28\textwidth]{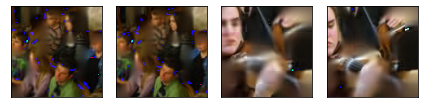}\\
\includegraphics[width=0.28\textwidth]{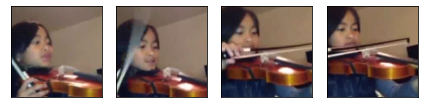}
&
\includegraphics[width=0.28\textwidth]{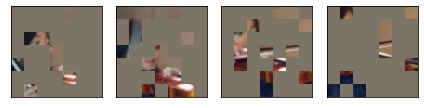}
& 
\includegraphics[width=0.28\textwidth]{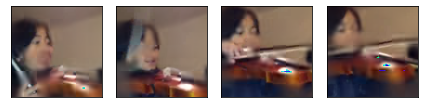}\\
\includegraphics[width=0.28\textwidth]{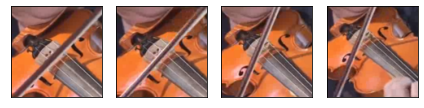}
&
\includegraphics[width=0.28\textwidth]{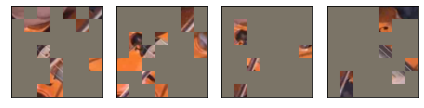}
& 
\includegraphics[width=0.28\textwidth]{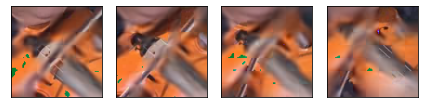}\\
\includegraphics[width=0.28\textwidth]{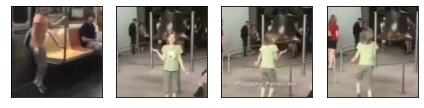}
&
\includegraphics[width=0.28\textwidth]{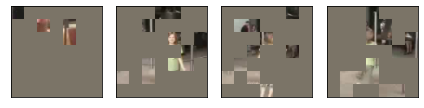}
& 
\includegraphics[width=0.28\textwidth]{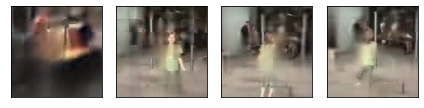}\\
\includegraphics[width=0.28\textwidth]{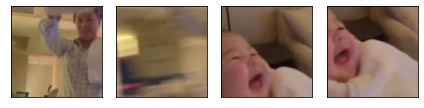}
&
\includegraphics[width=0.28\textwidth]{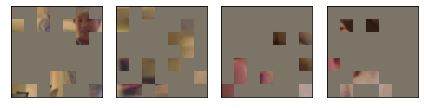}
& 
\includegraphics[width=0.28\textwidth]{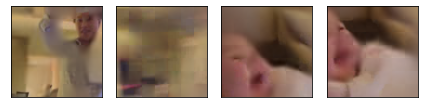}\\
\includegraphics[width=0.28\textwidth]{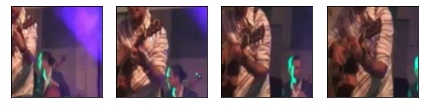}
&
\includegraphics[width=0.28\textwidth]{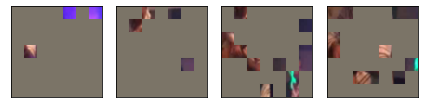}
& 
\includegraphics[width=0.28\textwidth]{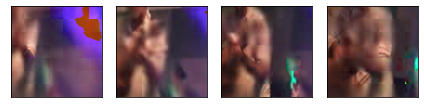}\\
\includegraphics[width=0.28\textwidth]{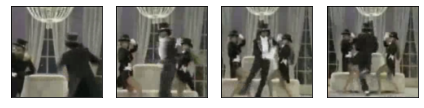}
&
\includegraphics[width=0.28\textwidth]{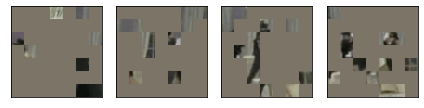}
& 
\includegraphics[width=0.28\textwidth]{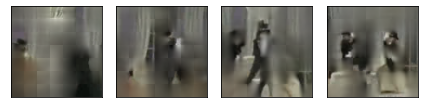}\\
\end{tabular}
}
\captionof{figure}{Examples of \textbf{video frame} reconstruction ability of XKD from highly masked ($80\%$) inputs. 
}
\label{fig:frame_recon}
\end{table*}

\begin{table*}
\centering
\setlength\tabcolsep{0pt}
\renewcommand{\arraystretch}{-10} %
\resizebox{0.9\linewidth}{!}{%
\begin{tabular}{ccc}
\textbf{Original} & \textbf{Masked} & \textbf{Reconstructed} \\
\includegraphics[width=0.28\textwidth]{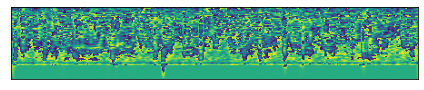}
&
\includegraphics[width=0.28\textwidth]{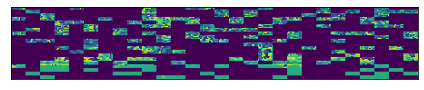}
& 
\includegraphics[width=0.28\textwidth]{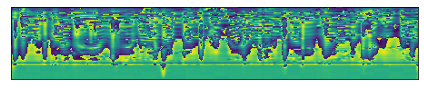}\\
\includegraphics[width=0.28\textwidth]{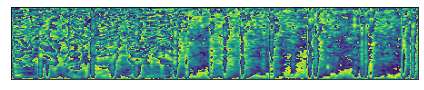}
&
\includegraphics[width=0.28\textwidth]{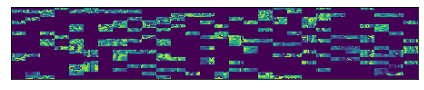}
& 
\includegraphics[width=0.28\textwidth]{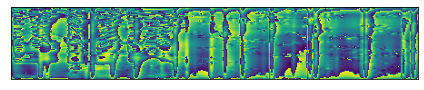}\\
\includegraphics[width=0.28\textwidth]{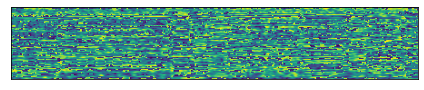}
&
\includegraphics[width=0.28\textwidth]{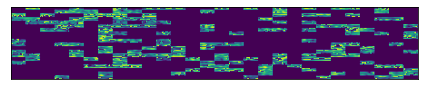}
& 
\includegraphics[width=0.28\textwidth]{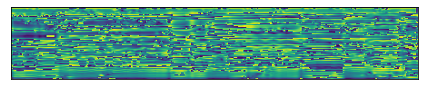}\\
\includegraphics[width=0.28\textwidth]{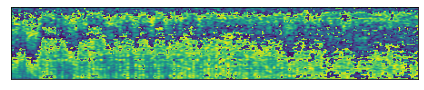}
&
\includegraphics[width=0.28\textwidth]{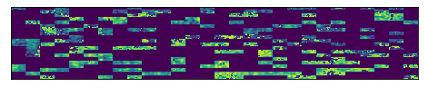}
& 
\includegraphics[width=0.28\textwidth]{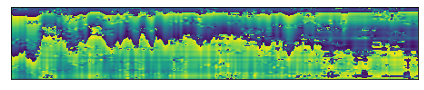}\\
\includegraphics[width=0.28\textwidth]{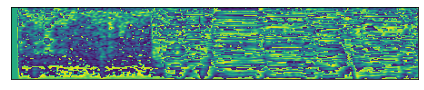}
&
\includegraphics[width=0.28\textwidth]{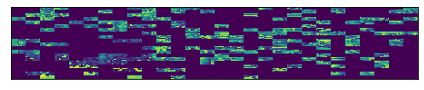}
& 
\includegraphics[width=0.28\textwidth]{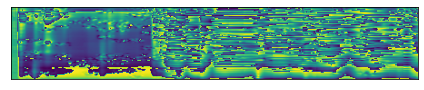}\\
\includegraphics[width=0.28\textwidth]{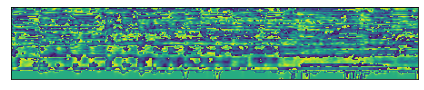}
&
\includegraphics[width=0.28\textwidth]{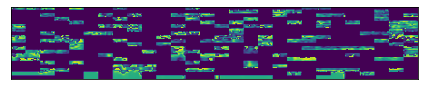}
& 
\includegraphics[width=0.28\textwidth]{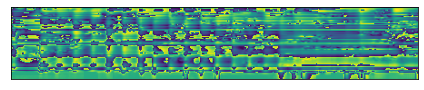}\\
\includegraphics[width=0.28\textwidth]{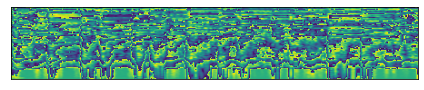}
&
\includegraphics[width=0.28\textwidth]{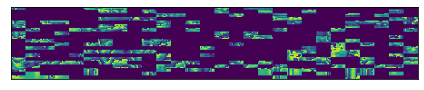}
& 
\includegraphics[width=0.28\textwidth]{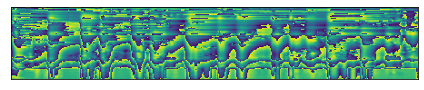}\\
\includegraphics[width=0.28\textwidth]{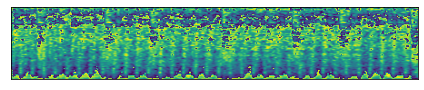}
&
\includegraphics[width=0.28\textwidth]{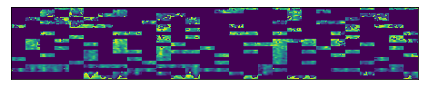}
& 
\includegraphics[width=0.28\textwidth]{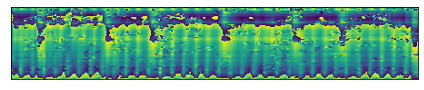}\\
\includegraphics[width=0.28\textwidth]{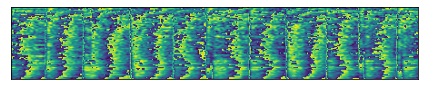}
&
\includegraphics[width=0.28\textwidth]{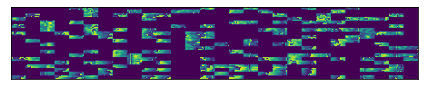}
& 
\includegraphics[width=0.28\textwidth]{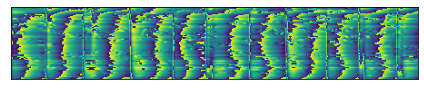}\\
\includegraphics[width=0.28\textwidth]{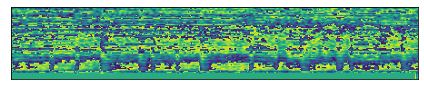}
&
\includegraphics[width=0.28\textwidth]{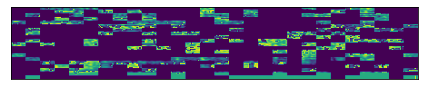}
& 
\includegraphics[width=0.28\textwidth]{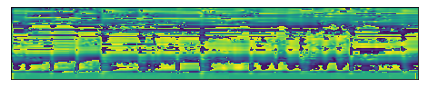}\\
\includegraphics[width=0.28\textwidth]{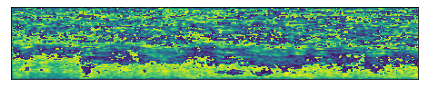}
&
\includegraphics[width=0.28\textwidth]{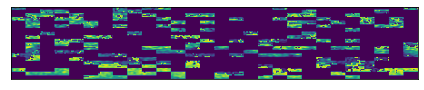}
& 
\includegraphics[width=0.28\textwidth]{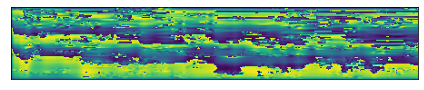}\\
\includegraphics[width=0.28\textwidth]{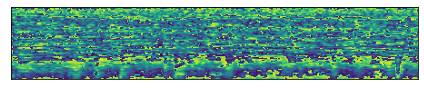}
&
\includegraphics[width=0.28\textwidth]{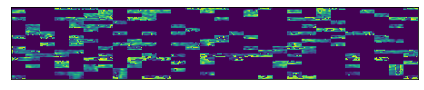}
& 
\includegraphics[width=0.28\textwidth]{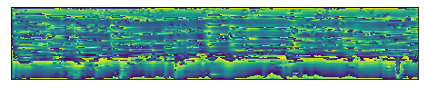}\\
\includegraphics[width=0.28\textwidth]{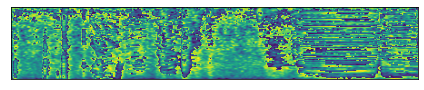}
&
\includegraphics[width=0.28\textwidth]{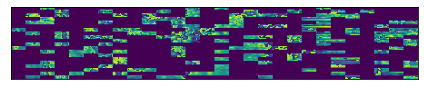}
& 
\includegraphics[width=0.28\textwidth]{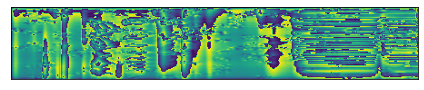}\\
\includegraphics[width=0.28\textwidth]{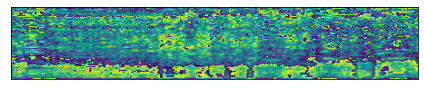}
&
\includegraphics[width=0.28\textwidth]{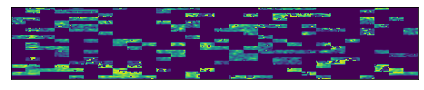}
& 
\includegraphics[width=0.28\textwidth]{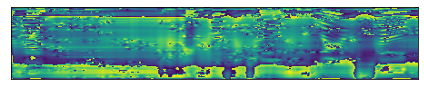}\\
\includegraphics[width=0.28\textwidth]{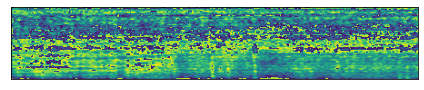}
&
\includegraphics[width=0.28\textwidth]{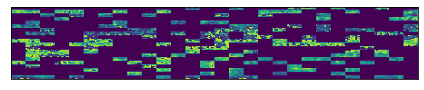}
& 
\includegraphics[width=0.28\textwidth]{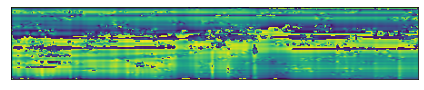}\\
\includegraphics[width=0.28\textwidth]{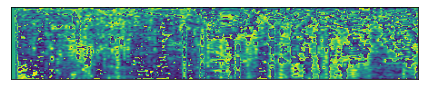}
&
\includegraphics[width=0.28\textwidth]{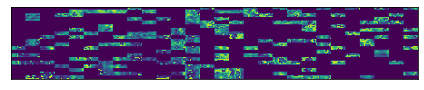}
& 
\includegraphics[width=0.28\textwidth]{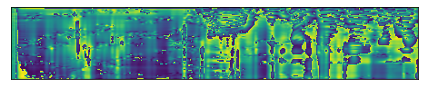}\\
\includegraphics[width=0.28\textwidth]{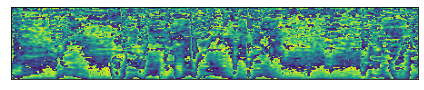}
&
\includegraphics[width=0.28\textwidth]{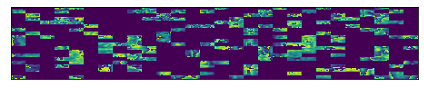}
& 
\includegraphics[width=0.28\textwidth]{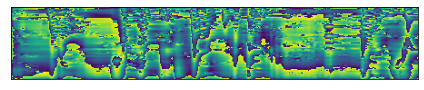}\\
\includegraphics[width=0.28\textwidth]{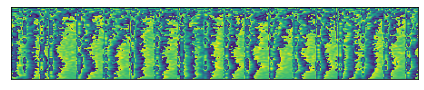}
&
\includegraphics[width=0.28\textwidth]{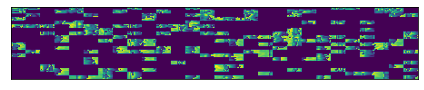}
& 
\includegraphics[width=0.28\textwidth]{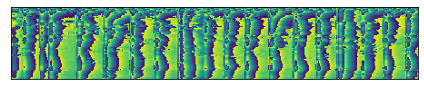}\\
\includegraphics[width=0.28\textwidth]{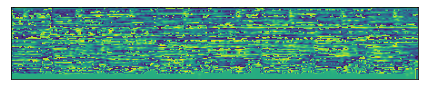}
&
\includegraphics[width=0.28\textwidth]{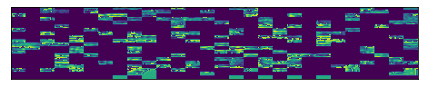}
& 
\includegraphics[width=0.28\textwidth]{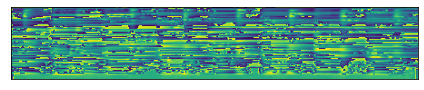}\\
\includegraphics[width=0.28\textwidth]{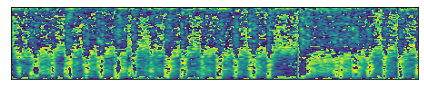}
&
\includegraphics[width=0.28\textwidth]{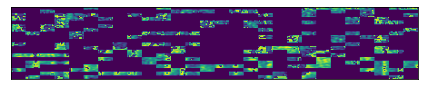}
& 
\includegraphics[width=0.28\textwidth]{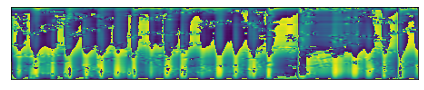}\\
\end{tabular}
}
\captionof{figure}{Examples of \textbf{audio spectrogram} reconstruction ability of XKD from highly masked ($70\%$) inputs. 
}
\label{fig:spec_recon}
\end{table*}

\clearpage

\end{document}